%% file: arxiv.tex
\documentclass{article}
\usepackage[T1]{fontenc}
\usepackage{textcomp}
\usepackage{iclr2027_conference,times}
\input{math_commands.tex}

\usepackage{amsmath,amssymb}
\usepackage{booktabs}
\usepackage{tabularx}
\usepackage{graphicx}
\usepackage{microtype}
\usepackage{xspace}
\usepackage[dvipsnames]{xcolor}
\usepackage[normalem]{ulem}
\usepackage{multirow}
\usepackage{wrapfig}
\usepackage{placeins}
\usepackage{tikz}
\usetikzlibrary{arrows.meta,positioning,fit,calc}
\usepackage{algorithm}
\usepackage{algpseudocode}
\usepackage{hyperref}
\usepackage{url}
\usepackage{adjustbox}

\title{\textsc{Nav4Manip}: From Navigation to Dexterous Mobile Manipulation through Co-Evolution}

\author{%
\bfseries
Yusen Feng\textsuperscript{1,2}\thanks{Equal contribution.}\quad
Zixi Kang\textsuperscript{1,2}\footnotemark[1]\quad
Zhijun Yin\textsuperscript{1}\footnotemark[1]\quad
Changxiao Liu\textsuperscript{1}\quad
Tianhong Gao\textsuperscript{1,2}\\
\bfseries
Jiazhao Zhang\textsuperscript{1}\quad
Jiahang Liu\textsuperscript{1,2}\quad
Libin Liu\textsuperscript{1}\thanks{Corresponding authors.}\quad
Zhizheng Zhang\textsuperscript{2}\footnotemark[2]\quad
He Wang\textsuperscript{1,2}\footnotemark[2]\\[0.4em]
\mdseries
\textsuperscript{1}Peking University\qquad
\textsuperscript{2}Galbot\qquad
}

\iclrfinalcopy

\newcommand{\method}{\textsc{Nav4Manip}\xspace}
\newcommand{\navmodel}{\textsc{Nav4Manip-N}\xspace}
\newcommand{\manipmodel}{\textsc{Nav4Manip-M}\xspace}
\newcommand{\kzx}[1]{{\color{black}#1}}
\newcommand{\yzj}[1]{{\color{purple}#1}}
\newcommand{\kzxp}[1]{{\color{black}#1}}
\usepackage[most]{tcolorbox}
\usepackage{listings}

\lstdefinestyle{promptstyle}{
    basicstyle=\ttfamily\small,
    breaklines=true,
    columns=fullflexible,
    keepspaces=true,
    showstringspaces=false,
    upquote=true,
    aboveskip=0pt,
    belowskip=0pt
}

\newtcblisting{promptbox}{
    enhanced,
    breakable,
    listing only,
    colback=white,
    colframe=black,
    boxrule=0.6pt,
    arc=0pt,
    outer arc=0pt,
    left=8pt,
    right=8pt,
    top=8pt,
    bottom=8pt,
    listing options={
        style=promptstyle
    }
}

\begin{document}
\maketitle
\lhead{}
\renewcommand{\headrulewidth}{0pt}
% \vspace{-10pt}
\begin{abstract}
Dexterous mobile manipulation is essential for robots to perform useful tasks in complex, unstructured environments. However, existing systems face two challenges. First, navigation remains too coarse for downstream manipulation, which often requires a precise base position and approach orientation rather than merely reaching a meter-scale goal region. Second, navigation and manipulation are commonly trained in isolation, lacking a feedback loop that enables the two capabilities to co-evolve. These limitations prevent current systems from reliably executing long-horizon dexterous mobile-manipulation tasks in real-world environments. To address them, we introduce \method, a bidirectionally coupled co-evolution system for long-horizon dexterous mobile manipulation. It comprises two complementary architectures: \navmodel learns long-horizon navigation and refines its policy through manipulation feedback, while \manipmodel trains a world-action model (WAM) for downstream dexterous manipulation. We train \navmodel in three stages: multi-task co-tuning learns navigation behavior and manipulation-aware affordance prediction; rejection sampling fine-tuning (RFT) rejects low-quality policy-generated paths and fine-tunes the model on retained trajectories; and reinforcement learning from AI feedback (RLAIF) uses counterfactual preferences to further improve navigation decisions. To support this framework, we establish the \textsc{Nav-N-Manip} benchmark and construct Nav4Manip-Pairs from benchmark rollouts, providing large-scale paired navigation--manipulation trajectories and affordance supervision. Extensive experiments on public vision-language navigation (VLN) benchmarks and \textsc{Nav-N-Manip} demonstrate state-of-the-art performance, while real-world evaluations show that \method completes diverse dexterous mobile-manipulation tasks. Project website: \url{https://nav4manip-web.github.io/}
\end{abstract}

\begin{figure*}[t]
\centering
\includegraphics[width=\textwidth]{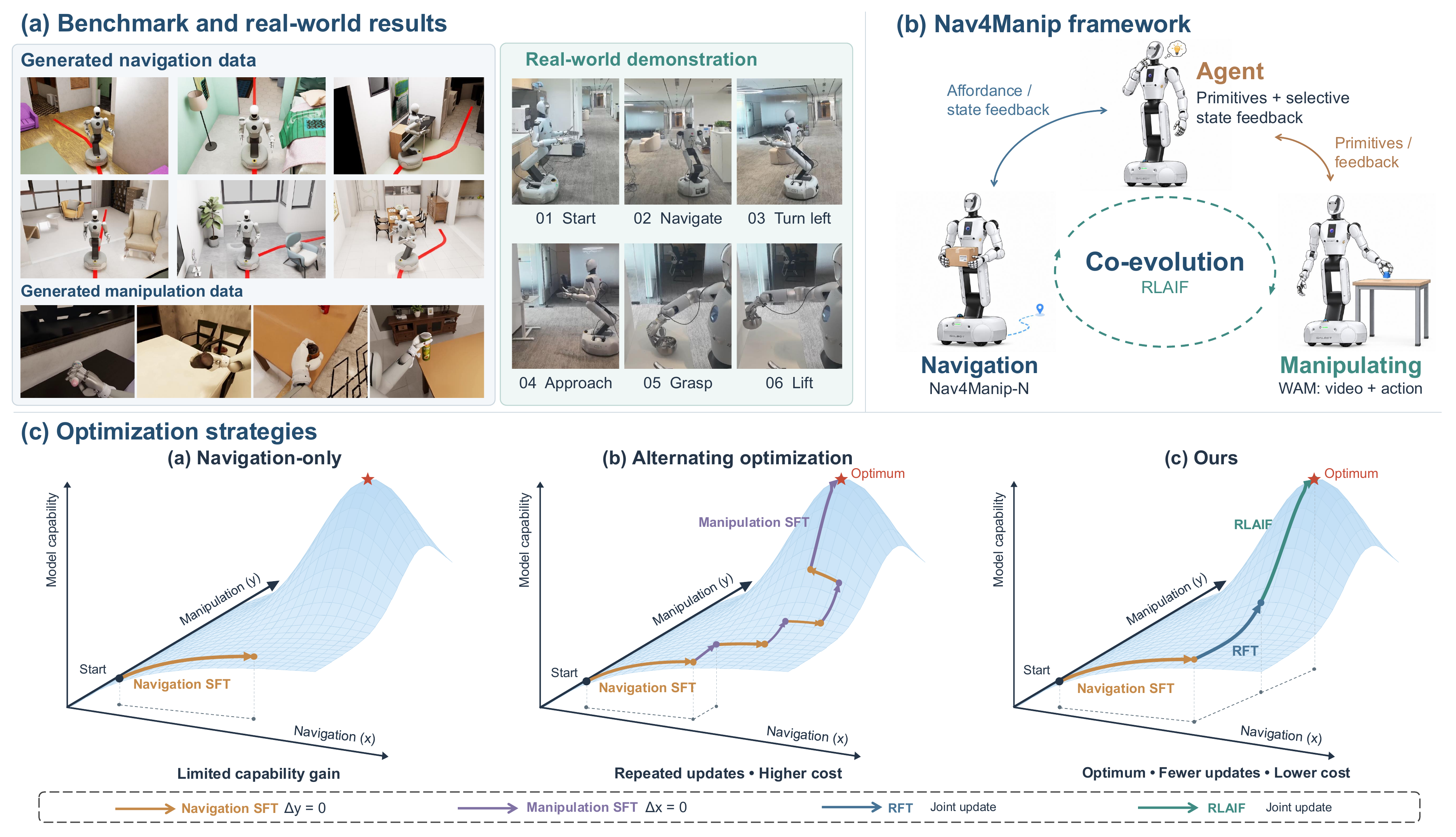}
\caption{\textbf{Overview of \method and its distinction from prior mobile-manipulation paradigms.} (a) The \textsc{Nav-N-Manip} benchmark provides generated navigation and manipulation data, alongside real-world demonstrations of long-horizon dexterous mobile manipulation. See Appendix~\ref{app:gallery_environments}, Appendix~\ref{app:gallery_assets}, and Appendix~\ref{app:gallery_real_mobile_manip} for more results. (b) \method couples \navmodel and \manipmodel through manipulation feedback, enabling navigation and manipulation to co-evolve.
(c) \kzxp{Comparison of three optimization strategies: navigation-only SFT, alternating optimization, and our co-evolution approach. By combining multi-task co-tuning, RFT, and RLAIF, our approach jointly improves navigation and manipulation with fewer optimization updates. See Appendix~\ref{app:task_definition} for details.}}
\label{fig:teaser}
% \vspace{-15pt}
\end{figure*}

\input{section/01_introduction}

\input{section/02_related_v2}

\input{section/04_method_v2}

\input{section/05_experiment}

\input{section/06_conclusion}

\newpage
\input{section/07_ethic}

\newpage
\bibliography{iclr2027_conference}
\bibliographystyle{iclr2027_conference}

\clearpage

\appendix
\renewcommand{\kzx}[1]{{\color{black}#1}}
\renewcommand{\yzj}[1]{{\color{purple}#1}}
\input{section/11_taskdefinition}

\input{section/12_systemarch}

\input{section/13_ablation}

\input{section/14_datagen}

\input{section/15_gallery}

\input{section/16_deploy}

\input{section/17_wam}

\input{section/18_prompt}
\end{document}

%% file: math_commands.tex
\usepackage{amsmath,amsfonts,bm}

\def\eqref#1{equation~\ref{#1}}
\def\1{\bm{1}}

\DeclareMathAlphabet{\mathsfit}{\encodingdefault}{\sfdefault}{m}{sl}
\SetMathAlphabet{\mathsfit}{bold}{\encodingdefault}{\sfdefault}{bx}{n}

%% file: section/01_introduction.tex
\section{Introduction}

Dexterous mobile manipulation integrates navigation~\citep{zhang2024navid,zhang2025uninavid,zhang2025navfom,xue2025omninav} and manipulation~\citep{brohan2023rt1,chi2023diffusionpolicy,octo2024,ye2026dreamzero,yuan2026fastwam}. However, existing VLN methods are insufficient as upstream policies for downstream manipulation. First, VLN benchmarks typically consider navigation successful when the robot stops within 3~m of a reference goal~\citep{anderson2018r2r,krantz2020vlnce}. In contrast, manipulation success is highly sensitive to the robot's terminal base pose, as workspace geometry, kinematic reachability, visibility, and collision constraints impose much stricter requirements on the final position and orientation. Consequently, an endpoint deemed successful by VLN metrics may still leave the target unreachable, occluded, or accessible only from an unsuitable approach direction. Second, existing mobile-manipulation systems typically adopt either a unified whole-body policy for room-scale tasks~\citep{fu2024mobilealoha,physicalintelligence2025pi05} or separately trained navigation and manipulation modules~\citep{yenamandra2023homerobot,chen2024okrobot}. \kzxp{These approaches primarily focus on gripper-based manipulation rather than dexterous multi-finger interaction. Additionally, }unified whole-body policies do not directly address long-horizon navigation, while modular pipelines lack a mechanism for manipulation outcomes to inform and refine the navigator. Moreover, independently fine-tuning navigation and manipulation policies provides no mechanism for the two capabilities to co-evolve toward a shared end-to-end objective.

We address these limitations by introducing \method, a bidirectionally coupled co-evolution system with two complementary architectures: \navmodel learns long-horizon navigation and \kzxp{refines its policy through manipulation feedback, while \manipmodel trains a WAM for downstream dexterous manipulation.} As illustrated in Figure~\ref{fig:teaser}, navigation-only optimization yields limited joint capability, while alternating optimization repeatedly trains both modules. Our staged co-evolution strategy achieves a 17.83-point improvement in end-to-end task success (52.13\% vs. 34.30\%) over alternating optimization, while avoiding repeated manipulation policy training. \navmodel jointly predicts navigation actions and manipulation affordances. Multi-task co-tuning first establishes broad navigation competence and manipulation endpoint prediction. \kzxp{We then introduce verifier-guided RFT, which samples trajectories from \navmodel and uses downstream manipulation executions as a verifier to retain trajectories with feasible navigation endpoints and successful manipulation outcomes.} Finally, RLAIF uses an execution agent to interpret navigation--manipulation outcomes and construct counterfactual preference pairs for further optimizing \navmodel. The resulting knowledge is distilled into the navigation model, enabling manipulation feedback to improve future navigation.

To support training and evaluation, we establish \textsc{Nav-N-Manip}, a benchmark that evaluates both long-horizon navigation and downstream manipulation. Rollouts on \textsc{Nav-N-Manip} yield Nav4Manip-Pairs, a 233-hour paired navigation--manipulation dataset. In summary, we make the following contributions:
\begin{itemize}
    \item We introduce \method, a dexterous mobile-manipulation co-evolution system comprising three-stage \navmodel and WAM \textsc{Nav4Manip-M}.
    \item We establish the \textsc{Nav-N-Manip} benchmark for long-horizon dexterous mobile manipulation and construct Nav4Manip-Pairs, a 233-hour dataset of paired navigation--manipulation trajectories and affordance scores.
    \item We conduct extensive experiments on established VLN tasks and \textsc{Nav-N-Manip}, together with real-robot evaluations, demonstrating strong navigation performance and reliable end-to-end dexterous mobile manipulation across diverse tasks.
\end{itemize}

%% file: section/02_related_v2.tex
\section{Related Work}

\paragraph{\textbf{\textsc{Vision-Language Navigation Datasets}.}}
Built on Matterport3D~\citep{Matterport3D}, R2R~\citep{anderson2018r2r}, R4R~\citep{jain2019r4r}, and RxR~\citep{ku2020rxr} provide increasingly long routes and multilingual grounding. VLN-CE~\citep{krantz2020vlnce} enables continuous navigation, while REVERIE~\citep{qi2020reverie} and OpenUAV~\citep{wang2025towards} extend evaluation to object grounding and aerial navigation. ScaleVLN~\citep{wang2023scalevln} scales synthetic supervision, and RoomTour3D~\citep{han2025roomtour3d} and Sekai~\citep{li2026sekai} leverage exploration videos. VLN-PE~\citep{wang2025vlnpe}, VLNVerse~\citep{lin2025vlnverse}, and InternData-N1~\citep{wei2025internvla} further broaden physical realism, embodiment coverage, and scene diversity. Although these efforts substantially expand navigation supervision and some consider downstream embodied tasks, most VLN benchmarks do not explicitly evaluate whether terminal poses support manipulation. Nav4Manip-Pairs addresses this gap by pairing navigation paths and candidate endpoints with manipulation trajectories and rollout-derived manipulation success rates.

\paragraph{\textbf{\textsc{Vision-Language Navigation Algorithms}.}}
VLN methods have progressed from recurrent policies and data augmentation~\citep{anderson2018r2r,fried2018speaker,tan2019learning} to navigation pre-training and history modeling~\citep{hao2020prevalent,hong2020recurrent,guhur2021airbert,chen2021hamt}, followed by explicit spatial memory and planning~\citep{chen2022duet,an2023bevbert,an2024etpnav}. Recent foundation-model approaches extend these capabilities along several directions. For reasoning and agentic planning, NavGPT~\citep{zhou2023navgpt}, MapGPT~\citep{chen2024mapgpt}, and InstructNav~\citep{long2024instructnav} prompt large models to interpret instructions and construct navigation plans, while JanusVLN~\citep{zeng2026janusvln} improves long-horizon reasoning through compact semantic and spatial memories. For embodied control, NaVid~\citep{zhang2024navid} and StreamVLN~\citep{wei2025streamvln} operate over streaming observations, NaVILA~\citep{cheng2025navila} connects navigation to legged locomotion, and DualVLN~\citep{wei2025internvla} separates high-level grounding from real-time local control. For multi-task and cross-embodiment generalization, NaviLLM~\citep{zheng2024navillm}, Uni-NaVid~\citep{zhang2025uninavid}, and NavFoM~\citep{zhang2025navfom} combine navigation with broader embodied supervision, while OmniVLA~\citep{hirose2025omnivla}, ABot-N0/N1~\citep{chu2026abotn0,gong2026abotn1}, InternVLA-N1~\citep{wei2025internvla}, and Qwen-RobotNav~\citep{zhang2026qwenrobotnav} further expand task and embodiment coverage. \kzxp{Our \navmodel learns more precise, manipulation-aware navigation and refines its policy through downstream manipulation execution feedback.}

\paragraph{\textbf{\textsc{Mobile Manipulation}.}}
\kzxp{HomeRobot~\citep{yenamandra2023homerobot}, OK-Robot~\citep{chen2024okrobot}, Mobile ALOHA~\citep{fu2024mobilealoha}, and $\pi_{0.5}$~\citep{physicalintelligence2025pi05} enable mobile manipulation through whole-body policies.} Humanoid systems such as EgoHumanoid~\citep{shi2026egohumanoid}, WholeBodyVLA~\citep{jiang2025wholebodyvla}, $\Psi_0$~\citep{wei2026psi0}, and $\omega$-0~\citep{li2026omega0} further unify locomotion and manipulation control. Focusing on the navigation--manipulation interface, MoMa-Kitchen~\citep{zhang2025momakitchen} and BaSeNet~\citep{naik2024basenet} investigate manipulation-aware base placement. However, they do not address how long-horizon navigation and dexterous manipulation can be coupled. \method addresses this gap through staged co-evolution: paired navigation--manipulation training aligns the two modules, \kzxp{while downstream manipulation outcomes progressively refine the navigation policy, improving end-to-end performance on long-horizon dexterous mobile-manipulation tasks.}

%% file: section/04_method_v2.tex
\section{\method}

\kzxp{We present \method, a manipulation-aware navigation framework that integrates automated navigation--manipulation data construction, and progressive refinement through multi-task co-tuning, verifier-guided RFT, and agentic counterfactual RLAIF.}

\subsection{Nav4Manip-Pairs Construction}
\label{sec:nav4manip_pairs_method}
\begin{figure*}[t]
    \centering
    \includegraphics[width=\textwidth]{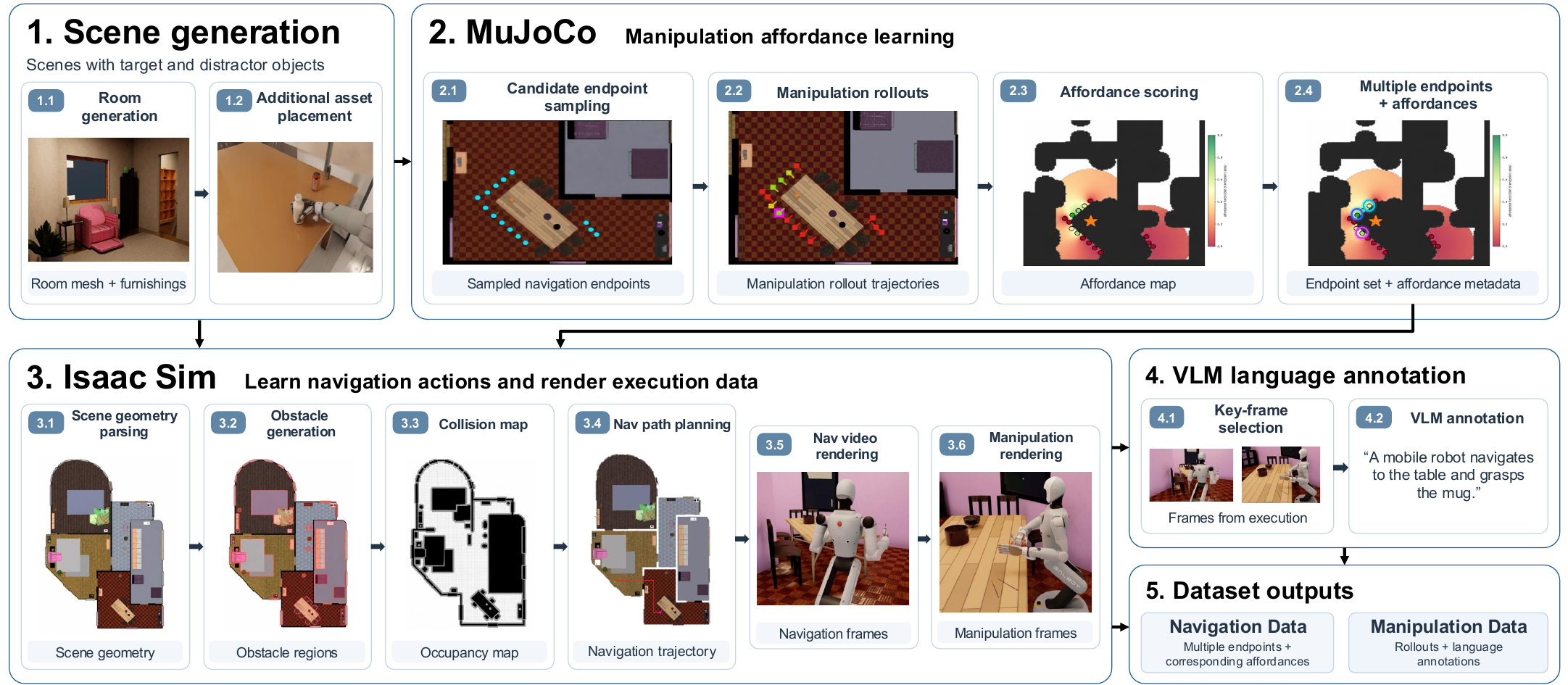}
    % \vspace{-10pt}
    \caption{\textbf{Nav4Manip-Pairs data-generation pipeline.} The pipeline constructs paired navigation trajectories, manipulation-validated endpoints, and aligned vision--language annotations.}
    \label{fig:overview}
    % \vspace{-10pt}
\end{figure*}

Figure~\ref{fig:overview} summarizes our automated data-generation pipeline. We combine 221 EmbodiedGen v2 houses~\citep{wang2026embodiedgen} with 176 scenes from the VLNVerse training set~\citep{lin2025vlnverse}, insert task objects, and standardize the scene configurations across MuJoCo and Isaac Sim. In MuJoCo, candidate base poses are proposed from the target-object location and surrounding support surface, then evaluated through repeated manipulation rollouts with object-pose perturbations to estimate endpoint affordances. Each selected navigation-reachable endpoint is paired with a collision-validated navigation path. The aligned navigation and manipulation trajectories are replayed in Isaac Sim to render synchronized observations, after which a VLM generates task instructions. The resulting records connect navigation trajectories with downstream manipulation outcomes. Full generation and validation details are provided in Appendix~\ref{app:data_affordance} to Appendix~\ref{app:quality_control}.

For each task, candidate endpoints around the support surface are evaluated through repeated manipulation rollouts with object-pose perturbations. We represent each Nav4Manip-Pairs sample $\mathcal D_i$ as
\begin{equation}
\begin{aligned}
\mathcal D_i&=\left(l_i,\mathcal N_i,\mathcal M_i\right),\\
\mathcal M_i&=\left(q_i,\{\tau_{i,k}^{m}\}_{k=1}^{K_i}\right),
\end{aligned}
\label{eq:nav4manip_record}
\end{equation}
where $l_i$ denotes the task instruction, $\mathcal N_i$ contains the navigation trajectory and associated annotations, \kzxp{and $\mathcal M_i$ comprises the endpoint-affordance score $q_i$ and $K_i$ successful manipulation trajectories, each consisting of state-action sequences. Both successful and unsuccessful endpoint candidates are retained for affordance prediction and ranking, while feasible endpoints are paired with validated navigation trajectories and successful manipulation rollouts.} The exact contents of $\mathcal N_i$ and $\mathcal M_i$ are provided in Appendix~\ref{app:data_architecture}.

We use both parts of each paired record for training: the navigation record $\mathcal N_i$ and the endpoint-affordance score $q_i$ in $\mathcal M_i$ jointly supervise navigation and endpoint affordances, while the successful manipulation trajectories in $\mathcal M_i$, their aligned observations, and task instructions supervise \manipmodel. Appendix~\ref{app:manipulation_deployment} gives the manipulation policy details.

\begin{figure}[t]
    \centering
    \includegraphics[width=\textwidth]{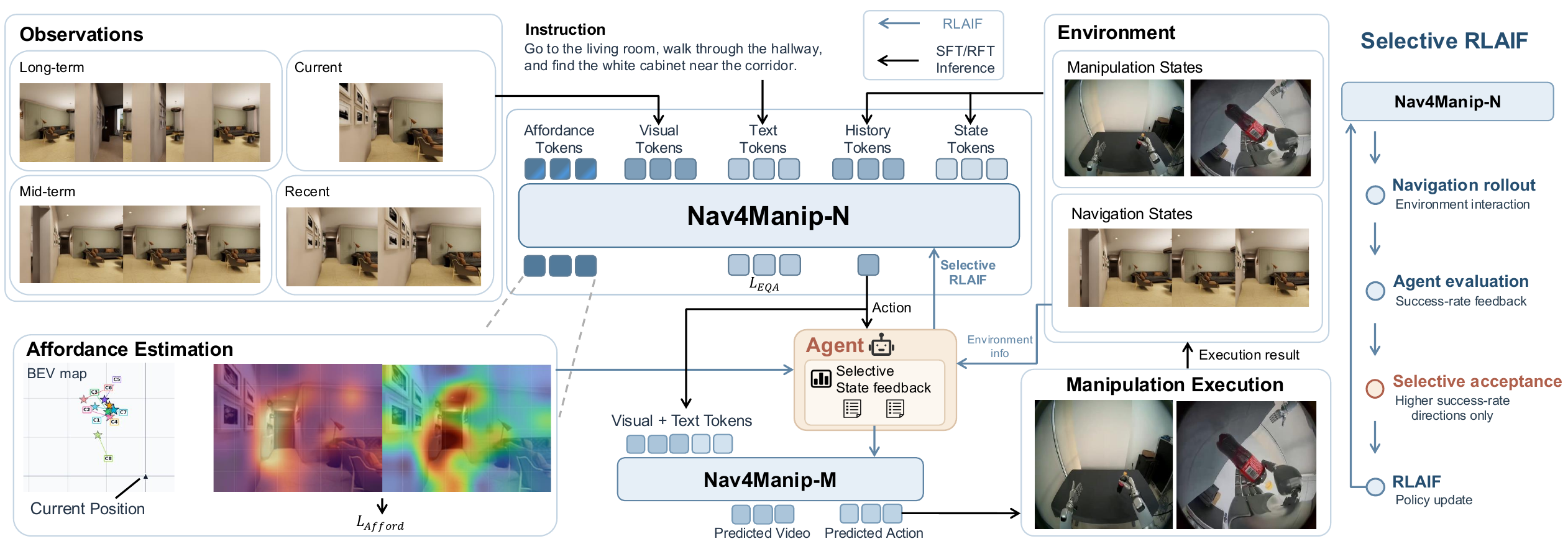}
    % \vspace{-10pt}
    \caption{\textbf{Overview of \method.} \navmodel jointly predicts navigation actions and manipulation-aware endpoint affordances. The bird's-eye-view (BEV) map shows the robot and endpoint candidates in ground-plane coordinates. \kzxp{The two affordance heatmaps visualize manipulation success rates across candidate object positions and robot navigation endpoints, respectively, indicating where the object can be placed and where the robot should navigate for successful manipulation.} Co-tuning, RFT, and counterfactual RLAIF progressively improve the policy, whose terminal pose is handed directly to \manipmodel at inference.}
    \label{fig:method}
    % \vspace{-10pt}
\end{figure}

\subsection{Manipulation-Aware Navigation Pipeline}
\label{sec:affordance_architecture}

As shown in Figure~\ref{fig:method}, at navigation step $t$, \navmodel receives learnable affordance queries, multi-scale visual tokens, instruction tokens, task-relevant history, and the current robot and environment state. The history summarizes temporally sampled observations and actions and, during RLAIF data collection, may additionally incorporate selected navigation--manipulation feedback. \kzxp{We denote this complete navigation context by $h_t^n$ and the selected navigation action by $a_t^n$. The navigation and manipulation policies are formulated as}
\begin{equation}
a_t^n \sim \pi_\theta(\cdot \mid h_t^n), \qquad
a_t^m \sim \pi_m(\cdot \mid h_t^m).
\label{eq:nav4manip_context_v2}
\end{equation}

Here, $\pi_\theta$ denotes the \navmodel navigation policy used throughout co-tuning, RFT, and RLAIF, while $\pi_m$ denotes the \manipmodel manipulation policy, activated by an agent and conditioned on its own manipulation context $h_t^m$. \kzxp{During RLAIF, an execution agent gates the invocation of $\pi_m$ based on navigation--manipulation memories and refine navigation endpoint.} Built upon a multimodal backbone initialized from Qwen3-VL~\citep{bai2025qwen3}, \navmodel predicts navigation actions, grounded EQA answers, a dense ego-centered affordance map, and image-grounded endpoint proposals to guide the robot toward manipulation-ready endpoints. Appendix~\ref{app:video_interface}, Appendix~\ref{app:nav4manip_architecture}, and Appendix~\ref{app:wam_manipulation_executor} provide further implementation details.

\subsection{Multi-Task Co-Tuning}
\label{sec:cotune_method}

Table~\ref{tab:data_v2} summarizes the composition of our multi-task co-tuning mixture. Our multi-task co-tuning mixture contains 4.9M samples (625.08 hours) from RxR~\citep{ku2020rxr}, HM3D ObjectNav~\citep{yadav2023hm3d}, R2R~\citep{anderson2018r2r}, our rendered EQA data, and Nav4Manip-Pairs. Public navigation datasets provide broad action supervision, the rendered EQA data improve grounded visual reasoning, and Nav4Manip-Pairs jointly supervise navigation and manipulation behaviors

\begin{table}[t]
\caption{Composition of the multi-task co-tuning mixture.  For Nav4Manip-Pairs, ``action + affordance'' denotes joint supervision of navigation behavior and manipulation-valid endpoints.  Policy-generated RFT records are collected separately after co-tuning.}
\label{tab:data_v2}
\centering
\small
\setlength{\tabcolsep}{3.5pt}
\resizebox{0.8\textwidth}{!}{%
\begin{tabular}{lrrrl}
\toprule
Source & Duration (h) & Samples & Episodes & Supervision \\
\midrule
RxR                    & 174.67 & 1.9M & 6,668  & action tokens \\
HM3D ObjectNav         & 115.66 & 416K & 6,603  & action tokens \\
R2R                    &  58.39 & 631K & 3,603  & action tokens \\
Rendered EQA (ours)    &  43.26 & 156K & 8,976  & answer text \\
Nav4Manip-Pairs (ours) & 233.11 & 1.8M & 17,927 & action + affordance \\
\midrule
Total                  & 625.08 & 4.9M & 43,777 & \\
\bottomrule
\end{tabular}
}
% \vspace{-10pt}
\end{table}

We optimize the co-tuning objective
\begin{equation}
\mathcal L_{\mathrm{co}}=
\mathcal L_{\mathrm{nav}}
+\lambda_{\mathrm{Afford}}\mathcal L_{\mathrm{Afford}}
+\lambda_{\mathrm{EQA}}\mathcal L_{\mathrm{EQA}},
\label{eq:cotune_v2}
\end{equation}
where $\mathcal L_{\mathrm{nav}}$ is the token-level navigation-action loss, $\mathcal L_{\mathrm{Afford}}$ supervises object positions and endpoint predictions with rollout-derived outcomes, and $\mathcal L_{\mathrm{EQA}}$ trains grounded answer generation. 
\kzxp{The affordance objective jointly considers object placements and navigation endpoints, incorporating navigation reachability and downstream manipulation success to guide the policy toward manipulation-feasible navigation. Further details are provided in Appendix~\ref{app:nav4manip_architecture}.} Joint co-tuning preserves broad navigation competence while initializing the policy to reason about downstream manipulation feasibility.

\subsection{Verifier-Guided Rejection Sampling Fine-Tuning}
\label{sec:rft_method}

\kzxp{Building on the co-tuned policy, RFT improves navigation-manipulation performance by collecting clean on-policy trajectories without intermediate state resets, exploring alternative navigation paths and manipulation endpoints. Specifically, for each task, we execute $\pi_{\theta}$ in closed loop from the task start state to generate full on-policy navigation rollouts toward candidate manipulation endpoints, forming the RFT trajectory $P_i^{\text{RFT}}$ terminating at $b_i^{\text{RFT}}$.} Repeated manipulation trials at this endpoint yield its empirical affordance $q_i^{\mathrm{RFT}}$. The resulting record $\mathcal D_i^{\mathrm{RFT}}$ follows the same format as $\mathcal D_i$, with trajectory-dependent fields regenerated by \navmodel.

The verifier discards unreachable endpoints and collision-invalid paths. The remaining samples are ranked by
\begin{equation}
s_i^{\mathrm{RFT}}
=q_i^{\mathrm{RFT}}
-\lambda_{\mathrm{path}}\ell(P_i^{\mathrm{RFT}}),
\label{eq:rft_score_v2}
\end{equation}
where $q_i^{\mathrm{RFT}}$ measures manipulation feasibility, $\ell(P_i^{\mathrm{RFT}})$ is the path length, and $\lambda_{\mathrm{path}}$ balances affordance against path efficiency. High-scoring records form the retained set $\mathcal D_{\mathrm{RFT}}^{+}$.

After expanding the retained trajectories into step-level context--action pairs, RFT performs supervised fine-tuning on the retained navigation actions:
\begin{equation}
\mathcal L_{\mathrm{RFT}}
=-
\mathbb E_{(h,a^n)\sim\mathcal D_{\mathrm{RFT}}^{+}}
\left[
\log\pi_\theta\!\left(a^n\mid h\right)
\right].
\label{eq:rft_v2}
\end{equation}
Here, $(h,a^n)$ denotes a model-context and navigation-action pair from an accepted policy-generated trajectory. Initialized from $\pi_{\theta_{\mathrm{co}}}$, the policy $\pi_\theta$ minimizes the expected token-level negative log-likelihood over the retained pairs. Thus, manipulation outcomes determine which policy-generated trajectories serve as SFT targets.

\subsection{Agentic Counterfactual RLAIF}
\label{sec:rlaif_method}

In the final training stage, we perform RLAIF following the broader paradigm of improving embodied policies through AI-generated feedback~\citep{guo2026vlaw}. Our approach combines agentic counterfactual refinement with execution-based validation. Given a navigation--manipulation rollout, an execution agent $\mathcal C$ analyzes the execution feedback and proposes a navigation correction, such as moving closer, adjusting the approach yaw, or reapproaching around an obstruction. We then execute an on-policy episode rollout under the refined navigation decision in closed loop and evaluate its terminal outcome using \manipmodel $\mathcal W$. This data collection and policy update process is repeated for $K_{\mathrm{iter}}$ RLAIF rounds.
\kzxp{For each task instruction $l_i$, we construct a counterfactual context $x_i$ consisting of the instruction, the navigation action, and visual observations capturing both navigation and manipulation states.} Based on this context, the agent proposes a refined navigation decision:
\begin{equation}
y_i^{+}=\mathcal C(x_i),
\end{equation}
where $y_i^{+}$ is the refined navigation decision. Let $r_i^{-}$ and $r_i^{+}$ be the manipulation scores before and after refinement. We retain only execution-validated improvements:
\begin{equation}
\Delta r_i=r_i^{+}-r_i^{-},
\qquad
\mathcal D_{\mathrm{AI}}=
\left\{(x_i,y_i^{+})\mid
\Delta r_i\geq\tau_{\Delta}\right\}.
\label{eq:rlaif_filter}
\end{equation}
We then fine-tune \navmodel on the retained refinements using supervised learning:
\begin{equation}
\mathcal L_{\mathrm{RLAIF}}
=
-\mathbb E_{(x,y^{+})\sim\mathcal D_{\mathrm{AI}}}
\left[
\log \pi_\theta(y^{+}\mid x)
\right],
\label{eq:rlaif_loss}
\end{equation}
where $\pi_\theta$ is initialized from the RFT navigator. This objective distills AI-generated, execution-validated refinements into \navmodel, so neither the agent nor counterfactual evaluation is required at inference time. Algorithm~\ref{alg:counterfactual_rlaif} summarizes the complete procedure. Appendix~\ref{app:agenticwam_system} and Appendix~\ref{app:training_implementation} provide the agent loop, prompts, and training settings.

\begin{algorithm}[t]
\caption{Execution-Validated Agentic Counterfactual RLAIF}
\label{alg:counterfactual_rlaif}
\begin{algorithmic}[1]
\Require Navigator $\pi_\theta$, agent $\mathcal C$, frozen \manipmodel $\mathcal W$,
         instructions $\{l_i\}$, threshold $\tau_\Delta$, rounds $K_{\mathrm{iter}}$
\Ensure Refined navigation policy $\pi_{\theta^*}$

\For{$k = 1, \ldots, K_{\mathrm{iter}}$}
    \State $\mathcal D_{\mathrm{AI}} \gets \varnothing$
    \For{each instruction $l_i$}
        \State $(x_i, r_i^-) \gets
            \Call{ClosedLoopEvaluate}{\pi_\theta, \mathcal W, l_i}$
        \State $y_i^+ \gets \mathcal C(x_i)$
        \State Execute full on-policy rollout for the refined decision from the episode start
        \State $r_i^+ \gets
            \Call{EvaluateRefinement}{\pi_\theta,\mathcal W,l_i,x_i,y_i^{+}}$
        \If{$r_i^+ - r_i^- \geq \tau_\Delta$}
            \State $\mathcal D_{\mathrm{AI}} \gets
                \mathcal D_{\mathrm{AI}} \cup \{(x_i,y_i^+)\}$
        \EndIf
    \EndFor
    \State Update $\pi_\theta$ on $\mathcal D_{\mathrm{AI}}$ using $\mathcal L_{\mathrm{RLAIF}}$
\EndFor

\State \Return $\pi_{\theta^*}$
\end{algorithmic}
\end{algorithm}

%% file: section/05_experiment.tex
\section{Experiments}

\kzxp{We comprehensively evaluate \method through standard VLN benchmarks, simulated and real-world mobile manipulation tasks, and training-pipeline ablations, assessing its navigation generalization, manipulation-aware endpoint precision, and end-to-end task success.}

\subsection{Setup}

We evaluate closed-loop continuous navigation on the val-unseen splits of R2R and RxR and on the official test split of the fine- and coarse-grained VLNVerse tasks. We further evaluate closed-loop navigation and manipulation on our \textsc{Nav-N-Manip} benchmark. At each terminal pose in \textsc{Nav-N-Manip}, we evaluate manipulation over 16 independently reset trials. SR@Manip reports average single-attempt manipulation success, while SR@E2E measures joint navigation-to-grasp success, requiring arrival within 0.6 m of the reference goal and successful grasping; navigation failures count as failures. Following R2R conventions, we report success rate (SR), oracle success rate (OSR), success weighted by path length (SPL), navigation error (NE), and trajectory length (TL). Navigation SR@$r$ measures the percentage of episodes whose final agent position is within $r$ meters of the reference goal, using the benchmark's prescribed distance metric. Unless otherwise noted, R2R and RxR use the standard 3~m success radius; for our benchmark, we additionally report results at radii of 1.5~m, 1~m, and 0.6~m. \navmodel operates on monocular RGB observations in a closed-loop manner, executing one discrete navigation action before observing and replanning. \kzxp{For fair comparison across observation settings, we distinguish single-view (S) and multi-view (M) methods in our results, allowing direct comparison under the single-view protocol while retaining multi-view baselines for reference.} Full navigation architecture, agent loop, and stage-wise training configurations are detailed in Appendices~\ref{app:video_interface}--\ref{app:training_implementation} (Table~\ref{tab:training_implementation}); simulation benchmark splits, affordance evaluation, and trajectory generation are specified in Appendices~\ref{app:data_architecture}--\ref{app:quality_control}; and real-world deployment, manipulation post-training, and inference protocols are provided in Appendix~\ref{app:real_world_deployment} (Section~\ref{app:navigation_deployment}) and Appendix~\ref{app:manipulation_deployment} (Sections~\ref{app:wam_manipulation_executor}--\ref{app:manipulation_rtc}).

\subsection{Main Results}

\begin{figure*}[t]
    \centering
    \includegraphics[width=\linewidth]{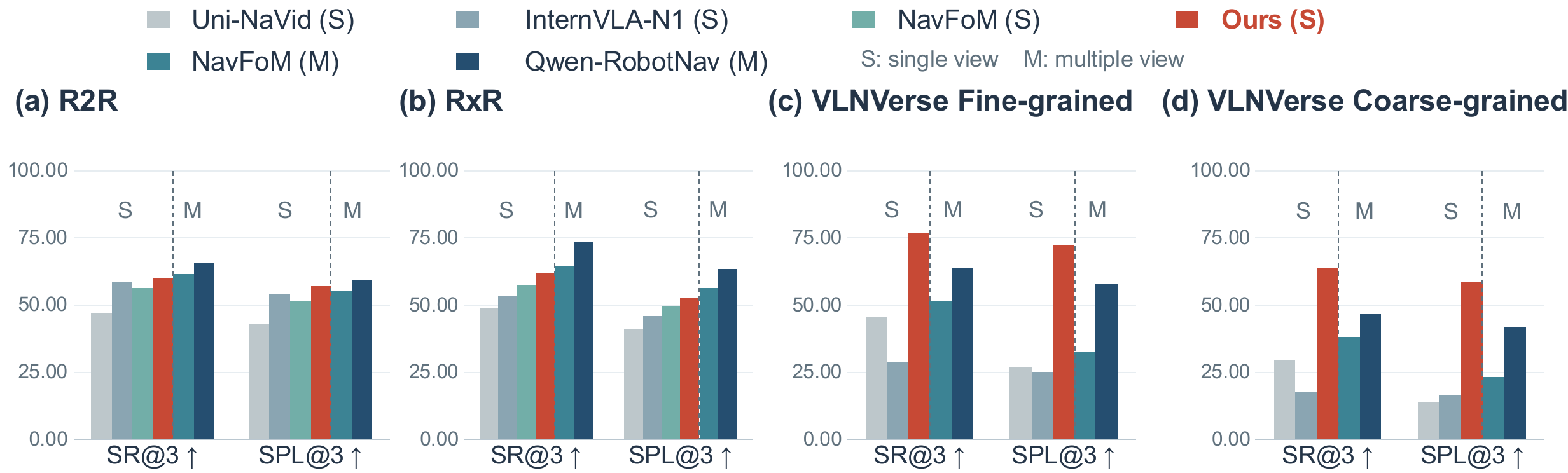}
    % \vspace{-10pt}
    \caption{Closed-loop navigation comparison on (a) R2R val-unseen, (b) RxR val-unseen, and the VLNVerse test split for (c) fine-grained and (d) coarse-grained navigation. Each panel reports SR@3 and SPL@3 (higher is better). S and M denote single-view and multi-view visual inputs, respectively. Complete numerical results are provided in Appendix~\ref{app:full_navigation_results}.}
    \label{fig:navigation_results}
    % \vspace{-10pt}
\end{figure*}

Figure~\ref{fig:navigation_results} presents a unified comparison on R2R and RxR val-unseen and on the official test split of the fine- and coarse-grained VLNVerse tasks. We report the two primary metrics shared by all settings: success rate within 3~m (SR@3) and success weighted by path length at the same threshold (SPL@3). Appendix Tables~\ref{tab:main} and~\ref{tab:vlnverse} report the complete benchmark-specific metrics, including OSR, NE, and trajectory length where available. Among single-view methods, \navmodel achieves the best SR@3 and SPL@3 on both R2R (60.03/56.79) and RxR (62.06/52.60). It also matches the multi-view NavFoM closely: on R2R, its SPL@3 is higher (56.79 vs. 55.30), while on RxR its SR@3 and SPL@3 are within 2.34 and 3.60 points, respectively. On the VLNVerse test split, \navmodel further surpasses the strongest multi-view baseline, Qwen-RobotNav, by 12.33/13.66 points in SR@3/SPL@3 on the fine-grained task and 16.58/16.95 points on the coarse-grained task.

We additionally evaluate whether navigation endpoints support downstream grasping in simulation. Table~\ref{tab:sim_grasp} compares instruction-following and ObjectNav task formulations. For this comparison, all models are fine-tuned for one epoch on the \textsc{Nav-N-Manip} training split and evaluated on its test split. Under instruction following, \method achieves 71.3\% SR@0.6 and 56.31\% end-to-end success; under ObjectNav, it achieves 71.7\% SR@0.6 and 48.15\% end-to-end success. Appendix Table~\ref{tab:downstream_agent_comparison} further compares our full system with general-purpose agents on ten \textsc{Nav-N-Manip} episodes. GPT-6 Astra reaches all ten goals within 1.5~m, nine within 1.0~m, and six within 0.6~m, yielding 5/10 manipulation successes with 445~s navigation latency. Our system succeeds in 8/10 episodes at every radius, completes manipulation in 7/10, and reduces navigation latency to 14~s.

\begin{table*}[t]
\caption{Simulation navigation-to-grasp results under instruction following and ObjectNav. All external navigation baselines use the same downstream manipulation interface as \textsc{Nav4Manip} for grasp execution and evaluation. All metrics follow the evaluation protocol in Section 4.1. Best/second-best results per setting are bolded/underlined.}
\label{tab:sim_grasp}
\centering
\scriptsize
\setlength{\tabcolsep}{3.5pt}
\resizebox{\textwidth}{!}{%
\begin{tabular}{llccccccc}
\toprule
Model & Task & SR@3.0 $\uparrow$ & SR@1.5 $\uparrow$ & SR@1.0 $\uparrow$ & SR@0.6 $\uparrow$ & NE $\downarrow$ & SR@Manip $\uparrow$ & SR@E2E $\uparrow$\\
% \midrule
% GPT-6 Astra & end-to-end & \kzx{todo} & \kzx{todo} & \kzx{todo} & \kzx{todo} & \kzx{todo} & \kzx{todo}\\
\midrule
OmniVLA & instruction & 0.8 & 0 & 0 & 0 & 7.29 & 0 & 0\\
InternVLA-N1 & instruction & 3.5 & 0.4 & 0.4 & 0 & 6.47 & 0 & 0\\
NaVid & instruction & \underline{35.7} & \underline{14.7} & 8.1 & 4.7 & \underline{4.33} & 0.78 & 0.39\\
Uni-NaVid & instruction & 30.2 & 14.0 & \underline{11.2} & \underline{5.0} & 5.07 & \underline{1.94} & \underline{1.94}\\
\method & instruction & \textbf{90.3} & \textbf{85.3} & \textbf{81.8} & \textbf{71.3} & \textbf{1.12} & \textbf{61.17} & \textbf{56.31}\\
\midrule
OmniVLA & ObjectNav & 0.8 & 0 & 0 & 0 & 7.26 & 0 & 0\\
InternVLA-N1 & ObjectNav & 2.7 & 0.4 & 0 & 0 & 6.97 & 0 & 0\\
NaVid & ObjectNav & \underline{37.2} & \underline{18.2} & 9.7 & 4.7 & \underline{4.37} & 0.78 & 0\\
Uni-NaVid & ObjectNav & 28.7 & 12.4 & \underline{10.9} & \underline{5.0} & 4.48 & \underline{3.88} & \underline{2.71}\\
\method & ObjectNav & \textbf{88.8} & \textbf{83.0} & \textbf{79.1} & \textbf{71.7} & \textbf{1.18} & \textbf{52.78} & \textbf{48.15}\\
\bottomrule
\end{tabular}
}
% \vspace{-10pt}
\end{table*}

\begin{figure*}[t]
    \centering
    \includegraphics[width=\linewidth]{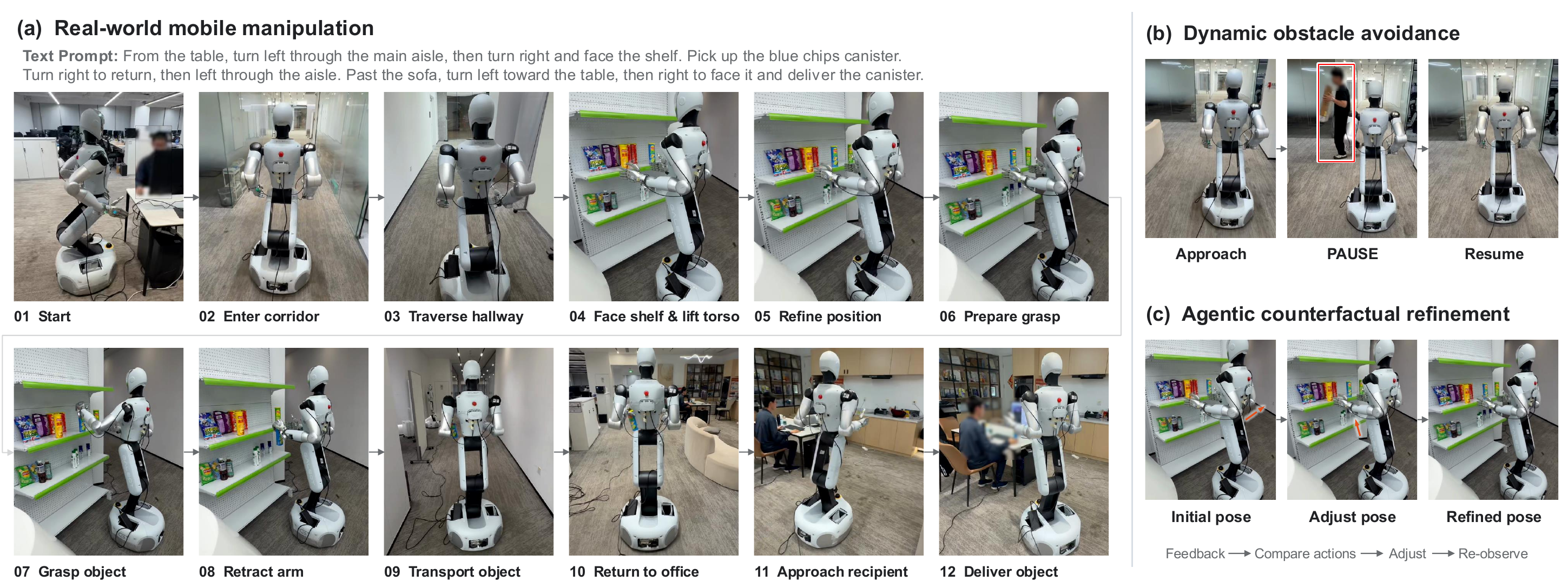}
    % \vspace{-10pt}
    \caption{\textbf{Real-world navigation and dexterous mobile manipulation.} (a) Long-horizon navigation and object delivery. (b) Dynamic obstacle avoidance. (c) Agentic counterfactual base-pose refinement. See Appendices~\ref{app:real_world_deployment} and~\ref{app:real_world_demonstrations} for deployment details and demonstration tasks.}
    \label{fig:demo}
    % \vspace{-10pt}
\end{figure*}

Figure~\ref{fig:demo} demonstrates the complete system on real-world dexterous mobile-manipulation tasks. In Figure~\ref{fig:demo}(a), the robot navigates to a shelf, refines its base pose, grasps and transports the target object, and delivers it to the recipient. Figure~\ref{fig:demo}(b) shows dynamic obstacle handling: the robot pauses when a person enters its path and resumes after the path clears. Figure~\ref{fig:demo}(c) illustrates agentic counterfactual refinement, where the agent compares corrective actions using task feedback, adjusts the base pose, and re-observes the scene before manipulation.

\subsection{Ablation of the Training Pipeline}

\kzxp{Table~\ref{tab:training_ablation} evaluates our training pipeline through progressive component ablations and comparisons with alternative optimization strategies.} The baseline is trained on RxR, HM3D ObjectNav, and R2R. Multi-task co-tuning (MTC) additionally incorporates our generated data, comprising Rendered EQA and Nav4Manip-Pairs; verifier-guided RFT then learns from manipulation-verified policy rollouts. MTC provides the largest initial gain, raising SR@0.6 from 10.7 to 52.7 and SR@E2E from 4.07 to 31.01. RFT further improves precise stopping: relative to MTC, SR@1.0 increases from 69.8 to 73.8, SR@0.6 from 52.7 to 60.9, and NE decreases from 1.81 to 1.76. These gains show that filtering policy-generated trajectories using downstream execution outcomes primarily improves the fine-grained endpoint accuracy required for manipulation.

Starting from the same RFT checkpoint, ACR performs counterfactual refinement at inference time without parameter updates, whereas RLAIF distills execution-validated refinements into the final \method policy. Both substantially improve the navigation--manipulation handoff: relative to RFT, ACR raises SR@Manip from 38.18 to 53.88 and SR@E2E from 31.79 to 47.10. 
% With two training iterations, RLAIF achieves the best result across all navigation and manipulation metrics, including 71.7 SR@0.6, 1.15~m NE, 56.87 SR@Manip, and 52.13 SR@E2E. 
With two training iterations, RLAIF achieves the best result across all navigation and manipulation metrics, including 71.7 SR@0.6, 1.15 m NE, 56.87 SR@Manip, and 52.13 SR@E2E. Notably, skipping the RFT stage and applying RLAIF directly to the co-tuned model (\texttt{+ MTC + RLAIF}) reaches only 43.02\% SR@E2E, confirming that RFT provides an essential behavioral prior that stabilizes subsequent RLAIF optimization.
With matched per-stage training sample counts, two rounds of alternating navigation--manipulation optimization require four module-training stages but achieve only 34.30\% SR@E2E, compared with 52.13\% for two-round RLAIF with a fixed manipulation policy. Compared with ACR, it improves SR@0.6 by 7.4 points, reduces NE by 0.40~m, and raises SR@Manip and SR@E2E by 2.99 and 5.03 points, respectively. RLAIF therefore provides stronger navigation and handoff quality while eliminating the inference-time refinement agent: Appendix Table~\ref{tab:acr_rlaif_latency} shows that total inference time decreases from 50.33~s to 13.80~s, a 72.6\% reduction. Furthermore, using the trained \navmodel's actual stopping distribution to collect new rollouts and post-train \manipmodel further boosts end-to-end success from 30\% to 52\% in simulation and from 17\% to 30\% on the physical robot, completing the bidirectional co-evolution loop. Additional diagnostics are provided in Appendix~\ref{sec:endpoint_diagnostics}.

\begin{table}[t]
\caption{Training-pipeline ablation on \textsc{Nav-N-Manip}. The baseline uses RxR, HM3D ObjectNav, and R2R; Multi-Task Co-Tuning (MTC) adds Rendered EQA and Nav4Manip-Pairs. Rejection-Sampling Fine-Tuning (RFT) builds on MTC; Agentic Counterfactual Refinement (ACR; inference time, no training) and two-iteration RLAIF build on RFT. Alternating optimization performs two navigation--manipulation rounds with the same number of training samples per stage as RLAIF. Best/second-best results are bolded/underlined.}
\label{tab:training_ablation}
\centering
\small
\setlength{\tabcolsep}{3.5pt}
\resizebox{\textwidth}{!}{%
\begin{tabular}{lccccccc}
\toprule
Model variant & SR@3.0 $\uparrow$ & SR@1.5 $\uparrow$ & SR@1.0 $\uparrow$ & SR@0.6 $\uparrow$ & NE $\downarrow$ & SR@Manip $\uparrow$ & SR@E2E $\uparrow$\\
\midrule
Baseline & 56.2 & 28.7 & 19.6 & 10.7 & 3.54 & 6.01 & 4.07\\
$+$ MTC & 80.2 & 77.1 & 69.8 & 52.7 & 1.81 & 39.34 & 31.01\\
\quad $+$ RFT & 80.8 & 76.7 & 73.8 & 60.9 & 1.76 & 38.18 & 31.79\\
\qquad $+$ ACR (IT) & 82.4 & 78.5 & 76.6 & 64.3 & 1.55 & \underline{53.88} & \underline{47.10}\\
\qquad $+$ RLAIF (\method) & \textbf{89.5} & \textbf{84.1} & \textbf{80.4} & \textbf{71.7} & \textbf{1.15} & \textbf{56.87} & \textbf{52.13}\\
\midrule
Alternating Optimization & 80.6 & 78.3 & 76.4 & \underline{68.4} & 1.61 & 41.47 & 34.30\\
MTC + RLAIF & \underline{84.5} & \underline{80.0} & \underline{78.5} & 64.2 & \underline{1.46} & 51.16 & 43.02 \\
\bottomrule
\end{tabular}
}
% \vspace{-10pt}
\end{table}

%% file: section/06_conclusion.tex
\section{Conclusion}

We introduced \method, a bidirectionally coupled dexterous mobile-manipulation system. Our automated data-generation pipeline pairs long-horizon navigation trajectories with rollout-derived manipulation affordances. Multi-task co-tuning, verifier-guided RFT, and RLAIF from execution-grounded counterfactual preferences jointly replace coarse, one-shot endpoint prediction with manipulation-informed navigation learning. Experiments on our \textsc{Nav-N-Manip} benchmark demonstrate that \method enables navigation and manipulation to co-evolve through their feedback, improving the system's ability to solve real-world dexterous mobile-manipulation tasks.

\paragraph{Limitations.}
Our evaluation uses a single Galbot G3 robot with Sharpa hands, and our navigation data cover only indoor environments. Future work should assess transfer across different robot embodiments and outdoor settings. Our bidirectional co-evolution framework retains separate navigation and manipulation models; extending it to a unified model with a shared action space for long-horizon mobile manipulation is a promising direction.

% \paragraph{Limitations.}
% Our current evaluation is limited to a single Galbot G3 robot equipped with Sharpa hands. Further experiments across a broader range of robot platforms, including humanoid robots, are needed to assess the generality of the proposed framework. In addition, our navigation data are currently generated only in indoor environments. Future work should extend the framework to broader embodied settings, such as autonomous driving and outdoor delivery, and investigate transfer across manipulation policies, robot embodiments, and operating environments.

%% file: section/07_ethic.tex
\subsection*{AI use statement}

Generative AI tools were used to annotate our generated dataset, provide feedback for RLAIF, serve as agent baselines during evaluation, and assist with language polishing. The authors reviewed the AI-assisted content and take full responsibility for the final manuscript.

\subsection*{Ethics statement}

This work primarily uses generated simulation environments and does not rely on personal or human-subject data. Learned navigation and manipulation components may fail under distribution shift and should be deployed with collision checking, hardware limits, emergency stops, and appropriate human supervision. Any data release will document model-generated annotations, known limitations, and applicable asset licenses.

\subsubsection*{Reproducibility Statement}
The main paper specifies the task, training objectives, and refinement procedure. Appendices~\ref{app:video_interface}--\ref{app:training_implementation} detail the navigation architecture, agentic loop, and training configurations (Table~\ref{tab:training_implementation}); Appendices~\ref{app:data_architecture}--\ref{app:quality_control} specify the benchmark splits and data-generation pipeline; Appendix~\ref{app:real_world_deployment} (Section~\ref{app:navigation_deployment}) and Appendix~\ref{app:manipulation_deployment} (Sections~\ref{app:wam_manipulation_executor}--\ref{app:manipulation_rtc}) provide the real-world deployment and manipulation policy configurations; and Appendices~\ref{app:full_navigation_results}--\ref{app:system_ablations} report complete benchmark and ablation results.

%% file: section/11_taskdefinition.tex
\section{Dexterous Mobile-Manipulation Task Definition}
\label{app:task_definition}

A dexterous mobile-manipulation task is defined by a task instance $s$, a language instruction $l$, a navigation policy $\pi_\theta$, and a downstream manipulation policy $\pi_m$. In our system, $\pi_m$ is implemented by \manipmodel. For episode $i$, the navigator generates a trajectory $\tau_i^n=(o_{1:T_i}^n,a_{1:T_i}^n)$ that terminates at the base pose $b_i=(x_i,y_i,\theta_i)$. 
Unlike standard VLN, merely reaching a meter-scale goal region is insufficient: dexterous mobile manipulation requires a precise terminal pose, including a base position and approach yaw that allow $\pi_m$ to complete the manipulation task without collisions. \kzxp{Therefore, navigation must be optimized to meet the requirements of downstream manipulation while maintaining long-horizon navigation capabilities.}

\begin{figure*}[t]
    \centering
    \includegraphics[width=\textwidth]{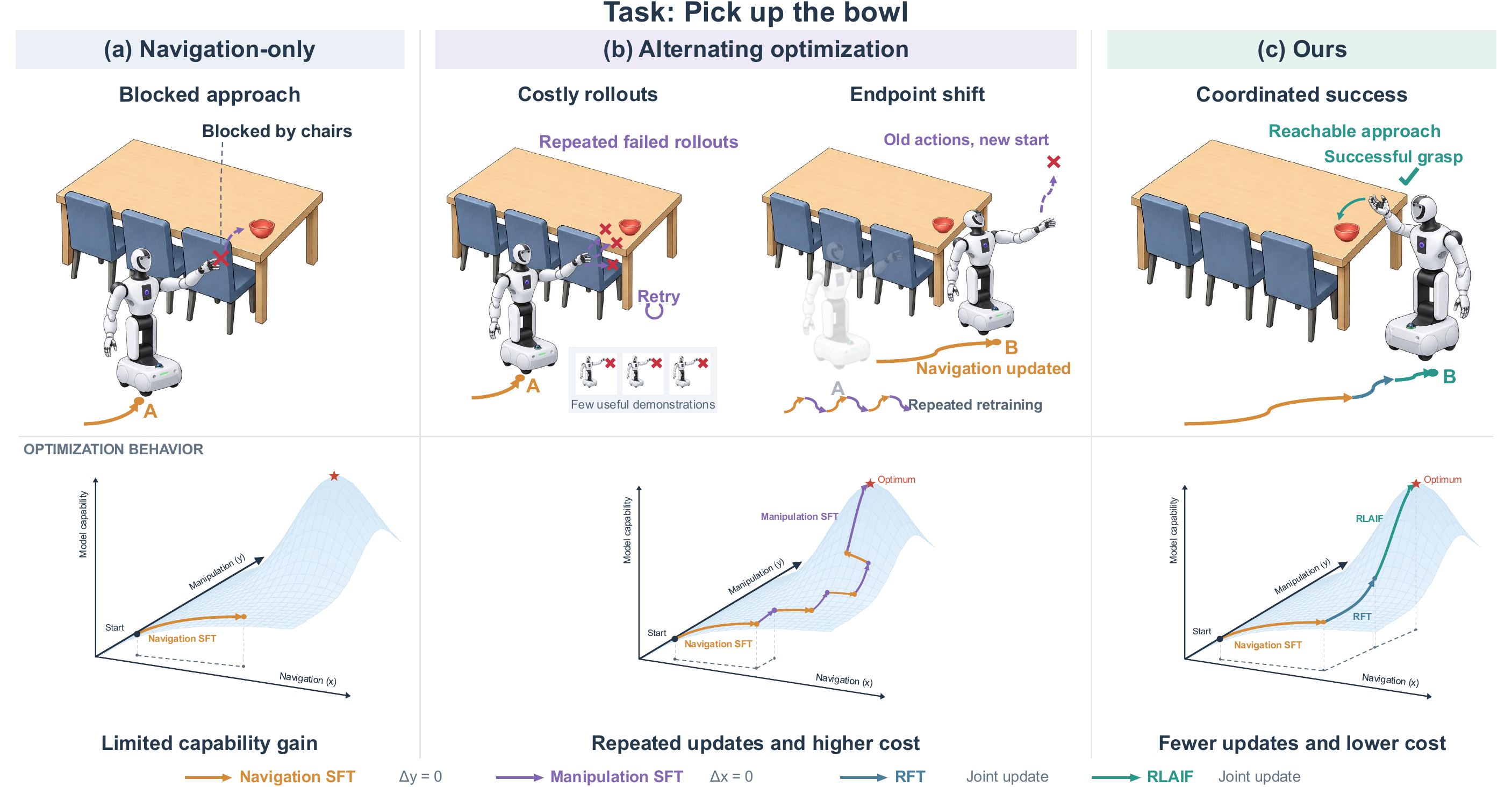}
    \caption{\textbf{Comparison of optimization strategies for mobile manipulation.} Navigation-only optimization stops at a chair-blocked endpoint. Alternating optimization wastes manipulation rollouts there and suffers action mismatch after navigation moves the endpoint from A to B. Our staged co-evolution instead selects a reachable approach and smoothly completes the grasp with fewer updates.}
    \label{fig:task_interface_example}
\end{figure*}

Figure~\ref{fig:task_interface_example} illustrates three optimization strategies for mobile manipulation through a bowl-picking task. Navigation-only training may reach endpoint A, which satisfies conventional navigation metrics but leaves the robot unable to grasp the bowl due to obstruction by the chairs. Optimizing navigation alone therefore fails to account for manipulation feasibility, limiting end-to-end task success.

A second strategy is coordinate-wise optimization, illustrated in the middle panels of Figure~\ref{fig:task_interface_example}, which alternates navigation updates along $x$ with manipulation updates along $y$. These panels expose two practical inefficiencies. First, manipulation rollouts at an unsuitable endpoint such as A repeatedly encounter the chair obstruction, making useful demonstrations difficult and costly to obtain. Second, after navigation moves the terminal pose from A to B, actions collected at A assume a different base pose, approach direction, and robot--object geometry. Reusing them at B creates an action mismatch. Our controlled comparison in Table~\ref{tab:training_ablation} evaluates two navigation--manipulation rounds, comprising four module-training stages.
With matched per-stage training sample counts, this baseline achieves lower end-to-end success than two-round RLAIF, which avoids repeated manipulation-policy training. Appendix~\ref{app:system_ablations} provides the experimental details.

Our staged co-evolution strategy, shown in the rightmost panel of Figure~\ref{fig:task_interface_example}, follows a shorter coupled path. Multi-task co-tuning establishes manipulation-aware navigation, allowing the navigator to favor endpoint B on the unobstructed side. We also train \manipmodel using diverse data collected from different navigation endpoints to achieve better manipulation results. RFT then uses downstream manipulation outcomes to improve navigation trajectories, while agentic counterfactual RLAIF distills execution-validated corrections into the navigator. Together, these stages enable effective navigation--manipulation co-evolution, guiding the robot toward manipulation-feasible endpoints and improving end-to-end task success. Unlike alternating optimization, our approach avoids repeatedly retraining the two modules.

%% file: section/12_systemarch.tex
\section{System Architecture}
\label{app:model_architecture}

\subsection{Multi-Scale Visual Encoding}
\label{app:video_interface}

Following the multi-scale visual-history encoding strategy of Uni-NaVid~\citep{zhang2025uninavid}, \navmodel preserves both fine-grained evidence at the current step and long-horizon context by organizing its visual observations into four temporal tiers: the current frame and recent-, middle-, and long-range history.  The current frame is retained at the highest spatial resolution, whereas progressively older observations are represented with progressively fewer visual tokens.  This hierarchy allocates most of the visual-token budget to information that is immediately relevant for control while maintaining a compact record of earlier scene layouts, landmarks, and interactions.  With a history stride of four, the default recent-, middle-, and long-range capacities are 8, 16, and 64 frames, respectively.  During training, we randomize the recent and middle capacities over $\{4,6,8,12\}$ and $\{8,12,16,24\}$, while keeping the long-range capacity at 64, thereby exposing the model to histories of varying duration.

All tiers are encoded by the shared Qwen3-VL vision encoder and multimodal spatial merger~\citep{bai2025qwen3}; no additional learned projector is introduced for history frames.  The vision tokenizer uses $16\!\times\!16$ patches with a spatial merge size of two.  Thus, the current $640\!\times\!480$ observation produces a $20\!\times\!15$ grid containing 300 merged visual tokens.  For an earlier observation $o_i$ assigned to tier $r\in\{\mathrm{recent},\mathrm{mid},\mathrm{long}\}$, we apply adaptive average pooling to the merged feature grid,
\[
v_i^{(r)}=\operatorname{Pool}_r\!\left(\operatorname{Enc}_{\mathrm{vis}}(o_i)\right),
\]
yielding grids of $5\!\times\!4=20$, $3\!\times\!2=6$, and $2\!\times\!2=4$ tokens for the recent, middle, and long tiers, respectively.  Letting $v_t^{(0)}$ denote the current-frame tokens, we concatenate the tiers from the oldest temporal scale to the current observation:
\begin{equation}
V_t=\left[\{v_i^{(\mathrm{long})}\};\ \{v_i^{(\mathrm{mid})}\};\
\{v_i^{(\mathrm{recent})}\};\ v_t^{(0)}\right].
\label{eq:multiscale_visual_sequence}
\end{equation}
The default visual-token budget is 1,024. During training, we randomly sample a budget between 384 and 1,024 for each batch. If the sequence exceeds the budget, older visual tokens are discarded first, while the current observation is always preserved. Validation and deployment use a fixed 1,024-token budget.

Temporal order is represented by textual timestamps, such as \texttt{<3.5 seconds>}, placed adjacent to each image block.  The visual M-RoPE temporal index remains zero because the observations are provided as individual images.  When odometry is available, relative motion $\Delta p=(\Delta x,\Delta y,\Delta\psi)$ is normalized by $(0.25\text{ m},0.25\text{ m},15^\circ)$, encoded using sine and cosine functions at frequencies $\{1,2,4,8\}$, and projected by a $24\!\rightarrow\!256\!\rightarrow\!4096$ MLP whose output is added to every token of the corresponding frame.  During training, we perturb each relative-motion estimate with zero-mean Gaussian noise whose standard deviations are $(0.02\text{ m},0.02\text{ m},0.017\text{ rad})$ for $(\Delta x,\Delta y,\Delta\psi)$, respectively.  When odometry is unavailable, we replace the motion encoding with a learned \texttt{no\_odom} embedding.  Recent and middle tiers additionally include serialized action history; the entire action-history side channel is dropped with probability 0.1 during training.

For a uniform interface across simulation, offline datasets, and real-robot deployment, we exchange observations as JPEG images.  Each image is decoded in BGR format, center-cropped to the expected $4{:}3$ view, resized using area interpolation, and converted to RGB before visual encoding.  The current, recent-, middle-, and long-range tiers use resolutions of $640\!\times\!480$, $320\!\times\!224$, $160\!\times\!128$, and $96\!\times\!96$, respectively.  To reduce sensitivity to a particular camera setup, the training data incorporate variations in camera height and pitch, illumination, and horizontal field of view.  Together with the history and token-budget randomization described above, these variations enable \navmodel to tolerate moderate changes in viewpoint, lighting conditions, and camera intrinsics in real-world applications.

\subsection{Detailed \navmodel Architecture}
\label{app:nav4manip_architecture}

We initialize \navmodel from Qwen3-VL-8B-Instruct~\citep{bai2025qwen3}.  The backbone comprises a vision transformer, a multimodal merger, and an autoregressive language model, with the configuration summarized in Table~\ref{tab:architecture_components}. During training, we jointly optimize the vision encoder, multimodal merger, and language model using full-parameter fine-tuning.

\begin{table}[t]
\caption{Backbone configuration of \navmodel.}
\label{tab:architecture_components}
\centering
\small
\setlength{\tabcolsep}{8pt}
\renewcommand{\arraystretch}{1.12}
\begin{tabular}{l l}
\toprule
\textbf{Component} & \textbf{Configuration} \\
\midrule
\multicolumn{2}{l}{\textit{Vision Encoder}} \\
\quad Layers / attention heads & 27 / 16 \\
\quad Hidden / intermediate dimensions & 1,152 / 4,304 \\
\quad Output dimension & 4,096 \\
\quad Patch / spatial merge / temporal patch & 16 / 2 / 2 \\
\quad Deep-stack layers & 8, 16, 24 \\
\addlinespace[2pt]
\multicolumn{2}{l}{\textit{Language Model}} \\
\quad Layers & 36 \\
\quad Hidden / intermediate dimensions & 4,096 / 12,288 \\
\quad Attention / KV heads & 32 / 8 \\
\quad Vocabulary size & 151,936 \\
\quad Maximum context length & 262,144 \\
\addlinespace[2pt]
\multicolumn{2}{l}{\textit{Positional Encoding}} \\
\quad M-RoPE sections & $[24,20,20]$ \\
\quad RoPE base & $5{,}000{,}000$ \\
\bottomrule
\end{tabular}
\end{table}

\paragraph{Learned queries and prediction heads.}
Inspired by the use of learnable queries to elicit task-specific visual and affordance representations in BLIP-2, AffordanceVLA, and Afford-VLA~\citep{li2023blip,yu2026affordancevla,wang2026afford}, we adapt the pretrained backbone by adding ten learned query tokens: one dense-affordance query, eight endpoint-candidate queries, and one action query.  Each query has a learnable embedding that is initialized as a model parameter and optimized end to end with the backbone and prediction heads.  The queries are registered as additional tokenizer entries, inserted immediately after the assistant prefix in map--candidate--action order, and excluded from the language-model token loss.  Their contextualized hidden states are passed to the corresponding prediction heads.

The hidden state of the dense-affordance query is decoded by a $4096\!\rightarrow\!4096\!\rightarrow\!8192$ GELU MLP and reshaped into the logit map $\hat M_t\in\mathbb R^{32\times32\times8}$.  This ego-centered grid covers $x,y\in[-3.2,3.2]$~m at a resolution of 0.2~m per cell, with the robot origin located at cell $(16,16)$.

All eight candidate queries share a $4096\!\rightarrow\!512\!\rightarrow\!512\!\rightarrow\!7$ MLP.  Candidate $k$ predicts $[\hat x,\hat y,\hat\theta,\hat u,\hat v,z^v,z^s]$. We map $(\hat x,\hat y)$ through $3.2\tanh(\cdot)$, $\hat\theta$ through $\pi\tanh(\cdot)$, and the normalized image coordinates $(\hat u,\hat v)$ through a sigmoid; $z^v$ and $z^s$ are the visibility and affordance-score logits, respectively. Candidate slots retain the ordering provided by Nav4Manip-Pairs. In addition, a goal/waypoint head operates on the fused multimodal state to predict two three-dimensional waypoints.

\paragraph{Multi-task co-tuning losses.}
For completeness, we expand all terms in the co-tuning objective from Eq.~\ref{eq:cotune_v2}:
\begin{equation}
\mathcal L_{\mathrm{co}}
=\mathcal L_{\mathrm{nav}}
+\lambda_{\mathrm{Afford}}\mathcal L_{\mathrm{Afford}}
+\lambda_{\mathrm{EQA}}\mathcal L_{\mathrm{EQA}}.
\label{eq:implemented_cotune_loss}
\end{equation}
\kzxp{Dataset-specific masks activate $\mathcal L_{\mathrm{nav}}$, $\mathcal L_{\mathrm{EQA}}$, and $\mathcal L_{\mathrm{Afford}}$ only when the corresponding supervision is available.}

The navigation term is the autoregressive negative log-likelihood of the target action-token sequence $a^n=(a_1^n,\ldots,a_{T_a}^n)$ under the complete multimodal context $h$:
\begin{equation}
\mathcal L_{\mathrm{nav}}
=-\mathbb E_{(h,a^n)}
\left[
\sum_{s=1}^{T_a}
\log \pi_\theta\!\left(a_s^n\mid h,a_{<s}^n\right)
\right].
\label{eq:implemented_navigation_loss}
\end{equation}
Only target action tokens contribute to this loss; prompt, observation, and learned-query positions are masked. The grounded EQA term has the same token-level form. For an answer sequence $e=(e_1,\ldots,e_{T_e})$,
\begin{equation}
\mathcal L_{\mathrm{EQA}}
=-\mathbb E_{(h,e)}
\left[
\sum_{s=1}^{T_e}
\log p_\theta\!\left(e_s\mid h,e_{<s}\right)
\right],
\label{eq:implemented_eqa_loss}
\end{equation}

The affordance objective supervises the dense map and discrete endpoint candidates with rollout-derived targets:
\begin{equation}
\mathcal L_{\mathrm{Afford}}
=\mathcal L_{\mathrm{dense}}
+\mathcal L_{\mathrm{cand}}
+0.5\mathcal L_{\mathrm{rank}}
+0.2\mathcal L_{\mathrm{cons}}.
\label{eq:implemented_affordance_loss}
\end{equation}
Let $n$ index samples in a mini-batch, $k\in\{1,\ldots,8\}$ index candidate slots, and $g$ index the $32\times32\times8$ position--yaw entries of the dense map.  For entry $(n,g)$, $z_{ng}$ is the predicted logit, $a_{ng}\in[0,1]$ is the affordance target, and $m_{ng}\in\{0,1\}$ indicates whether the entry is labeled.  With $\sigma$ denoting the sigmoid function, the masked dense-map loss is
\begin{align}
\ell_{\mathrm{BCE}}(z,y)
&=-y\log\sigma(z)-(1-y)\log(1-\sigma(z)),\label{eq:afford_bce}\\
\mathcal L_{\mathrm{dense}}
&=\frac{\sum_{n,g}m_{ng}\ell_{\mathrm{BCE}}(z_{ng},a_{ng})}
{\max(1,\sum_{n,g}m_{ng})}.
\label{eq:afford_dense_loss}
\end{align}
This binary-cross-entropy loss supports soft targets, while entries with $m_{ng}=0$ are ignored.

For candidate $(n,k)$, let $\mathbf p_{nk}=(x,y)$, $\theta_{nk}$, $\mathbf q_{nk}=(u,v)$, $v_{nk}\in\{0,1\}$, and $s_{nk}\in[0,1]$ be the target position, yaw, normalized image coordinates, visibility, and affordance score.  Their predictions are denoted by hats, while $z^v_{nk}$ and $z^s_{nk}$ are the visibility and score logits from the candidate head.  The mask $c_{nk}\in\{0,1\}$ indicates whether the candidate slot is valid.  The candidate loss is
\begin{align}
\mathcal L_{\mathrm{cand}}
&=\frac{1}{\max(1,\sum_{n,k}c_{nk})}
\sum_{n,k}c_{nk}\Big[
\lVert\hat{\mathbf p}_{nk}-\mathbf p_{nk}\rVert_1
+d_{\mathbb S^1}(\hat\theta_{nk},\theta_{nk})
+v_{nk}\lVert\hat{\mathbf q}_{nk}-\mathbf q_{nk}\rVert_1\notag\\[-2pt]
&\hspace{38mm}
+\ell_{\mathrm{BCE}}(z^v_{nk},v_{nk})
+\ell_{\mathrm{BCE}}(z^s_{nk},s_{nk})\Big].
\label{eq:afford_candidate_loss}
\end{align}
Thus, unlabeled map cells and invalid candidate slots do not contribute to the loss, and image-coordinate regression is applied only to visible candidates. Yaw is supervised by the shortest angular distance $d_{\mathbb S^1}(\hat\theta,\theta)=\left|\operatorname{atan2}\!\left(\sin(\hat\theta-\theta),\cos(\hat\theta-\theta)\right)\right|$.

For ranking, $\mathcal P$ contains valid candidate pairs $(n,i,j)$ from the same sample whose target scores satisfy $s_{ni}>s_{nj}+10^{-4}$.  We apply the margin loss directly to the pre-sigmoid score logits:
\begin{align}
\mathcal L_{\mathrm{rank}}
&=\frac{\sum_{(n,i,j)\in\mathcal P}
\max(0,0.1-z^s_{ni}+z^s_{nj})}
{\max(1,|\mathcal P|)}.
\label{eq:afford_ranking_loss}
\end{align}

For consistency, $\mathcal B_{\mathrm{cam}}$ denotes the mini-batch samples with camera parameters, $b\in\{1,\ldots,8\}$ indexes yaw bins, $Q$ maps a predicted ground position to its discrete BEV cell, and $\Pi_n$ projects that position to normalized image coordinates for sample $n$ and clips them to $[0,1]^2$.  The consistency loss is
\begin{align}
\mathcal L_{\mathrm{cons}}
&=\frac{1}{\max(1,\sum_{n\in\mathcal B_{\mathrm{cam}},k}c_{nk})}
\sum_{n\in\mathcal B_{\mathrm{cam}},k}c_{nk}\Big[
-\log\max\!\left(10^{-4},
\max_b\sigma\!\left(z_{n,Q(\hat{\mathbf p}_{nk}),b}\right)\right)\notag\\[-2pt]
&\hspace{42mm}
+\lVert\hat{\mathbf q}_{nk}-\Pi_n(\hat{\mathbf p}_{nk})\rVert_1\Big].
\label{eq:afford_consistency_loss}
\end{align}
This term encourages a candidate to lie on a high-response map cell while agreeing with its geometric image projection.  Following the implementation, it is applied to every valid candidate in camera-calibrated samples and is not masked by target visibility.

\paragraph{Action decoding.}
The action query conditions autoregressive generation of an action sequence.  The complete action interface contains nine commands: \texttt{forward}, \texttt{turn\_left}, \texttt{turn\_right}, \texttt{move\_left}, \texttt{move\_right}, \texttt{up}, \texttt{down}, \texttt{pause}, and \texttt{stop}.  During the initial multi-task co-tuning and RFT stages, explicit action supervision and constrained decoding cover the seven-action subset excluding \texttt{up} and \texttt{down}.

Agentic counterfactual RLAIF extends the action space with \texttt{up} and \texttt{down} to support adaptation to workspaces at different heights. Leveraging navigation and manipulation memories, the agent proposes corrections toward more manipulation-feasible navigation endpoints. \kzxp{During RLAIF training, we disable \texttt{up} and \texttt{down} to ensure that all training data use the same seven-action space as multi-task co-tuning and RFT, enabling a consistent comparison across training stages. At inference time, however, the agent can still invoke \texttt{up} and \texttt{down} to improve adaptability to real-world environments.}

\paragraph{Prompt format.}
For the Qwen3-VL-based \navmodel, we use the following fixed system prompt: ``You are a wheeled indoor navigation robot. Follow the instruction using the observation history and current image.''  The user message contains the task instruction and requests actions from the complete nine-action set described above.  Each observation is preceded by a textual timestamp; for observations in the recent and middle tiers, the actions executed after that observation are appended as additional context. The ten learned queries are inserted immediately after the assistant prefix, and the outer conversation markers follow the Qwen3-VL chat template.

\subsection{Agentic Navigation--Manipulation Loop}
\label{app:agenticwam_system}

\begin{figure*}[t]
    \centering
    \includegraphics[width=\textwidth,draft=false]{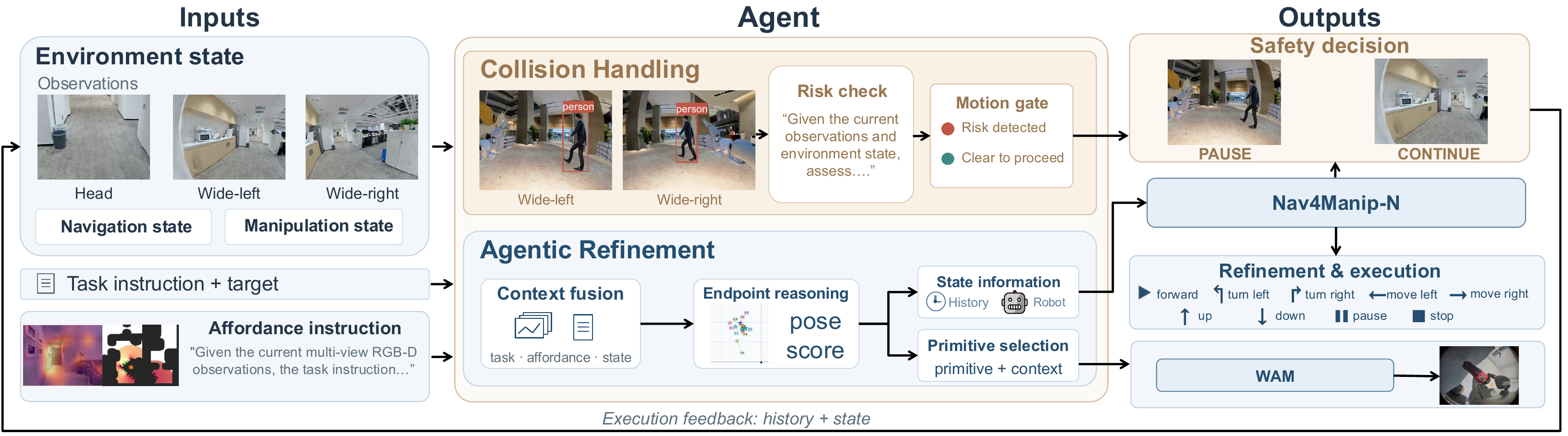}
    \caption{\textbf{Agent-driven counterfactual refinement in \method.} The coordinating agent integrates environment observations, robot state, task instructions, and endpoint affordances to refine or accept navigation endpoints. \kzxp{In parallel, Collision Handling uses detector feedback to issue a \texttt{PAUSE} decision when necessary.} The agent then selects a manipulation primitive and provides visual and textual context to \manipmodel. Execution outcomes update the task history and robot state, closing the feedback loop for subsequent refinement.}
    \label{fig:agentic_loop_appendix}
    \label{fig:agentic_loop}
\end{figure*}
\paragraph{Agent Loop and Prompt Interface.}
Figure~\ref{fig:agentic_loop} details how the agent $\mathcal C$ produces the counterfactual refinement $y_i^{+}=\mathcal C(x_i)$ used in Section~\ref{sec:rlaif_method}. For task $(l_i)$, the factual context $x_i$ contains the task instruction, factual navigation decision, execution outcome, and updated environment state. Within local refinement cycle $t$, the agent receives nearby manipulation-affordance predictions
\begin{equation}
\mathcal{A}_{i,t}=\{(b_{i,t,k},\hat q_{i,t,k})\}_{k=1}^{N_{i,t}},
\qquad
b_{i,t,k}=(p^x_{i,t,k},p^y_{i,t,k},\theta_{i,t,k}).
\label{eq:agent_affordance_candidates}
\end{equation}
Each pair contains a candidate base pose and its manipulation-success score. We denote the current base pose by $b_{i,t}$ and its predicted score by $\hat q_{i,t}=\hat q(b_{i,t})$. The affordance target $q$ predicted by $\hat q$ is defined in Appendix~\ref{app:data_affordance}, Eq.~\ref{eq:agenticwam_offline}. $\hat q_{i,t}$ at the current pose indicates whether the robot is ready for manipulation, while $\mathcal A_{i,t}$ provides alternative nearby poses for selecting a better navigation endpoint.

The local environment state is
\begin{equation}
S_{i,t}=(O_{i,t},s_{i,t}^{\mathrm{nav}},s_{i,t}^{\mathrm{manip}},H_{i,t}),
\qquad
O_{i,t}=I_{i,t}^{H},
\label{eq:agent_local_state}
\end{equation}
where $O_{i,t}$ contains only the current head-camera observation; $s_{i,t}^{\mathrm{nav}}$ is the navigation state; $s_{i,t}^{\mathrm{manip}}$ summarizes the arm, dexterous hand, and manipulation state; and $H_{i,t}$ contains the relevant factual interaction history. An external YOLO26n detector~\citep{jocher2026yolo26} processes the head-camera image and returns structured detections
\begin{equation}
D_{i,t}=g_{\mathrm{det}}(I_{i,t}^{H})
=\{(c_{i,t,j},B_{i,t,j},\rho_{i,t,j})\}_{j=1}^{M_{i,t}},
\label{eq:yolo_collision_detections}
\end{equation}
where $c_{i,t,j}$, $B_{i,t,j}$, and $\rho_{i,t,j}$ denote the class, bounding box, and confidence of detection $j$, respectively. The complete navigation action space is formatted over multiple lines for readability:
\begin{equation}
\begin{aligned}
\mathcal U=\{&\texttt{forward},\texttt{turn\_left},\texttt{turn\_right},\\
&\texttt{move\_left},\texttt{move\_right},\texttt{up},\\
&\texttt{down},\texttt{pause},\texttt{stop}\}.
\end{aligned}
\label{eq:agent_action_space}
\end{equation}

Given these inputs, Agentic Refinement proposes a local correction and, once the endpoint is accepted, provides the manipulation primitive and visual--textual context required by \manipmodel. Collision Handling supplies the YOLO26n results to the agent, which checks whether the proposed motion is safe to execute.

\textbf{Agentic Refinement.}
At each refinement step $t$, the agent $\mathcal C$ uses the task instruction, predicted affordances, and current environment state to propose a refinement action:
\begin{equation}
\tilde u_{i,t}=\mathcal C(l_i,\mathcal A_{i,t},S_{i,t}),
\qquad
\tilde u_{i,t}\in\mathcal U_{\mathrm{ref}},
\label{eq:agentic_refinement}
\end{equation}
where $\mathcal A_{i,t}$ and $S_{i,t}$ denote the predicted affordances and current environment state, respectively. The refinement action space excludes the \texttt{pause} action:
\begin{equation}
\mathcal U_{\mathrm{ref}}
=
\mathcal U\setminus\{\texttt{pause}\}.
\label{eq:agent_refinement_action_space}
\end{equation}

\textbf{Collision Handling.}
The agent receives the YOLO26n detections $D_{i,t}$ and analyzes their classes, locations, and confidence together with the proposed refinement action and current state:
\begin{equation}
c_{i,t}^{\mathrm{safe}}
=f_{\mathrm{col}}(D_{i,t},\tilde u_{i,t},S_{i,t}),
\qquad
c_{i,t}^{\mathrm{safe}}\in\{\texttt{PAUSE},\texttt{CONTINUE}\}.
\label{eq:agent_collision_decision}
\end{equation}
The safety decision produces the action sent to the robot controller,
\begin{equation}
u_{i,t}=
\begin{cases}
\texttt{pause}, & c_{i,t}^{\mathrm{safe}}=\texttt{PAUSE},\\
\tilde u_{i,t}, & c_{i,t}^{\mathrm{safe}}=\texttt{CONTINUE}.
\end{cases}
\label{eq:agent_safety_override}
\end{equation}

In each local cycle, the robot executes at most one bounded correction before acquiring a fresh state,
\begin{equation}
S_{i,t}
\xrightarrow{\,f_{\mathrm{ref}}\,}
\tilde u_{i,t}
\xrightarrow{\,f_{\mathrm{col}}(D_{i,t})\,}
u_{i,t}
\xrightarrow{\,\mathrm{execute}\,}
S_{i,t+1}.
\label{eq:agent_refinement_transition}
\end{equation}
The agent then reevaluates the current-pose manipulation affordance from the updated observations and robot state. This receding-horizon procedure conditions every correction on factual execution feedback. The existing \texttt{stop} action accepts the current endpoint and triggers the \manipmodel, so no additional acceptance token is required. The executed corrections and final accepted endpoint together define the refined navigation decision $y_i^{+}=\mathcal C(x_i)$.

At each local cycle, the prompt receives the following structured view of $x_i$:
\begin{equation}
(l_i,\mathcal A_{i,t},s_{i,t}^{\mathrm{nav}},
s_{i,t}^{\mathrm{manip}},O_{i,t},D_{i,t},H_{i,t})
\label{eq:agent_prompt_context}
\end{equation}
and returns one action from $\mathcal U_{\mathrm{ref}}$. Collision Handling may replace this proposal with \texttt{pause}; a replaced proposal is not recorded as an executed refinement target. Otherwise, the action is executed and the newly observed state is fed into the next cycle.

Once the endpoint is accepted, the agent combines the navigator representation, current affordance, EQA answer, and environment state to select a manipulation primitive and construct the visual--textual context for \manipmodel.
During counterfactual RLAIF collection, the refined navigation decision $y_i^+$ is evaluated in closed loop with the frozen \manipmodel $\mathcal{W}$ to obtain $r_i^+$.
The factual rollout supplies $r_i^{-}$. The pair $(x_i,y_i^{+})$ is retained only when $r_i^{+}-r_i^{-}\geq\tau_\Delta$, matching Eq.~\ref{eq:rlaif_filter}. Thus, bounded endpoint correction is the inner loop of $\mathcal C$, while same-initialization replay and execution-based filtering form the outer RLAIF loop.
\paragraph{Agent Prompt.}
The following fixed prompt is used by the code agent to perform counterfactual endpoint refinement during RLAIF data collection. It instructs the agent to interpret execution evidence, propose one bounded navigation correction at a time, prepare the \manipmodel handoff after endpoint acceptance, and return the resulting state to the refinement loop. We use GPT Astra with low reasoning effort as the code-agent backend.

\begin{promptbox}
You are the Agent for Nav4Manip, an embodied dexterous mobile-manipulation system. Your task is to coordinate navigation, endpoint refinement, and downstream manipulation using the task instruction, manipulation affordance, current head-camera observation, robot state, and interaction feedback.

## CURRENT TASK
{task_instruction}

## INPUT
- **Visual Observation**: One current RGB image from the head camera
- **Object Detections**: YOLO26n class labels, bounding boxes, and confidence scores for the current head-camera image
- **Affordance Information**: Current-pose manipulation affordance and nearby candidate base poses represented as `(x, y, yaw, affordance_score)`
- **Navigation State**: Current navigation status and navigation-related state
- **Manipulation State**: Current arm, dexterous hand, object-interaction, and manipulation-phase state
- **Navigator Context**: Current task-relevant outputs from the navigation model
- **Execution History**: Selected navigation and manipulation history from previous interaction cycles
- **Execution State**: Latest manipulation outcome and updated environment state, when available

## CORE RESPONSIBILITIES
1. **COLLISION HANDLING**: Analyze the YOLO26n detection results together with the proposed action and robot state; when they indicate a collision or person-safety risk, replace the proposed refinement action with `pause`
2. **ASSESS MANIPULATION READINESS**: Determine whether the current base configuration provides a suitable handoff for downstream manipulation using the task instruction, affordance, head-camera observation, and robot state
3. **REFINE ENDPOINT**: If local adjustment is required, select exactly one refinement action according to the current observation and state
4. **PREPARE NAV4MANIP-M HANDOFF**: When the endpoint is accepted, select a manipulation primitive and construct concise visual-textual context for Nav4Manip-M
5. **SELECT EXECUTION FEEDBACK**: After execution, retain navigation-relevant history and updated state, especially changes in visibility, reachability, object state, robot state, and approach feasibility
6. **CLOSED-LOOP UPDATE**: Make each refinement decision from the latest observation and state; after an executed correction, wait for fresh observations before making the next decision

## OUTPUT FORMAT

### Endpoint Refinement
Return exactly one action from `forward`, `turn_left`, `turn_right`, `move_left`, `move_right`, `up`, `down`, or `stop`. Use `stop` to accept the current endpoint and trigger the Nav4Manip-M handoff. The independent collision handler may replace the selected action with `pause` before execution.

### Nav4Manip-M Handoff
{
  "primitive": "<one manipulation primitive from the provided primitive set>",
  "context": "<concise task-relevant visual-textual context for Nav4Manip-M>"
}

### Navigation Feedback
{
  "history": "<task-relevant execution history>",
  "state": "<updated environment and robot state relevant to navigation>"
}

## DECISION RULES
1. **TASK CONDITIONING**: Judge endpoint quality according to downstream manipulation executability rather than target proximity alone
2. **AFFORDANCE AWARENESS**: Use the current-pose and nearby candidate affordance predictions when evaluating the handoff
3. **MINIMAL CORRECTION**: Prefer the smallest bounded correction sufficient to improve the current endpoint
4. **FRESH OBSERVATION**: Reassess the endpoint after every physical correction using newly observed state
5. **EVIDENCE BASED**: Do not infer visibility, reachability, object motion, or task success without support from the provided inputs
6. **MODULE SEPARATION**: Do not replace the navigation policy or generate low-level manipulation actions; navigation is handled by the navigator and manipulation actions are produced by Nav4Manip-M
7. **SAFETY SEPARATION**: Collision and person-safety gating is handled by an independent safety mechanism
\end{promptbox}

\subsection{Implementation Details}
\label{app:training_implementation}

We initialize \navmodel from Qwen3-VL-8B-Instruct and jointly optimize the vision encoder, multimodal merger, language model, learned queries, and prediction heads. Training uses bfloat16 precision and SDPA attention. The ten special-token embeddings are created through the standard tokenizer-resize path, and the dense and candidate heads use the default initialization of PyTorch linear layers. Samples are dynamically padded within each batch; the context ceiling of the backbone is 262,144-token.

Co-tuning uses the 4.9M-sample mixture in Table~\ref{tab:data_v2}. We set $\lambda_{\mathrm{Afford}}=\lambda_{\mathrm{EQA}}=1$ in Eq.~\ref{eq:cotune_v2}; within $\mathcal L_{\mathrm{Afford}}$, the dense, candidate, ranking, and consistency weights are $(1,1,0.5,0.2)$. All stages use only monocular RGB observations from the head camera. Table~\ref{tab:training_implementation} summarizes the remaining settings.

For RFT, we instantiate the path term in Eq.~\ref{eq:rft_score_v2} as normalized clipped excess length,
\begin{equation}
\bar\ell_i=\operatorname{clip}\!\left(
\frac{L_i-L_i^*}{\max(L_i^*,1\,\mathrm{m})},0,1\right),
\end{equation}
so $\lambda_{\mathrm{path}}=0.10$ contributes at most 0.10 to the ranking score. For both RFT and RLAIF, manipulation scores are success rates over 16 rollout executions. RLAIF evaluates the factual and refined endpoints using matched perturbation seeds and retains a pair when $r_i^+-r_i^-\geq0.25$.

\begin{table*}[t]
\caption{Training and rollout configuration of \navmodel.}
\label{tab:training_implementation}
\centering
\small
\setlength{\tabcolsep}{6pt}
\renewcommand{\arraystretch}{1.08}
\begin{tabular}{p{0.24\textwidth}p{0.70\textwidth}}
\toprule
\textbf{Setting} & \textbf{Value} \\
\midrule
\multicolumn{2}{l}{\textit{Model and runtime}}\\
Backbone & Qwen3-VL-8B-Instruct \\
Trainable components & vision encoder, merger, language model, queries, prediction heads \\
Initialization & 10 special-token embeddings via tokenizer resize; heads via PyTorch defaults \\
Attention / memory & SDPA; gradient checkpointing \\
Distributed training & replicated full model; 512~MiB chunked all-reduce \\
Sequence length & dynamic padding; 262,144-token context ceiling \\
Visual input & head-camera RGB only (all stages) \\
\addlinespace[2pt]
\multicolumn{2}{l}{\textit{Co-tuning}}\\
Hardware & $32\times$ RTX PRO 5000 Blackwell GPUs (72~GB each) \\
Precision / optimizer & bfloat16 / Adafactor \\
Batch size & $1$ per GPU $\times32$ GPUs $\times32$ accumulation steps $=1{,}024$ \\
Learning rate & $1.3\!\times\!10^{-5}$ (backbone); $1.0\!\times\!10^{-4}$ (prediction heads) \\
Schedule & 160-step warmup; cosine decay to $0.1\eta_{\max}$ \\
Regularization & weight decay $=0$; gradient clipping $=1$ \\
Training duration & 2 epochs \\
Data / input & 4.9M samples (Table~\ref{tab:data_v2}); head-camera RGB only \\
Task loss weights & $\lambda_{\mathrm{Afford}}=\lambda_{\mathrm{EQA}}=1$ \\
Affordance loss weights & dense / candidate / ranking / consistency: $1/1/0.5/0.2$ \\
\addlinespace[2pt]
\multicolumn{2}{l}{\textit{RFT}}\\
Rollout collection & 500 tasks; 8 intermediate states per trajectory \\
Manipulation verification & 16 trials per endpoint \\
Path penalty & $\lambda_{\mathrm{path}}=0.10$; denominator floor: $1$~m; clipping range: $[0,1]$ \\
Optimization & batch size $32$; learning rates $5\!\times\!10^{-6}$ / $1.0\!\times\!10^{-4}$; 1 epoch \\
\addlinespace[2pt]
\multicolumn{2}{l}{\textit{RLAIF}}\\
Collection & $K_{\mathrm{iter}}=2$; 500 tasks per round; GPT Astra agent (low reasoning effort) \\
Manipulation verification & 16 matched trials per endpoint; $\tau_\Delta=0.25$ \\
Optimization & batch size $32$; learning rates $5\!\times\!10^{-6}$ / $1.0\!\times\!10^{-4}$; 1 epoch per round \\
\bottomrule
\end{tabular}
\end{table*}

%% file: section/13_ablation.tex
\section{Additional Experiments}
\label{sec:endpoint_diagnostics}

This section supplements the main evaluation with complete navigation results and more detailed analyses of the proposed system. We first report benchmark-wide comparisons, then isolate the effects of navigation-model architecture and co-tuning data, and finally examine the agentic refinement and RLAIF pipeline through system-level accuracy, latency, and iteration studies.

\subsection{Full Navigation Results}
\label{app:full_navigation_results}
Tables~\ref{tab:main} and~\ref{tab:vlnverse} report the complete numerical results corresponding to Figure~\ref{fig:navigation_results}, covering R2R~\citep{anderson2018r2r} and RxR~\citep{ku2020rxr} val-unseen as well as the fine- and coarse-grained tasks on the official VLNVerse test split~\citep{lin2025vlnverse}.
Success Rate (SR@3) is the fraction of episodes whose final position lies within 3~m of the goal, while Oracle Success Rate (OSR@3) counts an episode if any visited position enters that radius. Success weighted by Path Length (SPL@3) combines SR@3 with a penalty for paths longer than the shortest-path reference. Navigation Error (NE) is the final shortest-path distance to the goal in meters, and Trajectory Length (TL) is the executed path length in meters. Higher values are better for SR, OSR, and SPL, whereas lower values are better for NE; TL is reported as a diagnostic without a universal preferred direction. For metrics with an indicated direction, the best and second-best results within each benchmark or task are shown in bold and underlined, respectively; Table~\ref{tab:main} additionally double-underlines third-best results.
The R2R and RxR comparisons include NaVid~\citep{zhang2024navid}, Uni-NaVid~\citep{zhang2025uninavid}, LookStep~\citep{yu2026lookstep}, NaVILA~\citep{cheng2025navila}, StreamVLN~\citep{wei2025streamvln}, InternVLA-N1~\citep{wei2025internvla}, PROSPECT~\citep{fan2026prospect}, MapDream~\citep{lian2026mapdream}, NavFoM~\citep{zhang2025navfom}, and Qwen-RobotNav~\citep{zhang2026qwenrobotnav}.
For VLNVerse, we compare against Uni-NaVid, InternVLA-N1, NavFoM, and Qwen-RobotNav. We also include the test-split Seq2Seq, CMA, HNR, RDP, and GAMA baselines reported in the original VLNVerse study~\citep{sutskever2014seq2seq,wang2019rcm,wang2024hnr,wang2025vlnpe,lin2025vlnverse}. We retain each method's stated single- or multi-view input configuration, as indicated in the tables, and evaluate \navmodel under the single-view setting used throughout the paper.

\begin{table}[t]
\caption{Closed-loop navigation on R2R and RxR val-unseen. SR, OSR, and SPL are percentages; NE is in meters. Best, second-best, and third-best results per benchmark are bolded, underlined, and double-underlined, respectively.}
\label{tab:main}
\centering
\small
\setlength{\tabcolsep}{5pt}
\resizebox{\ifdim\width>\linewidth\linewidth\else\width\fi}{!}{%
\begin{tabular}{llccccccc}
\toprule
& & \multicolumn{4}{c}{R2R} & \multicolumn{3}{c}{RxR}\\
\cmidrule(lr){3-6}\cmidrule(l){7-9}
Model & Views & SR@3 $\uparrow$ & SPL@3 $\uparrow$ & OSR@3 $\uparrow$ & NE $\downarrow$ & SR@3 $\uparrow$ & SPL@3 $\uparrow$ & NE $\downarrow$\\
\midrule
NaVid & single & 37.40 & 35.90 & 49.10 & 5.47 & 34.50 & 23.80 & 8.41\\
Uni-NaVid & single & 47.00 & 42.70 & 53.30 & 5.58 & 48.70 & 40.90 & 6.24\\
LookStep & single & 49.70 & 45.30 & 55.90 & 5.34 & 46.90 & 39.90 & 6.89\\
NaVILA & single & 54.00 & 49.00 & 62.50 & 5.22 & 49.30 & 44.00 & 6.77\\
StreamVLN & single & 56.90 & 51.90 & 64.20 & 4.98 & 52.90 & 46.00 & 6.22\\
InternVLA-N1 & single & 58.20 & 54.00 & 63.30 & 4.83 & 53.50 & 46.10 & 5.91\\
PROSPECT & single & 58.90 & 54.00 & \uuline{65.20} & 4.92 & 54.60 & 46.20 & 5.70\\
MapDream & single & 59.80 & 54.40 & 64.40 & \underline{4.59} & 59.40 & 49.20 & 4.96\\
NavFoM & single & 56.20 & 51.20 & 64.90 & 5.01 & 57.40 & 49.40 & 5.51\\
\midrule
NavFoM & multi & \underline{61.70} & \uuline{55.30} & \underline{72.10} & \uuline{4.61} & \underline{64.40} & \underline{56.20} & \underline{4.74}\\
Qwen-RobotNav & multi & \textbf{65.70} & \textbf{59.60} & \textbf{72.70} & \textbf{4.36} & \textbf{73.40} & \textbf{63.50} & \textbf{4.16}\\
\midrule
\navmodel & single & \uuline{60.03} & \underline{56.79} & 64.11 & 4.80 & \uuline{62.06} & \uuline{52.60} & \uuline{4.76}\\
\bottomrule
\end{tabular}
}
\end{table}

\begin{table}[t]
\caption{Closed-loop navigation on the official VLNVerse test split. SR, OSR, and SPL are percentages; NE and TL are in meters. Dashes denote metrics not reported for the test split. Best/second-best results per task are bolded/underlined; TL is not ranked.}
\label{tab:vlnverse}
\centering
\small
\setlength{\tabcolsep}{4.5pt}
\resizebox{\ifdim\width>\linewidth\linewidth\else\width\fi}{!}{%
\begin{tabular}{llccccc ccccc}
\toprule
& & \multicolumn{5}{c}{Fine-grained test} & \multicolumn{5}{c}{Coarse-grained test}\\
\cmidrule(lr){3-7}\cmidrule(l){8-12}
Model & Views & SR@3 $\uparrow$ & SPL@3 $\uparrow$ & OSR@3 $\uparrow$ & NE $\downarrow$ & TL & SR@3 $\uparrow$ & SPL@3 $\uparrow$ & OSR@3 $\uparrow$ & NE $\downarrow$ & TL\\
\midrule
Seq2Seq & single & 30.19 & 28.15 & -- & 4.79 & 5.63 & 27.17 & 25.53 & -- & 5.00 & 6.32\\
CMA & single & 28.83 & 25.54 & -- & 5.24 & 8.53 & 31.85 & 28.67 & -- & 4.72 & 7.53\\
HNR & single & 32.24 & 29.03 & -- & 4.93 & 7.98 & 34.59 & 30.45 & -- & 4.62 & 7.95\\
RDP & single & 36.38 & 32.29 & -- & 4.60 & 8.18 & 36.60 & 32.21 & -- & 4.65 & 8.35\\
GAMA & single & 37.72 & 33.85 & -- & 4.52 & 5.53 & 37.20 & 31.02 & -- & 4.23 & 3.58\\
Uni-NaVid & single & 45.74 & 26.91 & 67.02 & 4.11 & 12.35 & 29.68 & 13.47 & 42.70 & 5.05 & 10.33\\
InternVLA-N1 & single & 28.95 & 25.00 & 38.69 & 5.68 & 9.23 & 17.51 & 16.54 & 23.32 & 5.71 & 4.00\\
NavFoM & multi & 51.59 & 32.40 & 69.68 & 4.13 & 8.58 & 38.02 & 23.15 & 45.93 & 4.93 & 6.52\\
Qwen-RobotNav & multi & \underline{63.75} & \underline{57.93} & \underline{72.81} & \underline{3.00} & 8.16 & \underline{46.59} & \underline{41.54} & \underline{55.97} & \underline{4.08} & 7.68\\
\midrule
\navmodel & single & \textbf{76.08} & \textbf{71.59} & \textbf{82.94} & \textbf{2.20} & 8.29 & \textbf{63.17} & \textbf{58.49} & \textbf{73.66} & \textbf{3.05} & 8.14\\
\bottomrule
\end{tabular}}
\end{table}

On the VLNVerse test split, \navmodel achieves 76.08/71.59 SR@3/SPL@3 on the fine-grained task and 63.17/58.49 on the coarse-grained task. Relative to the strongest prior method in Table~\ref{tab:vlnverse}, Qwen-RobotNav, these results improve SR@3/SPL@3 by 12.33/13.66 points for fine-grained navigation and 16.58/16.95 points for coarse-grained navigation. Among the baselines originally reported with VLNVerse, GAMA gives the strongest fine-grained test result (37.72/33.85), while GAMA and RDP provide the two strongest coarse-grained SR/SPL trade-offs.

\subsection{VLN Model Ablations}
\label{app:architecture_ablations}

We evaluate three groups of factors affecting the navigation model: architectural components, the co-tuning data composition, and additional fine-tuning with real-to-sim scenes. For the backbone comparison, we replace Qwen3-VL-8B with the smaller Qwen3-VL-4B model~\citep{bai2025qwen3}.

\begin{table}[H]
\caption{Architecture ablations on \textsc{Nav-N-Manip}. SR@3 and SR@0.6 are navigation success rates at the indicated radii, in percent; NE is in meters. SR@Manip is the single-attempt manipulation success rate, while SR@E2E measures joint navigation and single-attempt manipulation success. Both follow Section 4.1 and are reported as percentages. The full model uses two RLAIF iterations. Best/second-best results are bolded/underlined.}
\label{tab:architecture_ablation}
\centering
\small
\setlength{\tabcolsep}{3.5pt}
\resizebox{\ifdim\width>\linewidth\linewidth\else\width\fi}{!}{%
\begin{tabular}{lccccc}
\toprule
Model variant & SR@3 $\uparrow$ & SR@0.6 $\uparrow$ & NE $\downarrow$ & SR@Manip $\uparrow$ & SR@E2E $\uparrow$\\
\midrule
w/o multi-scale history & 71.5 & 39.9 & 2.41 & 24.03 & 18.41\\
w/o affordance queries and losses & \underline{85.5} & 67.4 & \underline{1.43} & 41.28 & \underline{31.78}\\
Alternative backbone: Qwen3-VL-4B & 82.0 & \underline{68.2} & 1.61 & \underline{41.84} & 28.57\\
Ours & \textbf{89.5} & \textbf{71.7} & \textbf{1.15} & \textbf{56.87} & \textbf{52.13}\\
\bottomrule
\end{tabular}
}
\end{table}

\paragraph{Architectural Components.}
Table~\ref{tab:architecture_ablation} shows that removing multi-scale history lowers SR@0.6 from 71.7\% to 39.9\% and SR@E2E from 52.13\% to 18.41\%, consistent with temporal context supporting precise arrival and task completion. The smaller Qwen3-VL-4B backbone remains competitive in precise stopping but trails the full model in end-to-end success (28.57\% vs. 52.13\%). Removing affordance queries and losses lowers SR@3 from 89.5\% to 85.5\%, SR@0.6 to 67.4\%, and SR@E2E to 31.78\%. These results associate affordance supervision with improvements in both navigation accuracy and downstream success.

\begin{table}[H]
\caption{Ablation of co-tuning data sources for the \textsc{Nav4Manip-N} navigation module. The architecture is kept fixed, and the comparison focuses on the effects of training-data composition on \textsc{Nav4Manip-N}. Each ablated variant removes one source from the full training mixture; sampling ratios are not separately varied. VLNVerse results use the official test split. Success metrics are percentages; NE is in meters. Best/second-best results per task are bolded/underlined.}
\label{tab:data_ratio_ablation}
\centering
\scriptsize
\setlength{\tabcolsep}{3.5pt}
\resizebox{\ifdim\width>\linewidth\linewidth\else\width\fi}{!}{%
\begin{tabular}{lccccccccccc}
\toprule
& \multicolumn{3}{c}{R2R} & \multicolumn{3}{c}{VLNVerse (fine test)} & \multicolumn{3}{c}{VLNVerse (coarse test)} & \multicolumn{2}{c}{\textsc{Nav-N-Manip}}\\
\cmidrule(lr){2-4}\cmidrule(lr){5-7}\cmidrule(lr){8-10}\cmidrule(l){11-12}
Model variant & SR@3 $\uparrow$ & SPL@3 $\uparrow$ & NE $\downarrow$ & SR@3 $\uparrow$ & SPL@3 $\uparrow$ & NE $\downarrow$ & SR@3 $\uparrow$ & SPL@3 $\uparrow$ & NE $\downarrow$ & SR@Manip $\uparrow$ & SR@E2E $\uparrow$\\
\midrule
w/o Nav4Manip-Pairs & \underline{49.8} & 47.1 & \underline{5.69} & \underline{71.7} & \underline{70.6} & \underline{2.52} & 58.1 & 56.3 & 3.32 & 7.56 & 2.89\\
w/o Matterport 3D data & 17.4 & 16.3 & 9.11 & 69.4 & 67.5 & 2.66 & 57.7 & 56.0 & 3.31 & \underline{37.40} & \underline{27.33}\\
w/o EQA & \underline{49.8} & \underline{47.3} & 5.89 & 70.7 & 69.3 & 2.57 & \underline{59.6} & \underline{57.4} & \underline{3.26} & 36.43 & 26.55\\
\midrule
\navmodel & \textbf{60.0} & \textbf{56.8} & \textbf{4.80} & \textbf{76.1} & \textbf{71.6} & \textbf{2.20} & \textbf{63.2} & \textbf{58.5} & \textbf{3.05} & \textbf{38.18} & \textbf{31.79}\\
\bottomrule
\end{tabular}}
\end{table}

\paragraph{Training Data Sources.}
Table~\ref{tab:data_ratio_ablation} shows different effects across tasks. Removing Matterport 3D data most strongly affects R2R navigation: SR@3 falls from 60.0\% to 17.4\%, while NE rises from 4.80~m to 9.11~m. Removing Nav4Manip-Pairs most strongly affects the manipulation handoff, lowering SR@Manip/SR@E2E from 38.18\%/31.79\% to 7.56\%/2.89\%. Removing EQA also lowers performance, including SR@E2E to 26.55\%. Thus, component effects depend on evaluation task: Matterport removal most affects R2R, while Nav4Manip-Pairs removal is most consequential for manipulation and end-to-end success. The leave-one-source-out results do not identify an optimal mixture ratio.

\begin{table}[H]
\caption{Effect of fine-tuning on real-to-sim environments.}
\label{tab:real2sim_ablation}
\centering
\small
\setlength{\tabcolsep}{5pt}
\resizebox{\ifdim\width>\linewidth\linewidth\else\width\fi}{!}{%
\begin{tabular}{lcccc}
\toprule
Setting & SR@3 $\uparrow$ & SPL@3 $\uparrow$ & OSR@3 $\uparrow$ & SR@Manip $\uparrow$\\
\midrule
w/o real-to-sim fine-tuning & 19.17 & 19.14 & 30.83 & 0.83\\
with real-to-sim fine-tuning & \textbf{72.50} & \textbf{72.10} & \textbf{82.50} & \textbf{55.83}\\
\bottomrule
\end{tabular}
}
\end{table}

\paragraph{Real-to-Sim Fine-Tuning.}
Table~\ref{tab:real2sim_ablation} evaluates the effect of additional fine-tuning on real-to-sim environments. SR@3 rises from 19.17\% to 72.50\%, and SR@Manip from 0.83\% to 55.83\%, accompanied by improvements in SPL@3 and OSR@3. These gains show that adaptation to the evaluated environments improves both navigation reliability and the suitability of the resulting endpoints for manipulation.

\subsubsection{Diagnostic Analyses}
\label{app:closed_loop_diagnostics}

We further analyze the key factors affecting navigation performance through a series of diagnostic experiments.

\begin{table}[H]
\caption{Navigation diagnostics on R2R val-unseen. Results are reported as SR@3 (\%). Best results within each group are highlighted in bold.}
\label{tab:closed_loop_factors}
\centering
\small
\setlength{\tabcolsep}{3pt}
\begin{tabular}{lc@{\hspace{8pt}}lc@{\hspace{8pt}}lc@{\hspace{8pt}}lc}
\toprule
\multicolumn{2}{c}{\textbf{Instruction}} &
\multicolumn{2}{c}{\textbf{Initial Rotation}} &
\multicolumn{2}{c}{\textbf{History Length}} &
\multicolumn{2}{c}{\textbf{Stopping Rule}} \\
\cmidrule(lr){1-2}
\cmidrule(lr){3-4}
\cmidrule(lr){5-6}
\cmidrule(lr){7-8}
Setting & SR@3 &
Setting & SR@3 &
Setting & SR@3 &
Setting & SR@3 \\
\midrule
Matched & \textbf{60.03} &
$\theta<30^\circ$ & \textbf{73.00} &
88 frames & \textbf{60.03} &
$p_{\mathrm{stop}}\geq0.01$ & 60.03 \\
Mismatched & 19.36 &
$30^\circ\leq\theta<90^\circ$ & 58.39 &
5 frames & 49.16 &
Argmax & \textbf{60.20} \\
-- & -- &
$\theta\geq90^\circ$ & 24.11 &
-- & -- &
-- & -- \\
\bottomrule
\end{tabular}
\end{table}

\kzxp{Mismatched instructions reduce SR@3 by 40.67 points, highlighting the importance of language grounding. Navigation success also decreases with larger initial rotations, from 73.00\% for $\theta<30^\circ$ to 24.11\% for $\theta\geq90^\circ$. Reducing visual history from 88 to 5 frames lowers SR@3 by 10.87 points, demonstrating the benefit of longer temporal context. In contrast, changing the stopping rule from argmax to $p_{\mathrm{stop}}\geq0.01$ has little effect on SR@3 (60.20\% vs. 60.03\%). Together, these results highlight instruction grounding, initial orientation, and temporal context as important factors in navigation performance.}

\subsection{System Ablation Study}
\label{app:system_ablations}

\paragraph{Alternating optimization.}
The alternating-optimization baseline in Table~\ref{tab:training_ablation} performs two rounds of sequential navigation and manipulation training. Each round first optimizes the navigation policy and then the manipulation policy, yielding the sequence $N_1 \rightarrow M_1 \rightarrow N_2 \rightarrow M_2$ and four module-training stages in total. Each stage uses the same number of training samples as an RLAIF training stage. We evaluate the navigation and manipulation policies obtained after the second round. In comparison, two-round RLAIF performs two navigation-policy training stages while keeping the manipulation policy fixed. Alternating optimization achieves 68.4\% SR@0.6, 41.47\% SR@Manip, and 34.30\% SR@E2E, compared with 71.7\%, 56.87\%, and 52.13\%, respectively, for RLAIF. Thus, RLAIF achieves higher task success with fewer module-training stages under matched per-stage sample counts.

\paragraph{Agent-loop comparison in simulation.}
Table~\ref{tab:agent_loop_ablation} compares four refinement-agent backends across the same ten simulated episodes with the surrounding \method loop fixed. GPT-6 Astra achieves the highest SR@0.6 of 70\% and the lowest mean navigation latency of 61s per episode. GPT-5.6 Terra achieves the lowest NE of 2.22~m and the highest manipulation success rate of 40\%, while Terra and Sol both achieve an SR@0.6 of 60\%. Notably, the GPT models serve only as counterfactual-refinement agents in this experiment, unlike the direct-control baseline in Table~\ref{tab:downstream_agent_comparison}.

\begin{table}[H]
\caption{Refinement-agent comparison within the fixed \method loop in simulation, using ten episodes per backend. NE in meters, and mean navigation latency per episode in seconds. Best/second-best results are bolded/underlined.}
\label{tab:agent_loop_ablation}
\centering
\small
\setlength{\tabcolsep}{5pt}
\resizebox{\ifdim\width>\linewidth\linewidth\else\width\fi}{!}{%
\begin{tabular}{lcccc}
\toprule
Agent & SR@0.6 $\uparrow$ & NE $\downarrow$ & SR@Manip $\uparrow$ & Mean nav. latency (s) $\downarrow$\\
\midrule
GPT-5.6 Luna & 5/10 & 2.39 & 2/10 & 78\\
GPT-5.6 Terra & \underline{6/10} & \textbf{2.22} & \textbf{4/10} & \underline{70}\\
GPT-5.6 Sol & \underline{6/10} & 2.29 & \underline{3/10} & 73\\
GPT-6 Astra & \textbf{7/10} & \underline{2.25} & 2/10 & \textbf{61}\\
\bottomrule
\end{tabular}
}
\end{table}

\paragraph{Direct navigation agents.}
We evaluate GPT-6 Astra, GPT-5.6 Sol, GPT-5.6 Terra, and GPT-5.6 Luna as direct closed-loop navigation agents on ten \textsc{Nav-N-Manip} episodes, with results reported in Table~\ref{tab:downstream_agent_comparison}. All four models share identical natural-language instructions, head-camera RGB observations, collision feedback, action spaces, and action budgets, differing only in the underlying model backend. At each decision step, the agent predicts a sequence of one to six atomic navigation actions, which are executed sequentially before the next observation is returned. The agents have no access to global maps, robot poses, goal coordinates, reference trajectories, affordance annotations, or online evaluation metrics. The \textit{Shared prompt for direct navigation} box presents the complete prompt used across all four baselines, specifying the observation interface, action definitions, action budget, termination criteria, and required output format.

\begin{table}[H]
\caption{Direct-agent comparison on ten \textsc{Nav-N-Manip} episodes.}
\label{tab:downstream_agent_comparison}
\centering
\small
\setlength{\tabcolsep}{4pt}
\resizebox{\ifdim\width>\linewidth\linewidth\else\width\fi}{!}{%
\begin{tabular}{lccccccc}
\toprule
Method & SR@3.0 $\uparrow$ & SR@1.5 $\uparrow$ & SR@1.0 $\uparrow$ & SR@0.6 $\uparrow$ & NE $\downarrow$ & SR@Manip $\uparrow$ & Mean nav. latency (s) $\downarrow$\\
\midrule
GPT-5.6 Luna & 1/10 & 0/10 & 0/10 & 0/10 & 7.021 & 0/10 & \underline{123}\\
GPT-5.6 Terra & 6/10 & 4/10 & 0/10 & 0/10 & 3.011 & 0/10 & 848\\
GPT-5.6 Sol & 7/10 & 5/10 & 1/10 & 0/10 & 2.281 & 0/10 & 614\\
GPT-6 Astra & \textbf{10/10} & \textbf{10/10} & \textbf{9/10} & \underline{6/10} & \textbf{0.573} & \underline{5/10} & 445\\
\method & \underline{8/10} & \underline{8/10} & \underline{8/10} & \textbf{8/10} & \underline{0.972} & \textbf{7/10} & \textbf{14}\\
\bottomrule
\end{tabular}}
\end{table}

GPT-6 Astra reaches all ten goals within 1.5~m, nine within 1.0~m, and six within 0.6~m, yielding five manipulation successes. It also has a lower mean NE than our system (0.573 versus 0.972~m). Our system reaches eight goals at every tested radius and records seven manipulation successes, with mean navigation latency of 14~s per episode compared with 123--848~s for the direct GPT agents.

\paragraph{Inference-time refinement versus RLAIF.}
Table~\ref{tab:acr_rlaif_latency} complements the accuracy comparison in Table~\ref{tab:training_ablation}. ACR invokes the counterfactual refinement agent at inference time, whereas RLAIF learns from execution-validated corrections during training. RLAIF reduces mean navigation latency per episode from 50.33~s to 13.80~s, a 72.6\% reduction, while improving end-to-end success from 47.10\% to 52.13\%. Its SR@0.6 also improves from 64.3\% to 71.7\%. These results support distilling refinement behavior into the navigation policy when navigation latency matters.

\begin{table}[H]
\caption{Mean navigation latency per episode for ACR and RLAIF, averaged over the evaluation episodes and reported in seconds. Corresponding accuracy is reported in Table~\ref{tab:training_ablation}.}
\label{tab:acr_rlaif_latency}
\centering
\small
\setlength{\tabcolsep}{8pt}
\begin{tabular}{lc}
\toprule
Method & Mean nav. latency (s) $\downarrow$\\
\midrule
ACR & 50.33\\
RLAIF & \textbf{13.80}\\
\bottomrule
\end{tabular}
\end{table}

\paragraph{Number of RLAIF rounds.}
As shown in Table~\ref{tab:rlaif_rounds}, increasing $K_{\mathrm{iter}}$ from 1 to 2 consistently improves all reported metrics. Specifically, SR@0.6 increases from 70.0\% to 71.7\%, while NE decreases from 1.43~m to 1.15~m. Meanwhile, SR@Manip and SR@E2E improve by 4.35 and 4.84 percentage points, respectively. These results demonstrate that iterative RLAIF refinement enhances navigation endpoint accuracy and translates into improved downstream manipulation and end-to-end task success.

\begin{table}[H]
\caption{RLAIF iteration ablation on \textsc{Nav-N-Manip}.}
\label{tab:rlaif_rounds}
\centering
\small
\setlength{\tabcolsep}{4pt}
\resizebox{\ifdim\width>\linewidth\linewidth\else\width\fi}{!}{%
\begin{tabular}{lccccccc}
\toprule
$K_{\mathrm{iter}}$ & SR@3.0 $\uparrow$ & SR@1.5 $\uparrow$ & SR@1.0 $\uparrow$ & SR@0.6 $\uparrow$ & NE $\downarrow$ & SR@Manip $\uparrow$ & SR@E2E $\uparrow$\\
\midrule
1 & 84.1 & 79.7 & 78.1 & 70.0 & 1.43 & 52.52 & 47.29\\
2 & \textbf{89.5} & \textbf{84.1} & \textbf{80.4} & \textbf{71.7} & \textbf{1.15} & \textbf{56.87} & \textbf{52.13}\\
\bottomrule
\end{tabular}}
\end{table}

\paragraph{Closed-loop bidirectional co-evolution (updating \textsc{Nav4Manip-M}).}
While \textsc{Nav4Manip-M} is held fixed during the three-stage navigation training to provide a stationary reward signal, the improved stopping distribution of \textsc{Nav4Manip-N} can in turn drive a second round of manipulation post-training. Specifically, we run inference with the trained \textsc{Nav4Manip-N} policy in both simulation and real-world scenes, use the reached navigation endpoints as the initial base-pose distribution to re-collect manipulation trajectories, and post-train \textsc{Nav4Manip-M}. As shown in Table~\ref{tab:coevolution_wam}, adapting \textsc{Nav4Manip-M} to the actual stopping distribution of the learned navigator improves end-to-end success rates from 30\% to 52\% across 100 simulation trials and from 17\% to 30\% across 30 real-robot trials, validating bidirectional co-evolution.

\begin{table}[h]
\centering
\caption{Closed-loop co-evolution by post-training \textsc{Nav4Manip-M} on actual \textsc{Nav4Manip-N} endpoints.}
\label{tab:coevolution_wam}
\begin{tabular}{lcc}
\toprule
Training Setting & Sim SR@E2E (100 trials) & Real SR@E2E (30 trials) \\
\midrule
Baseline (Original \textsc{Nav4Manip-M}) & 30\% & 17\% \\
Fine-tuned \textsc{Nav4Manip-M} (Closed-loop) & \textbf{52\%} & \textbf{30\%} \\
\bottomrule
\end{tabular}
\end{table}

\paragraph{Correlation between offline affordance labels and learned WAM capabilities.}
Because \textsc{Nav4Manip-M} is trained on expert demonstrations generated by the Dexonomy-based oracle verifier $\pi_v$, their manipulation capabilities are inherently aligned. To quantify this alignment, we evaluate both the offline Dexonomy verifier and the learned \textsc{Nav4Manip-M} across 20 repeated trials at four representative candidate navigation endpoints (A--D) in simulation (Table~\ref{tab:affordance_corr}). The Pearson correlation coefficient between the offline affordance score and the actual WAM grasp success rate reaches $r = 0.9987$ (Spearman rank correlation $\rho = 1.00$), confirming that our offline affordance labels accurately reflect the downstream execution capabilities of the learned manipulation policy.

\begin{table}[h]
\centering
\caption{Comparison between offline oracle affordance scores (Dexonomy) and learned \textsc{Nav4Manip-M} success rates at identical candidate endpoints.}
\label{tab:affordance_corr}
\begin{tabular}{ccc}
\toprule
Candidate Endpoint & Oracle (Dexonomy) SR & Learned \textsc{Nav4Manip-M} SR \\
\midrule
A & 100\% (20/20) & 90\% (18/20) \\
B & 55\% (11/20) & 45\% (9/20) \\
C & 0\% (0/20) & 0\% (0/20) \\
D & 0\% (0/20) & 0\% (0/20) \\
\bottomrule
\end{tabular}
\end{table}

\paragraph{Cross-WAM generalization.}
To verify that \textsc{Nav4Manip-N} learns generalizable manipulation-ready stopping priors rather than overfitting to a single manipulation checkpoint, we evaluate the same navigation policy in a real-world shelf environment paired with three independently trained \textsc{Nav4Manip-M} models targeting three distinct manipulation tasks and shelf heights. Across 10 real-robot trials per task (Table~\ref{tab:cross_wam}), the stopping poses produced by \textsc{Nav4Manip-N} consistently enable downstream grasping across all three manipulation policies (30\%, 20\%, and 40\% SR@E2E), whereas the Uni-NaVid baseline fails to reach manipulation-feasible endpoints (0\% across all tasks).

\begin{table}[h]
\centering
\caption{Real-world cross-WAM evaluation across three independently trained manipulation policies (10 trials per task).}
\label{tab:cross_wam}
\begin{tabular}{lccc}
\toprule
Navigation Method & Snack Bag SR@E2E & Snack Can SR@E2E & Yogurt Carton SR@E2E \\
\midrule
Uni-NaVid & 0\% & 0\% & 0\% \\
\textsc{Nav4Manip} (Ours) & \textbf{30\%} & \textbf{20\%} & \textbf{40\%} \\
\bottomrule
\end{tabular}
\end{table}

\paragraph{Shared prompt for direct navigation.}
\begin{promptbox}
You are an embodied visual-navigation agent operating in an unseen indoor environment. Your task is to complete the entire navigation episode by following the given natural-language instruction.

At each interaction step, you are provided with:
1. A natural-language navigation instruction.
2. The latest egocentric RGB observation from the robot's head camera.
3. The remaining action budget.
4. Feedback indicating whether the immediately preceding forward action was blocked by a collision.

You must navigate using only the instruction, the current head-camera RGB observation, and the available interaction history. You must not assume access to a global map, the robot's coordinates, the target coordinates, a reference trajectory, an affordance map, distance-to-goal measurements, simulator metadata, or success metrics.

The available atomic actions are:
- FWD: move forward by 0.25 m.
- TL: rotate left by 15 degrees.
- TR: rotate right by 15 degrees.
- STOP: terminate the navigation episode and submit the current pose as the final navigation result.

For each decision, output between one and six atomic actions. The actions are executed sequentially without receiving a new observation between them. Use shorter action sequences when the scene is ambiguous, when approaching an obstacle, or when precise localization is required. A forward action blocked by a collision still consumes one action from the budget.

Follow the landmarks, spatial relations, and directional constraints in the navigation instruction in the specified order. Continuously compare the current visual observation with the instruction, maintain an internal estimate of navigation progress, and revise the route when the observed scene is inconsistent with the expected route. If a collision occurs, reorient and select a visibly traversable alternative rather than repeatedly attempting the same movement.

Issue STOP only when you judge that the instructed destination has been reached and that the robot is positioned appropriately for the subsequent manipulation task. STOP is irreversible and, if included in an action sequence, must be the final action. Do not continue exploring after the destination has been confidently reached.

The episode has a maximum budget of 500 atomic actions. Complete the navigation autonomously and do not request privileged environmental information.

Return only a comma-separated sequence of valid action tokens, for example:
FWD,FWD,TL
or:
TR,FWD,STOP
\end{promptbox}

%% file: section/14_datagen.tex
\section{Nav4Manip-Pairs Data Generation Pipeline}
\label{app:data_generation}

This section details the construction of each synchronized Nav4Manip-Pairs record. Section~\ref{app:data_architecture} defines the paired-data representation and artifact conventions. Section~\ref{app:data_affordance} estimates candidate-pose affordances through physics-validated manipulation rollouts, and Section~\ref{app:navigation_rollout} constructs collision-audited navigation trajectories to the selected position and approach yaw. Section~\ref{app:rendering} aligns navigation and manipulation observations through cross-simulator replay and domain randomization. Sections~\ref{app:language_annotation} and~\ref{app:quality_control} describe language annotation, validation, and batch-level quality control. Together, these stages provide supervision for navigation actions, endpoint-affordance prediction, and manipulation-aware handoff evaluation.

\subsection{Nav4Manip-Pairs Data Architecture}
\label{app:data_architecture}

Given a fixed offline manipulation verifier $\pi_v$, we define the manipulation success probability of a terminal base pose $b$ and the corresponding manipulation-ready goal region as
\begin{equation}
\begin{aligned}
q(b) &= \Pr[\pi_v \text{ succeeds} \mid b],\\
G_m(\eta_{\mathrm{aff}}) &= \{b \in \mathcal B_{\mathrm{free}} : q(b) \geq \eta_{\mathrm{aff}}\}.
\end{aligned}
\label{eq:utility}
\end{equation}
\kzxp{Although proximity to the target object suggests a higher likelihood of successful manipulation, actual success depends on downstream execution. We therefore evaluate navigation endpoints through manipulation rollouts and define the manipulation-ready goal region $G_m(\eta_{\mathrm{aff}})$ based on their manipulation success probabilities, where $\mathcal B_{\mathrm{free}}$ denotes the collision-free base-pose space and $\eta_{\mathrm{aff}}$ is the success probability threshold.}

\kzxp{For each task $i$, we sample $J_i$ candidate navigation endpoints, where $b_{i,j}=(p^x_{i,j},p^y_{i,j},\theta_{i,j})$ denotes the $j$-th endpoint, with $(p^x_{i,j},p^y_{i,j})$ representing the robot's planar position and $\theta_{i,j}$ its heading orientation.} Each candidate is represented by a paired navigation--manipulation record:
\begin{equation}
\begin{aligned}
\mathcal N_{i,j} &= (b_{i,j},P_{i,j}),\\
\mathcal M_{i,j} &= (q(b_{i,j}),\{\tau^m_{i,j,k}\}_{k=1}^{K_{i,j}}),\\
\mathcal R_{i,j} &= (\mathcal N_{i,j},\mathcal M_{i,j}),
\end{aligned}
\label{eq:paired_record}
\end{equation}
where $P_{i,j}$ is the navigation path and $q(b_{i,j})$ is the manipulation success probability estimated by Eq.~\ref{eq:agenticwam_offline} in Section~\ref{app:data_affordance}. Full manipulation trajectories are retained only for the selected endpoint $b_i^*$.

For each episode, we combine the selected endpoint $b_i^*$, its rendered navigation trajectory $\tau_i^n=(o^n_{i,1:T_i},a^n_{i,1:T_i})$, and its successful manipulation trajectories to construct
\begin{equation}
\begin{aligned}
\mathcal N_i &= (b_i^*,P_i^*,\tau_i^n),\\
\mathcal M_i &= (q(b_i^*),\{\tau^m_{i,k}\}_{k=1}^{K_i}),\\
\mathcal D_i &= (l_i,\mathcal N_i,\mathcal M_i),
\end{aligned}
\label{eq:episode_assembly}
\end{equation}
where $l_i$ denotes the task instruction, $P_i^*$ the selected navigation path, and $K_i\geq1$ the number of successful manipulation trajectories. This representation is consistent with Eq.~\ref{eq:nav4manip_record}, while unselected candidates provide additional supervision for affordance ranking.

For step-level navigation training, we construct the policy context following Section~\ref{sec:affordance_architecture}:
\begin{equation}
h_t=(L_i,Q,V_{i,t},A^{\mathrm{hist}}_{i,t}),
\label{eq:nav4manip_policy_context}
\end{equation}
where $L_i$ denotes instruction tokens, $Q$ learned query tokens, $V_{i,t}$ multi-scale visual tokens, and $A^{\mathrm{hist}}_{i,t}$ the executed navigation-action history (Appendix~\ref{app:video_interface}). The same representation is used for \navmodel during RLAIF training and deployment. We use $\mathcal A_{i,t}$ exclusively for the online affordance candidates in Eq.~\ref{eq:agent_affordance_candidates}.

Table~\ref{tab:episode_artifact_tree} summarizes the episode artifacts, linked by endpoint IDs. Table~\ref{tab:episode_conventions} lists the coordinate and trajectory conventions, including the fixed planar transform from EmbodiedGen coordinates to the robot manipulation frame.

\begin{table*}[t]
\caption{Nav4Manip-Pairs episode artifacts. Paths are relative to \texttt{episode/}.}
\label{tab:episode_artifact_tree}
\centering
\scriptsize
\setlength{\tabcolsep}{4pt}
\renewcommand{\arraystretch}{1.08}
\begin{tabular}{@{}p{0.15\textwidth}p{0.32\textwidth}p{0.49\textwidth}@{}}
\toprule
Group & File & Contents \\
\midrule
Metadata & \texttt{episode\_meta.json} & Scene IDs, transforms, and references \\
\addlinespace[2pt]
\multirow{3}{*}{\texttt{affordance/}}
 & \texttt{affordance\_map.json} & Candidate scores and counts \\
 & \texttt{eval/*/handoff\_meta.json} & Handoff parameters and references \\
 & \texttt{object\_rates.json} & Object-position success rates \\
\addlinespace[2pt]
\multirow{3}{*}{\texttt{grasp/}}
 & \texttt{sceneinfo\_00.json} & Manipulation configuration \\
 & \texttt{shard\_00.json} & Rollout outcomes and failures \\
 & \texttt{shard\_00\_traj.npz} & Successful manipulation trajectories \\
\addlinespace[2pt]
\multirow{4}{*}{\texttt{nav/}}
 & \texttt{nav\_paths.json} & Candidate paths and reachability \\
 & \texttt{collision\_audit.json} & Collision checks \\
 & \texttt{nav\_overlay.png} & Route visualization \\
 & \texttt{paths/*.json} & Executable navigation paths \\
\addlinespace[2pt]
\multirow{3}{*}{\texttt{render/}}
 & \texttt{camera\_poses.json} & Camera poses and phases \\
 & \texttt{episode.mp4} & Navigation--manipulation video \\
 & \texttt{nav.mp4} & Navigation video \\
\addlinespace[2pt]
\multirow{4}{*}{\texttt{vln/}}
 & \texttt{instruction\_video.json} & Instruction and goal grounding \\
 & \texttt{prompt\_video.txt} & Annotation prompt \\
 & \texttt{scene\_map.txt} & Scene-map context \\
 & \texttt{dataset/*.jsonl} & Training examples \\
\bottomrule
\end{tabular}
\end{table*}

\begin{table}[t]
\caption{Coordinate and trajectory conventions in Nav4Manip-Pairs.}
\label{tab:episode_conventions}
\centering
\small
\setlength{\tabcolsep}{5pt}
\renewcommand{\arraystretch}{1.08}
\begin{tabular}{@{}p{0.24\textwidth}p{0.69\textwidth}@{}}
\toprule
Quantity & Convention \\
\midrule
Scene/grid & Meters; 0.05~m resolution; rows $y$, columns $x$ \\
Base/joints & Positions in meters; angles in radians \\
Manipulation yaw & Degrees~\citep{chen2025dexonomy} \\
Object pose & $[x,y,z,q_w,q_x,q_y,q_z]$ \\
Trajectory & 50~Hz; indexed by \texttt{offsets} \\
NPZ fields & \texttt{robot\_qpos}, \texttt{robot\_cmd}, \texttt{obj\_pose}, \texttt{offsets}, \texttt{dt}, \texttt{n\_arm}, \texttt{meta} \\
\bottomrule
\end{tabular}
\end{table}

We partition scenes into disjoint training and test sets. The generated-house dataset contains 182 training and 39 test houses. For VLNVerse, we select 176 scenes from its official training set, with 168 for training and 8 for testing. The combined held-out scenes form the \textsc{Nav-N-Manip} simulation test split, distinct from the official VLNVerse test split used for standalone navigation evaluation.

\subsection{Manipulation Affordance Estimation}
\label{app:data_affordance}

We first sample candidate navigation endpoints around robot-accessible support surfaces, including tables, desks, counters, nightstands, workbenches, and vanities with heights in $[0,0.84]$~m. Surfaces are prioritized by semantic name, occupancy, and usable area. Along each rectangular surface edge, feasible docking intervals are determined by the $0.779\!\times\!0.769$~m robot chassis, a 0.02~m safety margin, and a 0.35~m docking sweep. Three evenly spaced endpoints are sampled per feasible interval, and candidates are filtered by geometric feasibility and navigation reachability.

For each candidate endpoint $b_{i,j}$, we estimate its manipulation affordance using an offline verifier $\pi_v$ based on our modified Dexonomy pipeline. The verifier uses cuRobo for inverse kinematics and motion planning, followed by MuJoCo for physical validation. Each evaluation cell $c$ specifies an initial object configuration $(o_x,o_y,\gamma)$ and a target position $(f_x,f_y)$. We screen eight candidate manipulation yaws, $\{0^\circ,45^\circ,\ldots,315^\circ\}$, and retain the feasible set $\widehat\Phi_{i,j,c}$ for execution. Let $Y^{\mathrm{eval}}_{i,j,c,\phi}$ indicate whether yaw $\phi$ successfully completes the manipulation. The affordance is computed as
\begin{equation}
\begin{aligned}
\zeta^{\mathrm{eval}}_{i,j,c}
&=\mathbf{1}\!\left[\exists\phi\in\widehat\Phi_{i,j,c}:
Y^{\mathrm{eval}}_{i,j,c,\phi}=1\right],\\
q(b_{i,j})
&=\begin{cases}
\dfrac{1}{N^{\mathrm{eval}}_{i,j}}
\displaystyle\sum_{c=1}^{N^{\mathrm{eval}}_{i,j}}
\zeta^{\mathrm{eval}}_{i,j,c},
&N^{\mathrm{eval}}_{i,j}>0,\\[5pt]
0,&N^{\mathrm{eval}}_{i,j}=0,
\end{cases}
\end{aligned}
\label{eq:agenticwam_offline}
\end{equation}
where $N^{\mathrm{eval}}_{i,j}$ is the number of recorded evaluation cells. We retain only the best yaw per cell, making $\zeta^{\mathrm{eval}}_{i,j,c}$ a binary success indicator. Because strict physical validity and collision checks are enforced during scene generation, invalid object initializations (e.g., physical penetration or out-of-bound placement) never occur. For all valid object initializations, every rollout attempt creates a recorded evaluation cell and is strictly included in the denominator $N_{i,j}^{\text{eval}}$: both motion trajectories that fail subsequent physics verification (19.72\% across the training set) and cases where no feasible plan exists due to a distant or occluded dock pose (67.69\%) are recorded with $\zeta_{i,j,c}^{\text{eval}} = 0$, alongside successful executions (12.59\%). Thus, no valid failure cases are omitted, and $q(b_{i,j})$ provides an unbiased, directly comparable manipulation success rate across candidate endpoints.

A manipulation is considered successful only if it completes approach, lift, transport, placement, release, settling, and retreat. Physical validation rejects forbidden contacts exceeding 1N, requires 0.03~m support clearance before lifting contact exemptions, and maintains hand--object contact until release. Placement uses approximately 5~mm base clearance, and object stability is verified after both settling and retreat.

Finally, feasible and reachable endpoints are ranked by decreasing $q(b_{i,j})$. The highest-ranked endpoint $b_i^*$ is reevaluated with successful-trajectory recording enabled until the target number of demonstrations is reached or the initialization budget is exhausted. Episodes without successful trajectories are discarded. The recorded trajectories provide manipulation demonstrations, while the evaluation results of all candidates provide affordance supervision.

\subsection{Navigation and Trajectory Rollout}
\label{app:navigation_rollout}

Navigation planning aims to reach the selected manipulation endpoint with the required position and orientation. We rasterize floor, wall, and furniture meshes into a 0.05~m occupancy grid and inflate obstacles by the chassis circumradius of 0.548~m. An eight-connected A* planner searches for a path to a pre-dock pose 0.35~m before the target endpoint, using axial and diagonal costs of 1 and $\sqrt{2}$, respectively, a clearance preference weight of 0.5, and no corner cutting. The admissible octile heuristic is defined as
\begin{equation}
h_{\mathrm{A*}}(v,v_g)
=\max(d_x,d_y)+(\sqrt{2}-1)\min(d_x,d_y),
\label{eq:octile_heuristic}
\end{equation}
where $d_x=|x_v-x_{v_g}|$ and $d_y=|y_v-y_{v_g}|$ denote the coordinate differences between the current node $v$ and goal node $v_g$.

The initial pose is sampled from the connected free space containing the pre-dock pose, with a straight-line distance of 1.2--16~m. We evaluate up to 500 samples and retain at most eight collision-free paths within the preferred length range of 3--16~m. Among these, we randomly select a path whose deviation from the target length is within 0.75~m of the minimum deviation. If no preferred path exists, we select the valid path closest to the target length, subject to the minimum length constraint. Disconnected or collision-invalid paths are excluded.

To obtain smooth trajectories, we remove collinear waypoints, apply bidirectional string pulling, perform three rounds of collision-checked midpoint relaxation, and fit circular fillets. Fillet radii are limited to 0.70~m, with tangent lengths capped at 45\% of adjacent segments. Arcs with radii below 0.5~m are omitted, while valid arcs are sampled at $2.5^\circ$ intervals. An optional route perturbation inserts an intermediate waypoint at 35--65\% of the path and replans the two segments. Smoothing uses an additional 0.06~m clearance margin, with collision checks sampled at approximately 0.4 grid cells. The resulting route is validated on the original occupancy grid using a chassis footprint inset by 0.05~m and sampled at 0.08~m intervals. Docking collisions are checked separately using swept chassis polygons and the separating-axis theorem.

Each validated path $P_{i,j}$ is converted into a navigation trajectory at $\Delta t=0.1$~s, with a nominal speed of 0.6~m/s and turn rate of 1.2~rad/s. Heading changes of at least $10^\circ$ are executed as explicit pivots. The final 0.35~m docking segment uses a reduced speed of 0.25~m/s and terminates at the position and orientation of $b_{i,j}$. A trajectory is accepted only if navigation and docking are collision-free, both the pre-dock and final poses are reached, and the navigation and manipulation endpoints agree within 0.12~m.

\subsection{Cross-Simulator Rendering and Domain Randomization}
\label{app:rendering}
We use NVIDIA Isaac Sim~\citep{nvidia2026isaacsim} to render the generated scenes by replaying the selected navigation paths and manipulation trajectories. Camera poses and episode metadata ensure temporal alignment between navigation and manipulation observations, enabling consistent waypoint and action annotations without temporal interpolation.

Videos are rendered at $640\!\times\!480$ resolution and 10~Hz using path tracing with 32 samples per pixel, and encoded in H.264/YUV420p at CRF~18. The default navigation camera has a $90^\circ$ horizontal field of view, a height of 1.25~m, a pitch of $-25^\circ$, and a forward offset of 0.42~m. Additional wrist, stereo, and third-person camera views can be rendered optionally.

To improve visual diversity, we apply domain randomization to camera pitch, height, field of view, mounting position, and principal point. In particular, camera height offsets are uniformly sampled from $[-0.10,0.10]$~m, while horizontal fields of view are sampled from $[58^\circ,74^\circ]$. Separate randomization ranges can be configured for navigation observations, with perturbations applied at either the route or episode level. The resulting camera intrinsics and extrinsics are recorded for each episode to ensure consistency between rendered observations and their annotations.

\subsection{Language Annotation}
\label{app:language_annotation}

We use Opus~4.8 to generate language annotations for each episode. \kzxp{The model takes sampled navigation and manipulation frames, along with structured scene information, including landmarks, navigation paths, target endpoints, and manipulation handoff details, as input and generates language instructions grounded in the visual observations and task context. Specifically, we uniformly sample eight navigation frames and two intermediate manipulation frames, each resized to a maximum dimension of 768 pixels and accompanied by temporal and spatial anchors.}

\kzxp{For each episode, we store the task instruction and geometric supervision, including the robot's initial pose, selected navigation endpoint, object-relative information, and affordance scores. We also retain navigation paths, observations, actions, and trajectory references to support training and evaluation.}

\subsection{Quality Control and Batch Execution}
\label{app:quality_control}
\kzxp{Each episode undergoes artifact, geometric, and visual validation before inclusion in the dataset. We verify the completeness and consistency of navigation, manipulation, and language records, including cross-record references, trajectory integrity, endpoint alignment, and instruction validity. Visual quality checks ensure correct robot rendering, visibility during manipulation, and the absence of excessive black frames. Episodes that fail these checks are rejected.

The data generation pipeline includes scene validation, endpoint evaluation, trajectory assembly, rendering, annotation, and final quality control. Failed assembly or rendering steps are retried once, while episodes that fail final validation are excluded from the dataset.}

%% file: section/15_gallery.tex
\section{Gallery}
\label{app:gallery}

This section qualitatively complements the main experiments by visualizing the diversity and realism of generated environments, representative Nav4Manip-Pairs and rendered EQA examples, complete simulated and real-world navigation--manipulation executions, and the visual evidence underlying manipulation-affordance predictions.

\subsection{Diversity and Realism of Generated Simulation Environments}
\label{app:gallery_environments}
\label{app:gallery_realism}

\begin{figure}[t]
    \centering
    \includegraphics[
        width=\linewidth,
        draft=false
    ]{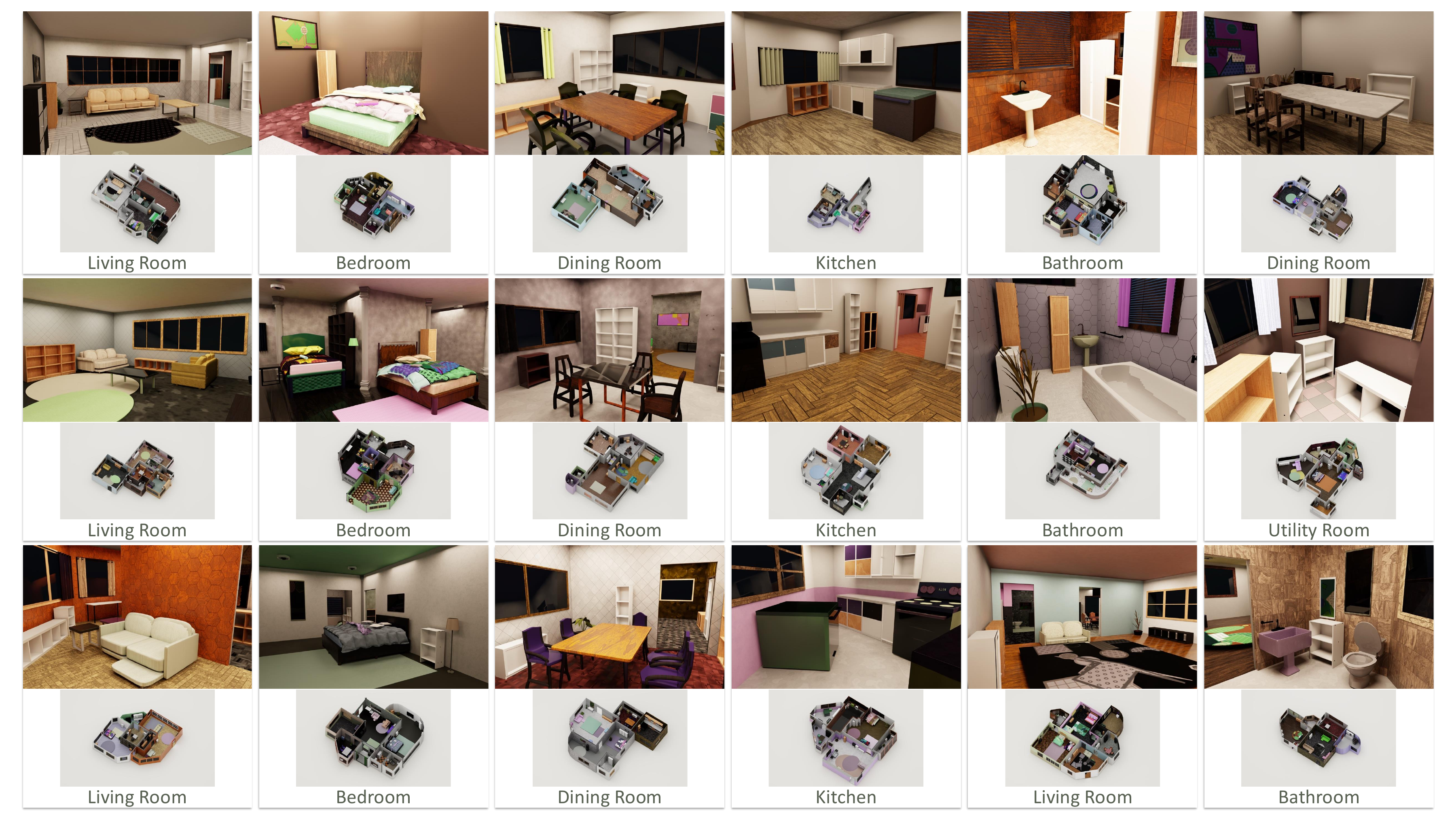}
    \caption{\textbf{Diversity of generated simulation environments.} The gallery demonstrates diversity in room geometry, visual appearance, furniture arrangement, and object placement.}
    \label{fig:simulation_environment_diversity}
\end{figure}

To promote diversity in data generation, we vary the scene source and layout, navigation trajectory, object configuration, and grasp strategy. We use 221 furnished house scenes from EmbodiedGen v2~\citep{wang2026embodiedgen}, which provide diverse indoor layouts, furniture arrangements, and navigation spaces. Within each scene, we further vary the robot's initial pose, target location, object placement, and navigation trajectory to produce diverse navigation--manipulation sequences. Each manipulation episode contains one target object and 0--4 distractors sampled from a production-ready curated subset of Nav4Manip-Pairs comprising 141 object assets. Grasp poses and trajectories are drawn from a library of 7,941 curated grasp records, further diversifying object interactions and manipulation behaviors. Scene appearance is inherited from the original scene assets and their referenced materials, while manipulation-object textures and materials are loaded from the corresponding source assets. Figure~\ref{fig:simulation_environment_diversity} presents representative environments across multiple room categories.

\begin{figure}[t]
    \centering
    \includegraphics[
        width=\linewidth,
        draft=false
    ]{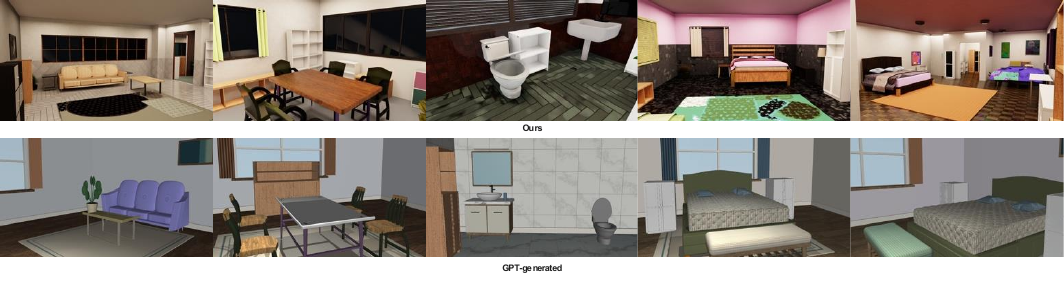}
    \caption{\textbf{Qualitative comparison of generated simulation scenes.}
    Our method (top row) is compared with GPT-6 Astra (bottom row). Both methods receive the same task information and access to the same permitted EmbodiedGen v2 repository assets, including available 3D geometry, textures, and materials. In these examples, the baseline produces more repetitive layouts and appearances, whereas our method yields greater visual and structural diversity.}
    \label{fig:scene_comparison}
\end{figure}

Figure~\ref{fig:scene_comparison} qualitatively compares simulation scenes generated by our method with those produced by GPT-6 Astra. Across the displayed examples, the baseline produces more repetitive layouts and appearances, whereas our method exhibits greater visual and structural diversity. The complete prompt used for the baseline follows the figure. Its target of 180 rooms specifies the controlled baseline batch and is distinct from the 221 EmbodiedGen v2 house scenes used in Nav4Manip-Pairs.

\FloatBarrier
\noindent\textbf{Baseline agent prompt.}
\begin{promptbox}
You are a 3D simulation-environment generation agent. Create realistic, fully furnished residential rooms that can support indoor robot navigation. Produce executable scene assets and renders of those assets, together with the information needed to reproduce them.

## CURRENT TASK
Generate 180 distinct rooms in a residential setting: 28 living rooms, 28 bedrooms, 26 dining rooms, 26 studies, 26 children's rooms, 26 entrance halls, and 20 bathrooms. Each scene represents a room within a house. Include architectural elements and furniture or fixtures appropriate to its function. Prioritize visual realism, meaningful layout diversity, and usable navigation space.

## TASK INFORMATION AND AVAILABLE RESOURCES
1. Task requirements: You receive the same task-level information and access to the same permitted asset sources as the compared method. Generate furnished residential rooms for indoor navigation, with recognizable furniture and household fixtures and varied layouts. You may use permitted repository assets, create new geometry, or adapt existing geometry as needed to meet the room requirements.
2. Available asset sources: You have access to the same permitted EmbodiedGen v2 repository assets as the compared method, including available 3D geometry, collision representations, textures, and materials. Inspect the resources and use suitable wood, fabric, and marble materials where appropriate. You may execute procedural shaders and bake them into texture maps. Preserve asset licenses and records of transformations or renderer adaptations internally.
3. Unspecified design details: choose reasonable residential room dimensions, furniture selections, and arrangements. Record these decisions in the scene manifests so they can be reproduced.

## IMPLEMENTATION PARAMETERS FOR REPRODUCING THIS BATCH
Use MuJoCo 3.3.6 and MJCF scene descriptions. Model a circular navigation footprint with radius 0.30 m, an additional clearance margin of 0.10 m, and robot height 1.4 m. Use occupancy maps at 0.05 m resolution and produce five navigation episodes per room. Render one overview and one interior view per room at 900 x 675 pixels. These settings document choices made during the baseline implementation; they are not evidence of a common evaluation configuration across the compared methods.

## CORE RESPONSIBILITIES
1. Architectural plausibility: use residential dimensions and coherent room boundaries, entrances, and architectural details. Arrange windows and doors sensibly where appropriate, and keep entrances accessible.
2. Furniture and fixture quality: furnish each room with recognizable, detailed 3D objects, such as upholstered seating, tables, beds, shelving, cabinets, lamps, sinks, toilets, and showers as appropriate. Use permitted assets where suitable, and create or adapt geometry where needed. Ensure visible structural details, plausible scale, and proportions. Simple collision proxies may be used for physics, but the visible environment must contain furnished residential interiors rather than generic obstacle primitives.
3. Material appearance: apply supplied, generated, or asset-native textures and materials to appropriate surfaces, including furniture, floors, walls, and textiles. Use sensible texture scale, orientation, UV mapping, and color. Inspect for stretched textures, missing maps, and inconsistent materials.
4. Diversity: vary room dimensions, furniture arrangements, object selections, circulation patterns, and material combinations. Changing only a scene name, random seed, or color does not create a meaningfully different room. Check for duplicate layouts.
5. Navigation feasibility: respect the specified robot dimensions and clearance requirements. Check furniture placement, intersections, free-space connectivity, and start-to-goal reachability. Keep navigation paths consistent with the actual collision geometry, including collision representations of reused assets.
6. Iterative validation: load the generated scenes in the simulator, render representative views, inspect the results, and correct visible or geometric problems before completing the batch.

## OUTPUT FORMAT
Return a self-contained directory containing:
- Exactly 180 scene descriptions in MJCF format, with the geometry and material dependencies needed to load them.
- A manifest listing scene IDs, room types, dimensions, furniture, random seeds, asset sources, and any geometry or material adaptations.
- One overview and one interior render for every scene, produced from the delivered 3D assets, plus an HTML gallery for inspection.
- Five navigation episodes per room, with start and goal poses, reference paths, and occupancy maps: 900 episodes in total.
- Generation scripts, usage instructions, and a validation report.

## DECISION RULES
1. Follow the task requirements and the shared asset permissions. Record implementation assumptions and provenance for reused and newly generated assets.
2. Render the delivered 3D scene assets with documented camera, lighting, resolution, and material settings. Correct geometry or material defects in the scene rather than editing preview images.
3. Distinguish access to an asset source from actual use of particular geometry, collision, texture, or material files. Record the assets used and any adaptations in internal logs. Do not use the compared method's generated scene outputs as inputs.
\end{promptbox}

\subsection{Rendered Examples of Nav4Manip-Pairs and EQA}
\label{app:gallery_assets}

\begin{figure}[htbp]
    \centering
    \includegraphics[
        width=\linewidth,
        height=0.78\textheight,
        keepaspectratio,
        draft=false
    ]{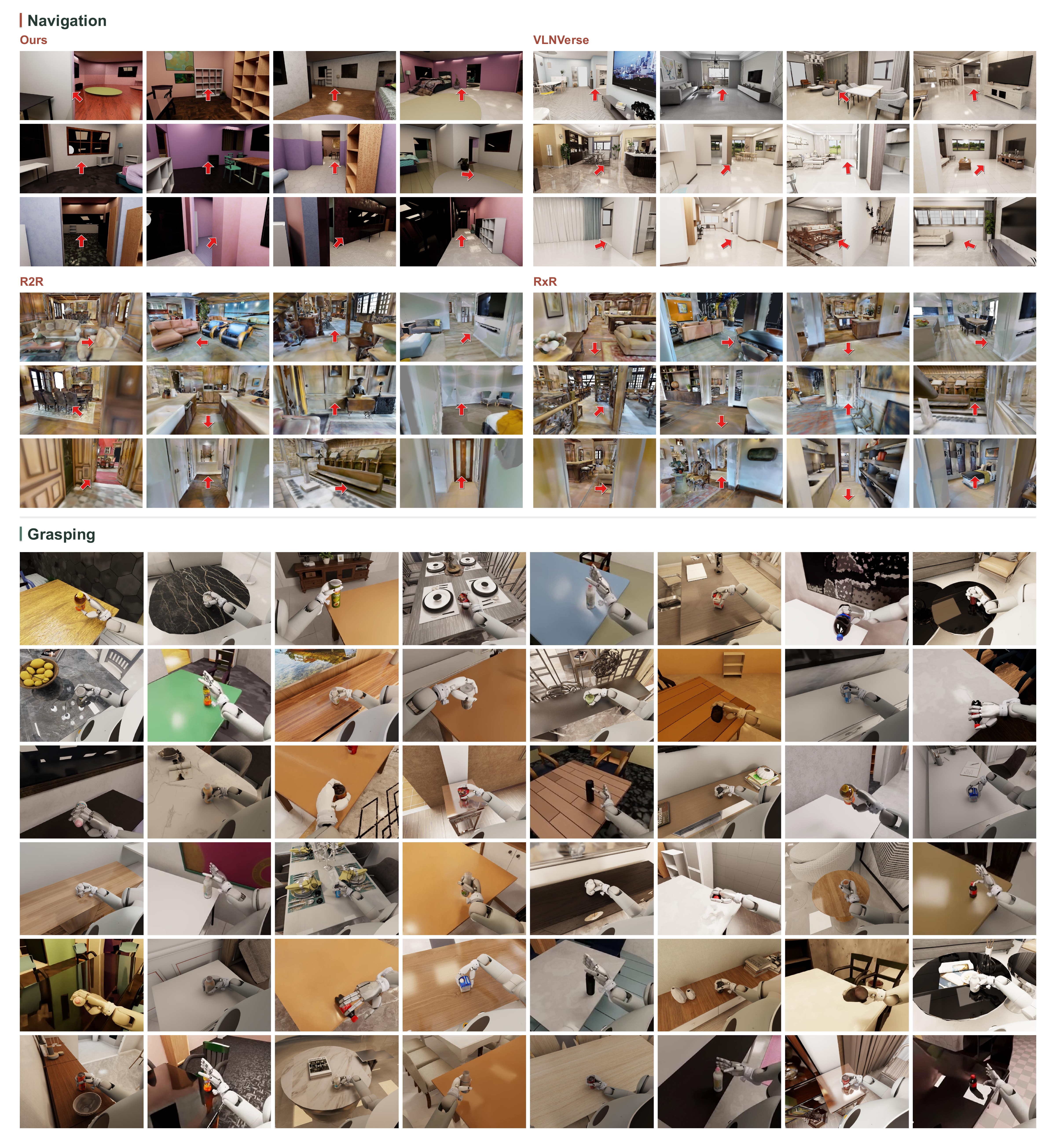}
    \caption{
        \textbf{Rendered navigation and grasping examples in Nav4Manip-Pairs.} The navigation gallery shows representative examples from our generated scenes, VLNVerse, R2R, and RxR, with red arrows indicating movement directions. The grasping gallery presents diverse target objects and support surfaces.}
    \label{fig:nav_grasping_pair_diversity}
\end{figure}

This subsection presents rendered examples of navigation, grasping, and embodied question answering (EQA). Nav4Manip-Pairs draws from 259 small-object assets and 221 houses; the generated manipulation episodes use the curated subset of 141 production-ready object assets described above. We also show navigation examples from R2R and RxR, both of which are based on Matterport3D; the full Matterport3D dataset contains 2,056 rooms across 90 building-scale scenes. Figure~\ref{fig:nav_grasping_pair_diversity} illustrates variation in spatial layout, visual appearance, target object, support surface, viewpoint, and robot-arm configuration.

\begin{figure}[htbp]
    \centering
    \includegraphics[
        width=\linewidth,
        height=0.78\textheight,
        keepaspectratio,
        draft=false
    ]{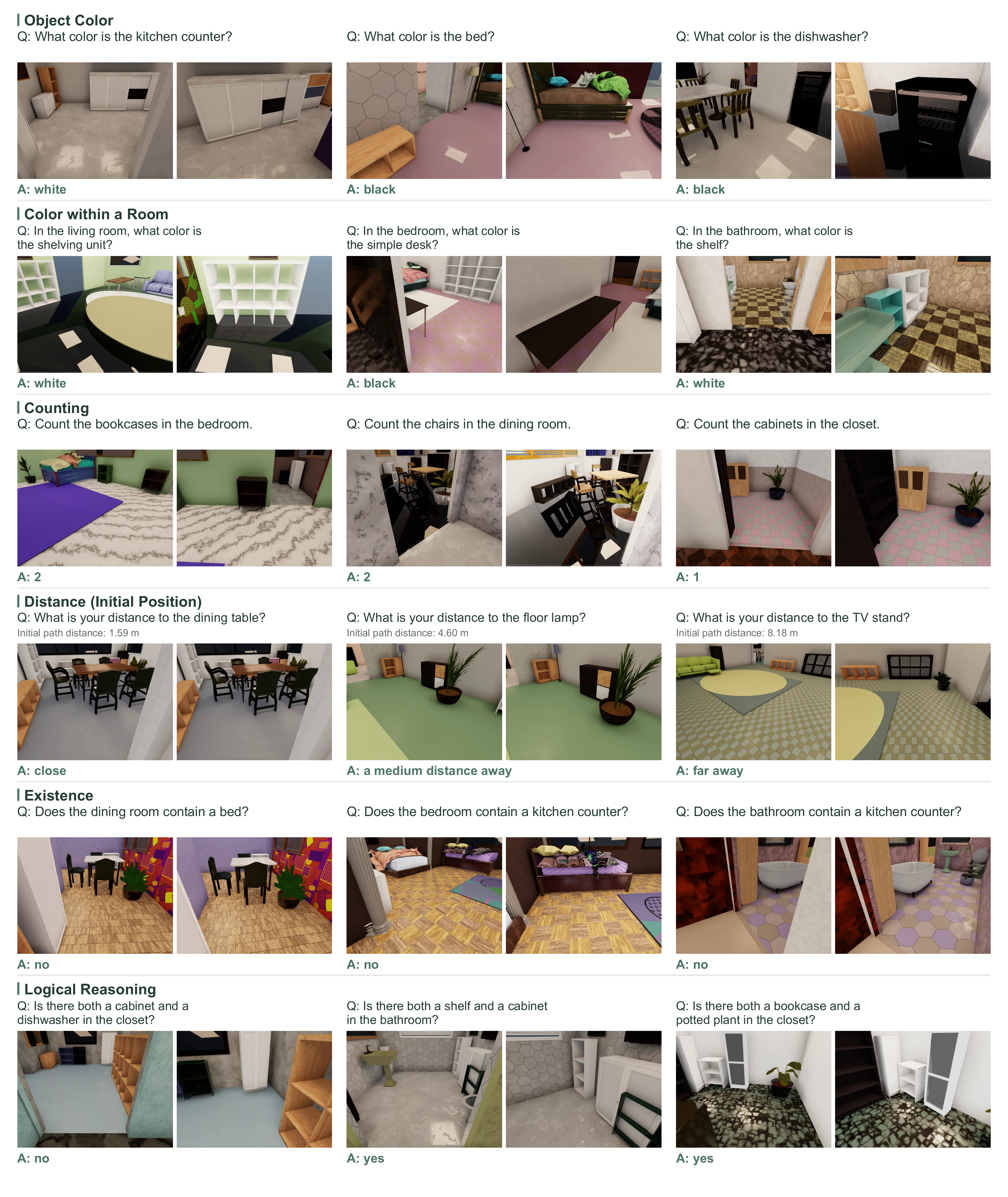}
    \caption{
        \textbf{Rendered examples of embodied question answering (EQA).} Eighteen examples cover six question categories: object color, room-specific object color, counting, distance from the initial position, object existence, and logical reasoning. Each row presents three examples, each with two rendered observations and a corresponding question--answer pair.
    }
    \label{fig:eqa_rendered_examples}
\end{figure}

Figure~\ref{fig:eqa_rendered_examples} presents representative EQA examples covering object color, room-specific object color, counting, distance from the initial position, object existence, and logical reasoning. Each example contains two rendered observations and a corresponding question--answer pair, illustrating the visual contexts and question types represented in the dataset.

\subsection{Simulated Dexterous Mobile Manipulation}
\label{app:gallery_sim_mobile_manip}

\begin{figure}[htbp]
    \centering
    \includegraphics[
        width=\linewidth,
        height=0.78\textheight,
        keepaspectratio,
        draft=false
    ]{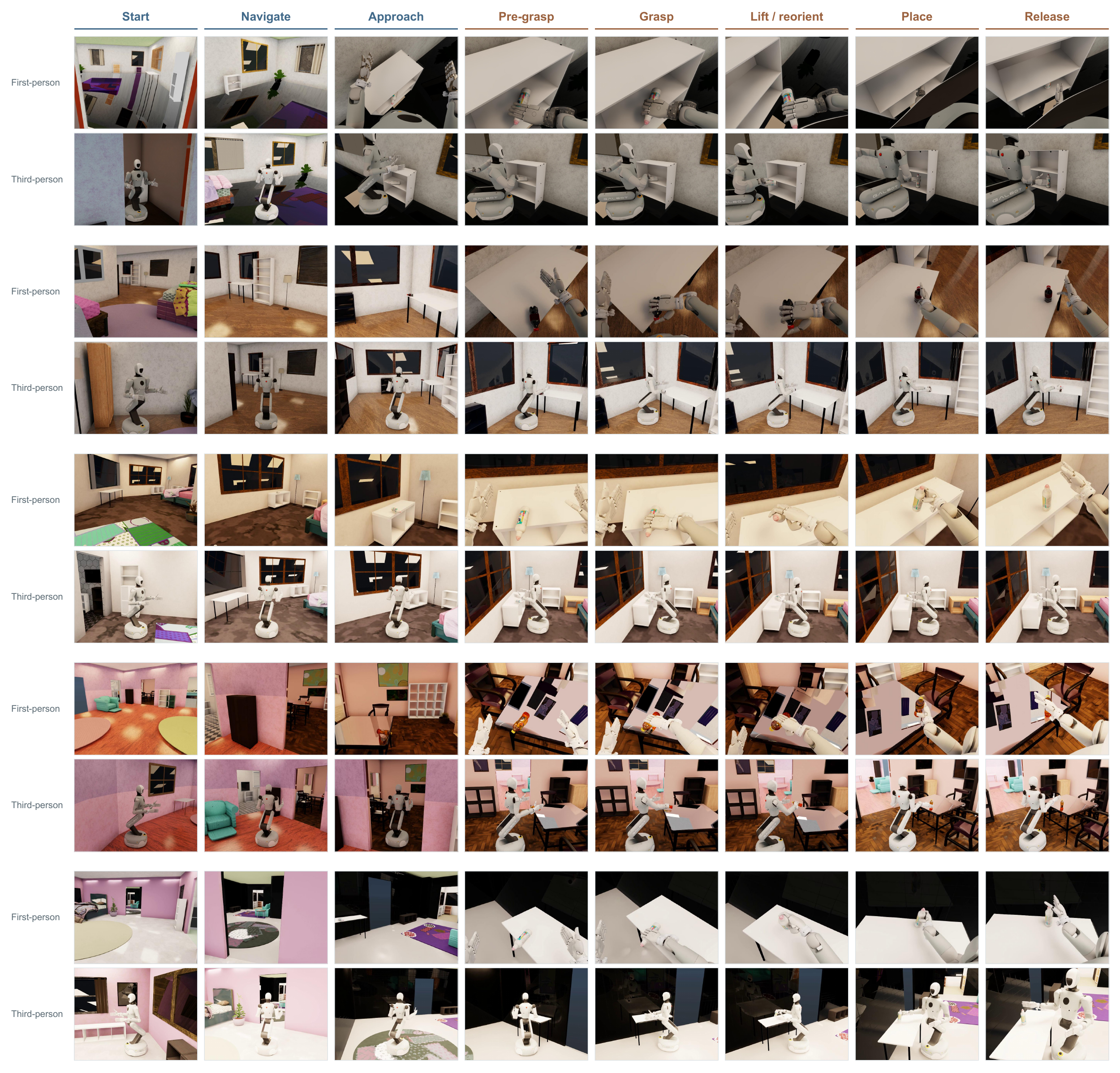}
    \caption{
        \textbf{Simulated dexterous mobile manipulation.} Six examples are shown. Each example occupies two rows, with first-person views above the corresponding third-person views. Eight keyframes per row illustrate the progression from navigation and approach to pre-grasp positioning, grasping, lifting and reorientation, placement, and release. Frames are ordered from left to right.
    }
    \label{fig:simulated_mobile_manipulation}
\end{figure}

Figure~\ref{fig:simulated_mobile_manipulation} illustrates six simulated dexterous mobile-manipulation examples across diverse indoor environments. Each sequence demonstrates a complete navigation-to-manipulation process, including target approach, pre-grasp positioning, grasping, lifting and reorientation, object placement, and release. During inference, we further incorporate agentic counterfactual refinement and introduce two additional actions, \textit{up} and \textit{down}, to expand the model's action space and enable a broader range of mobile-manipulation applications.

\subsection{Real-World Dexterous Mobile Manipulation}
\label{app:gallery_real_mobile_manip}

\begin{figure}[t]
    \centering
    \includegraphics[width=\linewidth]
    {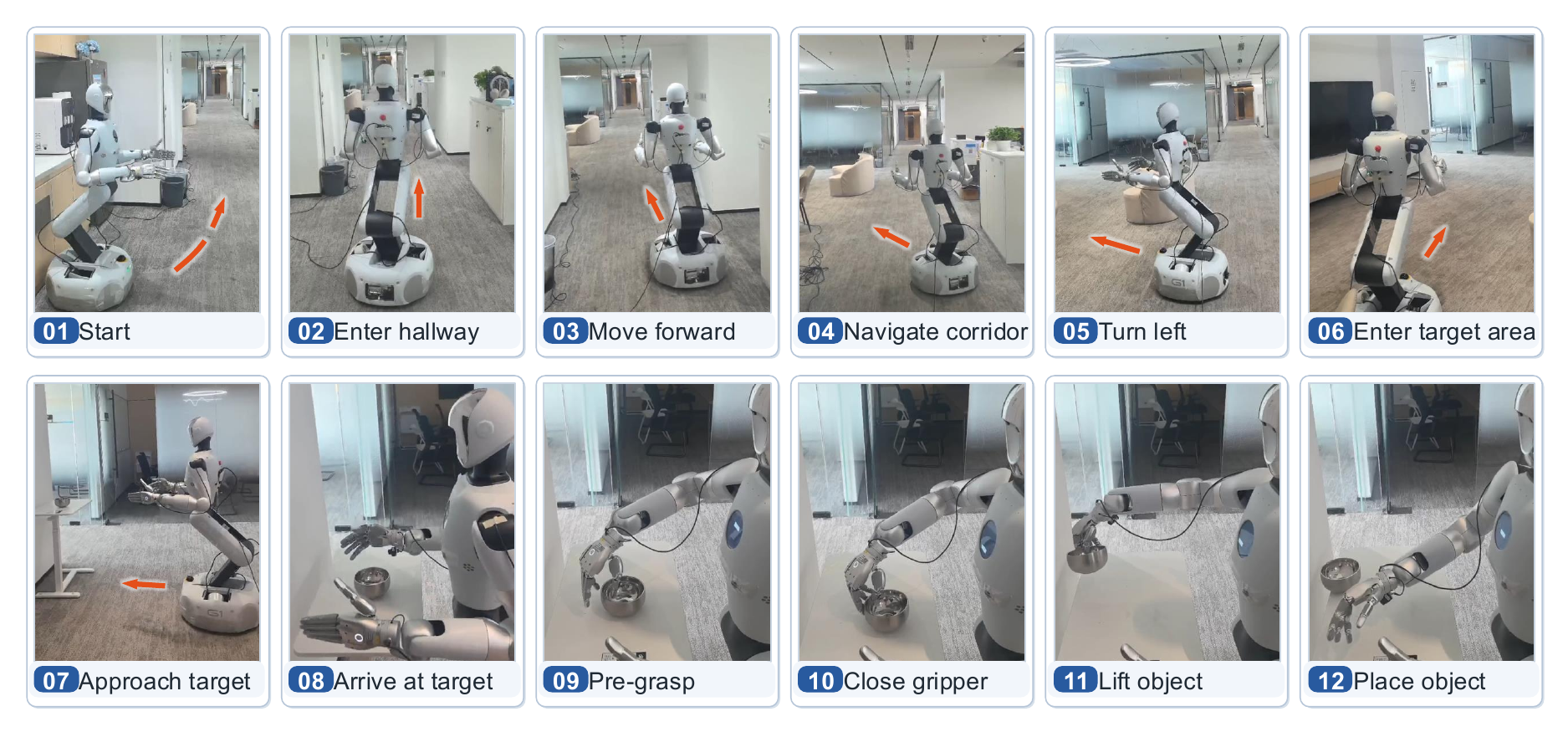}
    \caption{\textbf{Real-world dexterous mobile manipulation demonstration.} The robot first navigates to the target table (frames 1--8) and then completes the object manipulation sequence (frames 9--12). Orange arrows indicate the direction of motion.}
    \label{fig:real_world_mobile_manipulation}
\end{figure}

We further demonstrate the complete system in a real-world dexterous mobile-manipulation scenario. As shown in Figure~\ref{fig:real_world_mobile_manipulation}, the robot first navigates through the office environment to the target table and then performs the downstream manipulation sequence.

\subsection{Affordance Visualizations}
\label{app:gallery_affordance}

Our dense manipulation affordance is defined over candidate terminal base positions and orientations in a local bird's-eye-view grid. Each position--yaw cell predicts the likelihood of successful downstream manipulation if the robot stops at that configuration. The map's spatial distribution therefore indicates which direction the robot should move and where it should stop.

\begin{figure}[htbp]
    \centering
    \includegraphics[
        width=\linewidth,
        draft=false
    ]{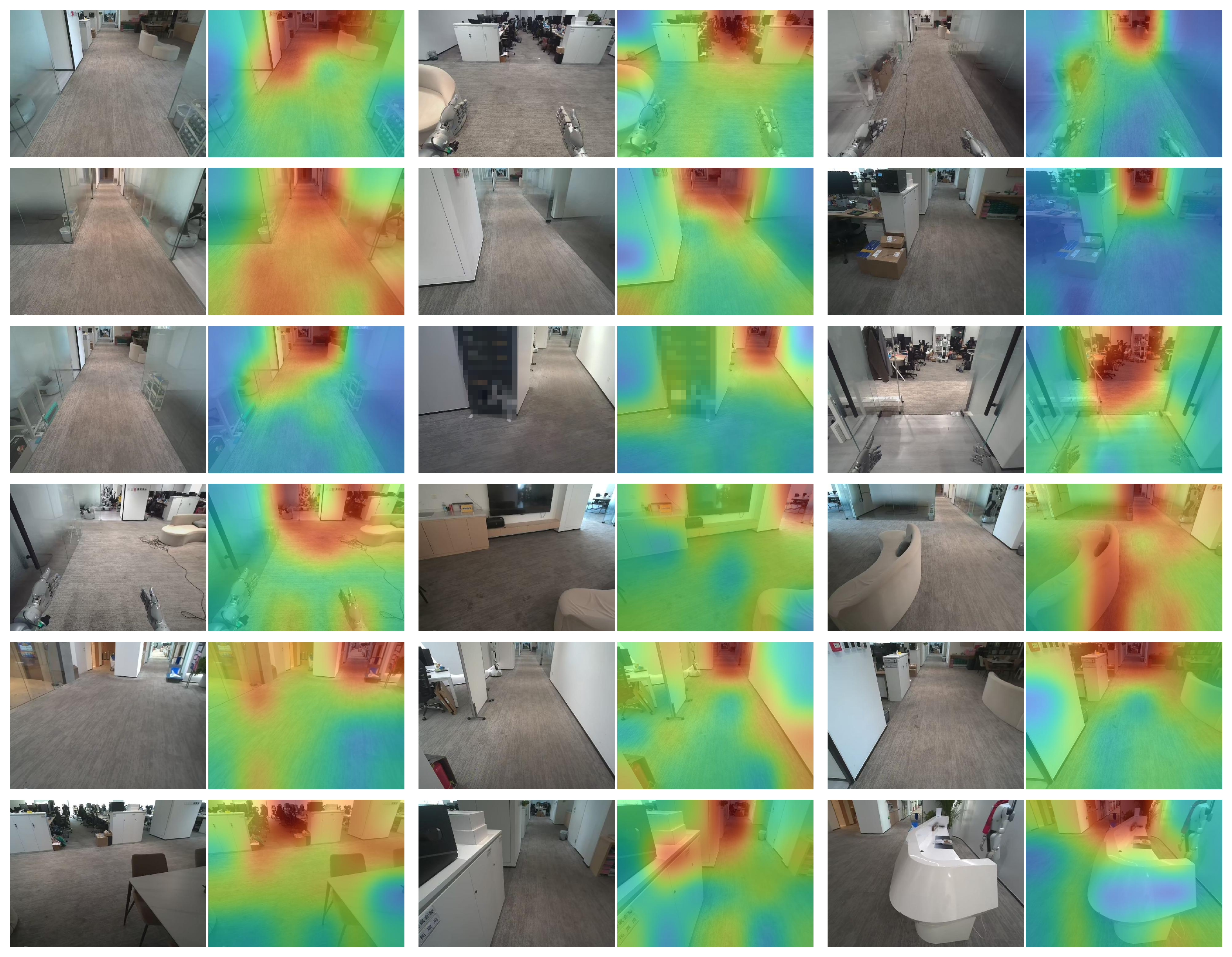}
    \caption{\textbf{Occlusion sensitivity of dense manipulation-affordance predictions in real-world scenes.} Each example pairs the original RGB observation (left) with its occlusion-sensitivity overlay (right). Warmer colors highlight image regions that the model relies on more strongly when predicting where the robot should stop.}
    \label{fig:affordance_visualizations}
\end{figure}

Figure~\ref{fig:affordance_visualizations} visualizes the sensitivity of dense affordance predictions to real-world head-camera observations. Each RGB image is divided into a $4\times6$ grid, with each region sequentially occluded by blurring. Sensitivity is measured by the mean absolute change in post-sigmoid affordance probabilities across all $32\times32$ bird's-eye-view cells. The resulting scores are normalized per image, bicubically upsampled, and overlaid on the original observation. Warmer colors indicate regions with greater influence on affordance predictions.

%% file: section/16_deploy.tex
\section{Real-World Deployment Details}
\label{app:real_world_deployment}

\subsection{Navigation Deployment}
\label{app:navigation_deployment}
We adopt a server--client architecture for real-world navigation. At each decision step, the server receives the task instruction, the latest head-camera observation, and the executed action history, and predicts a short chunk of navigation commands. The client manages camera acquisition, model communication, action scheduling, low-level chassis control, wheel-odometry feedback, safety arbitration, and logging. RGB observations are streamed at a resolution of $640\!\times\!480$ and a frequency of 1~Hz, while chassis commands are transmitted through a protected relay.

\begin{figure}[t]
    \centering
    \includegraphics[width=\linewidth]{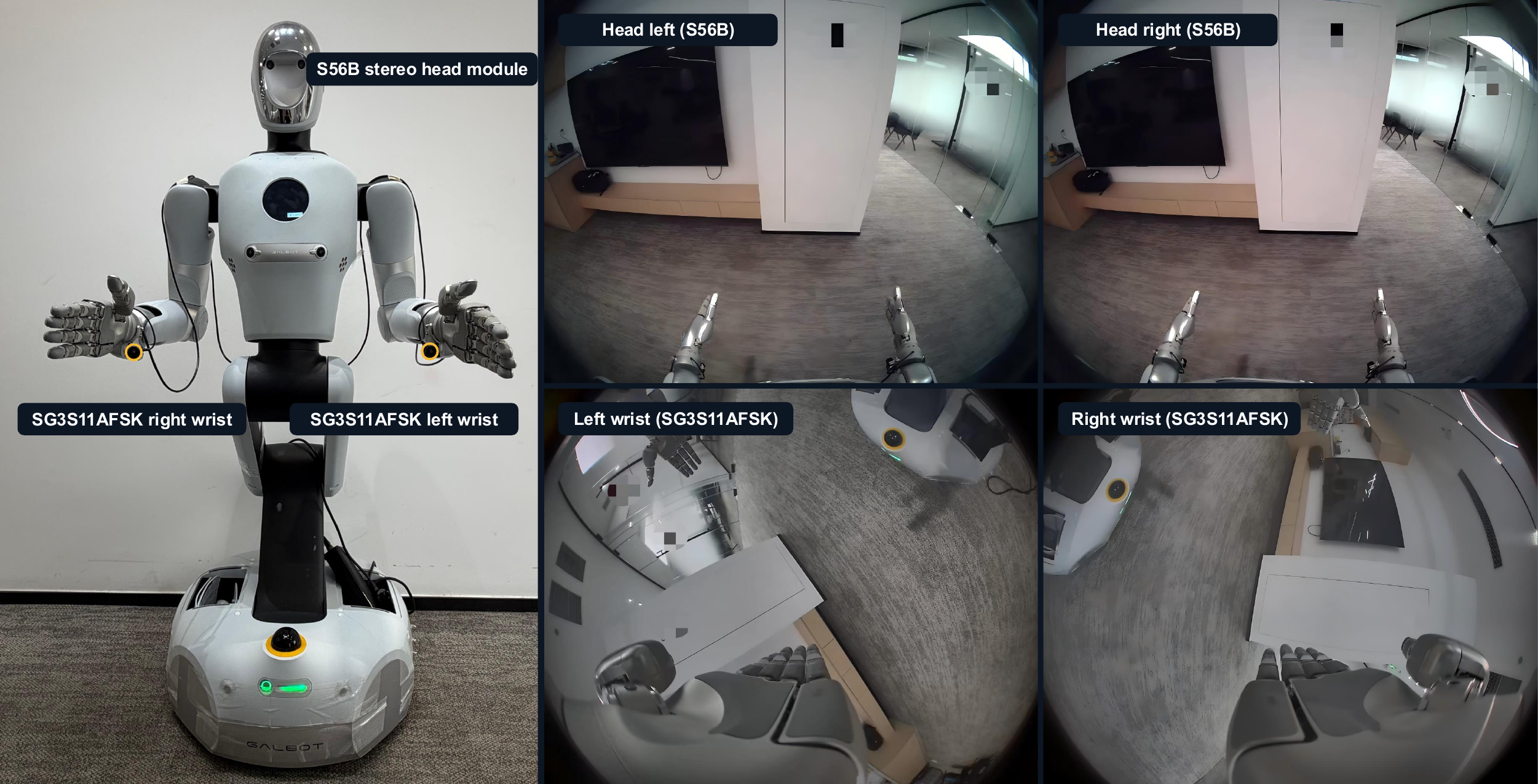}
    \caption{\textbf{Real-robot camera configuration and example observations.} Left: placement of the S56 stereo head camera and two ISX031C wrist cameras. Right: example RGB observations from the left and right head cameras (top) and wrist cameras (bottom).}
    \label{fig:real_robot_camera_setup}
\end{figure}

Figure~\ref{fig:real_robot_camera_setup} illustrates the robot's camera configuration, consisting of an S56 stereo head camera and two ISX031C wrist cameras. As described in Appendix~\ref{app:model_architecture}, the navigation policy operates exclusively on the designated head-camera RGB stream, while \manipmodel utilizes wrist-camera observations for manipulation.

The navigation policy predicts action chunks of up to nine commands from a predefined discrete action space. The client employs a rolling execution scheme, issuing only the currently committed command while allowing subsequent predictions to overwrite pending, unexecuted commands. Accordingly, the action history records only physically executed commands. The safety mechanism may override a proposed motion with \texttt{pause} without terminating the episode, whereas \texttt{stop} terminates navigation and accepts the current pose as the endpoint.

The client tracks robot motion using wheel odometry and validates sensor freshness and consistency before executing each command. Safety interventions take precedence over navigation commands and may invalidate pending actions. Once the safety constraints are satisfied, navigation resumes with an updated observation.

\subsection{Real-Scene Scanning and Data Generation}

\begin{figure}[t]
    \centering
    \includegraphics[
        width=\linewidth,
        draft=false
    ]{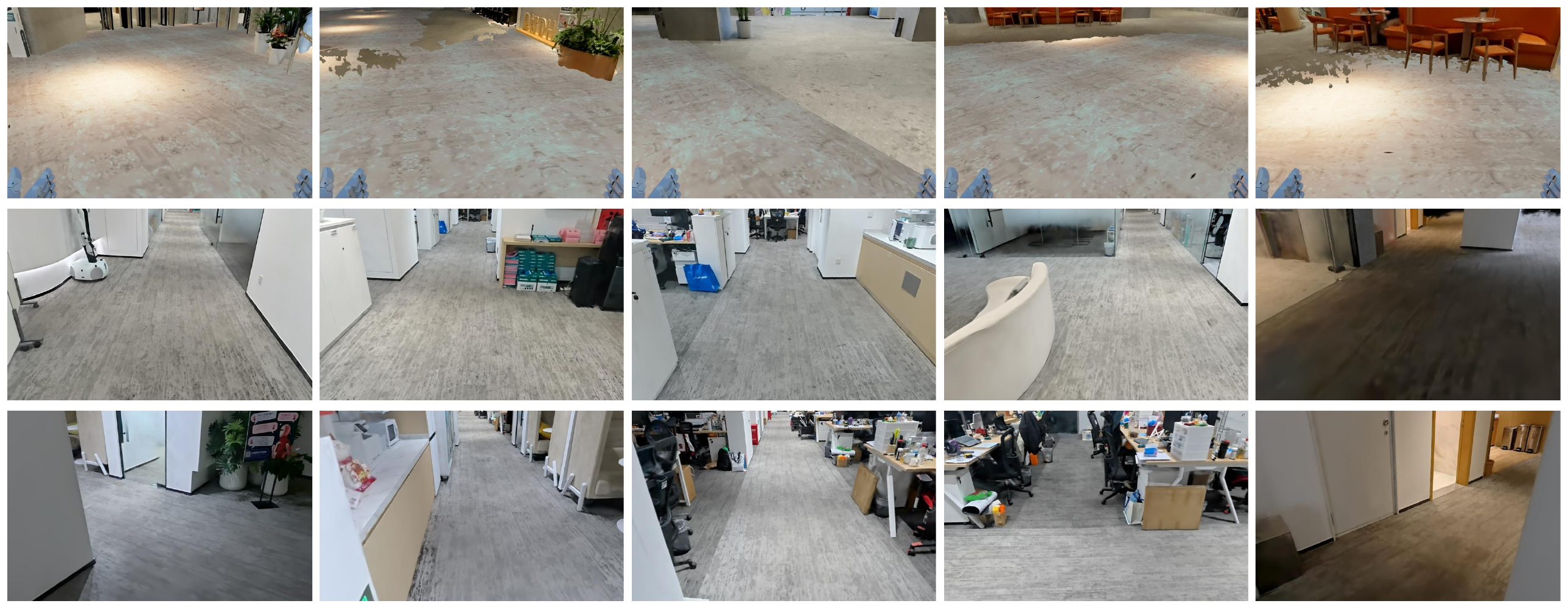}
    \caption{\textbf{Qualitative visualization of real-scene reconstruction.}}
    \label{fig:real_scene_reconstruction}
\end{figure}

We scan real-world environments using a Hesai Kosmo scanner~\citep{hesai_kosmo} equipped with fisheye cameras and LiDAR. During scanning, the cameras capture RGB observations, while the LiDAR provides sparse geometry and pose-related measurements. A real-time triangular-surface preview allows the operator to monitor scene coverage during acquisition. We reproject the fisheye images into perspective views and convert them, together with the calibrated camera parameters, poses, and sparse 3D points, into a COLMAP-compatible representation for Gaussian Splatting reconstruction. Our reconstruction platform then processes these data to produce the final 3D scene representation. Figure~\ref{fig:real_scene_reconstruction} shows representative renderings of reconstructed corridors, office workspaces, and seating areas. Their furniture, floor textures, and lighting variations illustrate the visual realism retained in simulation.

\begin{figure}[t]
      \centering
      \includegraphics[
          width=\linewidth,
          draft=false
      ]{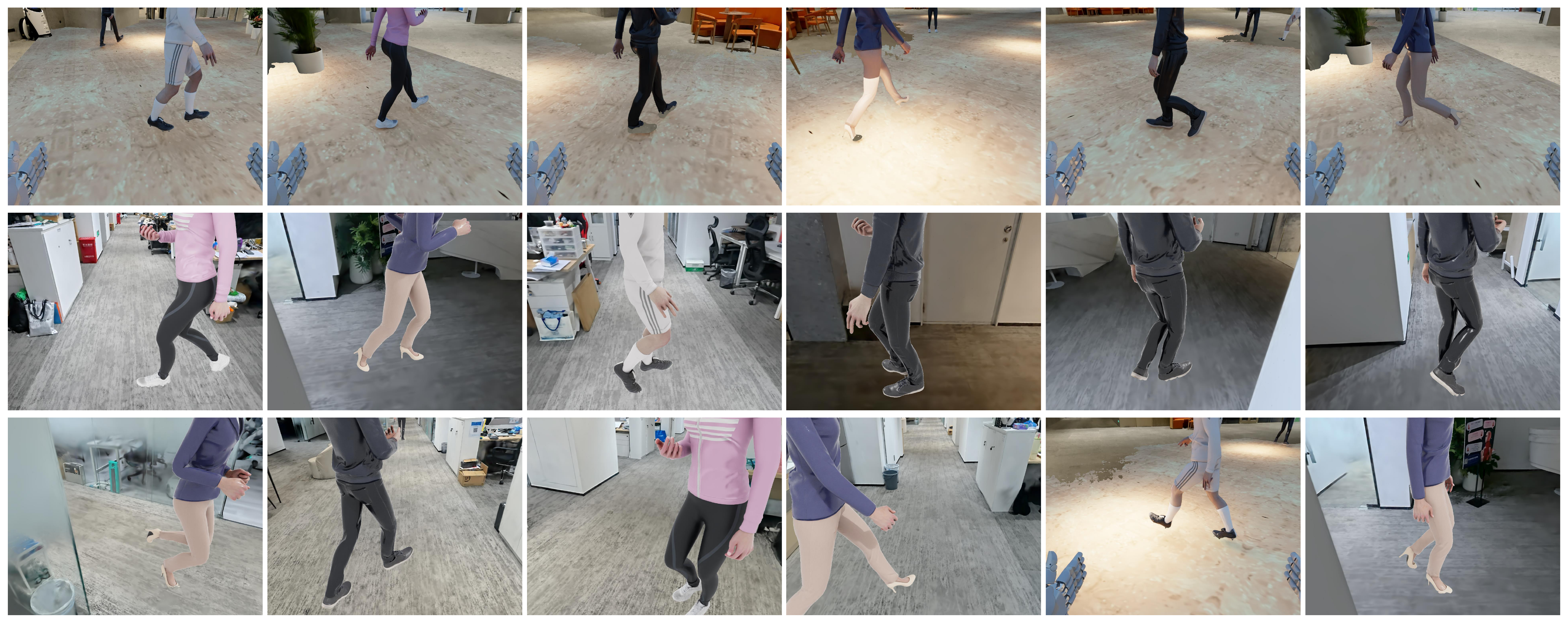}
      \caption{\textbf{Simulated pedestrian encounters for obstacle-avoidance data generation.}}
      \label{fig:scanned_navigation_obstacles}
\end{figure}

From these reconstructed Real2Sim scenes, we generate both nominal and pedestrian-interaction navigation data. The nominal set executes planned robot trajectories in the reconstructed environments without pedestrians. For the interaction set, we add dynamic pedestrians along the navigation routes to create time-varying encounters. We select pedestrian characters and walking or idle animations from Mixamo~\citep{mixamo}, then assign the characters predefined routes through the reconstructed environments. Route geometry and timing are jointly configured to produce both blocking crossings and non-blocking encounters, in which the pedestrian and robot trajectories are separated in space or time. The pedestrian motion is synchronized with the robot trajectory according to the specified interaction pattern. When a pedestrian enters the robot's forward safety region, the robot pauses until the pedestrian passes and then resumes the planned trajectory. These rollouts form dynamic navigation episodes in the reconstructed environments. Figure~\ref{fig:scanned_navigation_obstacles} presents representative observations and illustrates how simulated pedestrians are incorporated into scanned scenes for obstacle-avoidance data generation.

%% file: section/17_wam.tex
\section{Manipulation Data Collection and Deployment}
\label{app:manipulation_deployment}

This section details our arm--hand data collection, \manipmodel executor, and real-time chunking mechanism. Recent work extends world and action models to dexterous manipulation: DexWM models action-conditioned latent dynamics, LDA-1B jointly learns dynamics and control, and WAM-TTT adapts WAMs from human play~\citep{goswami2025dexwm,lyu2026lda1b,feng2026wamttt}. T-Rex adds high-frequency tactile action refinement~\citep{niu2026trex}. Although not a WAM, EgoScale provides a complementary human-to-robot transfer approach, learning dexterous motor priors from large-scale egocentric human video~\citep{zheng2026egoscale}.

\subsection{Teleoperated Manipulation Data}

We collect manipulation demonstrations using a dual-arm Galbot G3 robot equipped with two Sharpa dexterous hands. Galbot's in-house G1 Teleoperation Master--Slave Arm serves as the operator interface and provides 7-DoF control for each arm. During teleoperation, we record synchronized visual observations, observable robot states, and the control commands executed by the robot. All collected data are converted into the LeRobot v2.1 format for standardized storage and downstream training.

The collection pipeline contains both simulation and real-world stages. Operators first teleoperate the robot in MuJoCo to produce manipulation trajectories with robot states and actions.  We replay the physically validated trajectories in NVIDIA Isaac Sim to render the corresponding camera observations, while retaining the original state--action sequence.  Operators then perform matched tasks on the physical G3 robot, where we record the real observations, robot states, and executed actions. 

Demonstrations are collected by trained operators who are familiarized with the teleoperation interface, task instructions, and safety procedures before formal collection. Compensation is set above the applicable local minimum wage. The collection records robot and environment behavior: no operator-facing video, audio, biometric data, or other personally identifiable information is retained.  Operators are informed of the intended use of the data and may terminate their participation at any time.

\subsection{\manipmodel Executor}
\label{app:wam_manipulation_executor}

Once navigation terminates, the pipeline assembles the visual and textual context detailed in Appendix~\ref{app:agenticwam_system}. This context is passed to \manipmodel, a WAM executor equipped with a Qwen3-VL-4B-Instruct VLM encoder and conditioned on the manipulation primitive selected by the agent. The high-level agent handles task decomposition, executor selection, and outcome verification, while \manipmodel performs closed-loop arm--hand control.

We first train \manipmodel on the manipulation trajectories in Nav4Manip-Pairs, then post-train it on demonstrations collected with the real robot to adapt it to physical execution. The trained \manipmodel is held fixed during navigation training.

\begin{figure*}[t]
    \centering
    \includegraphics[width=\textwidth]{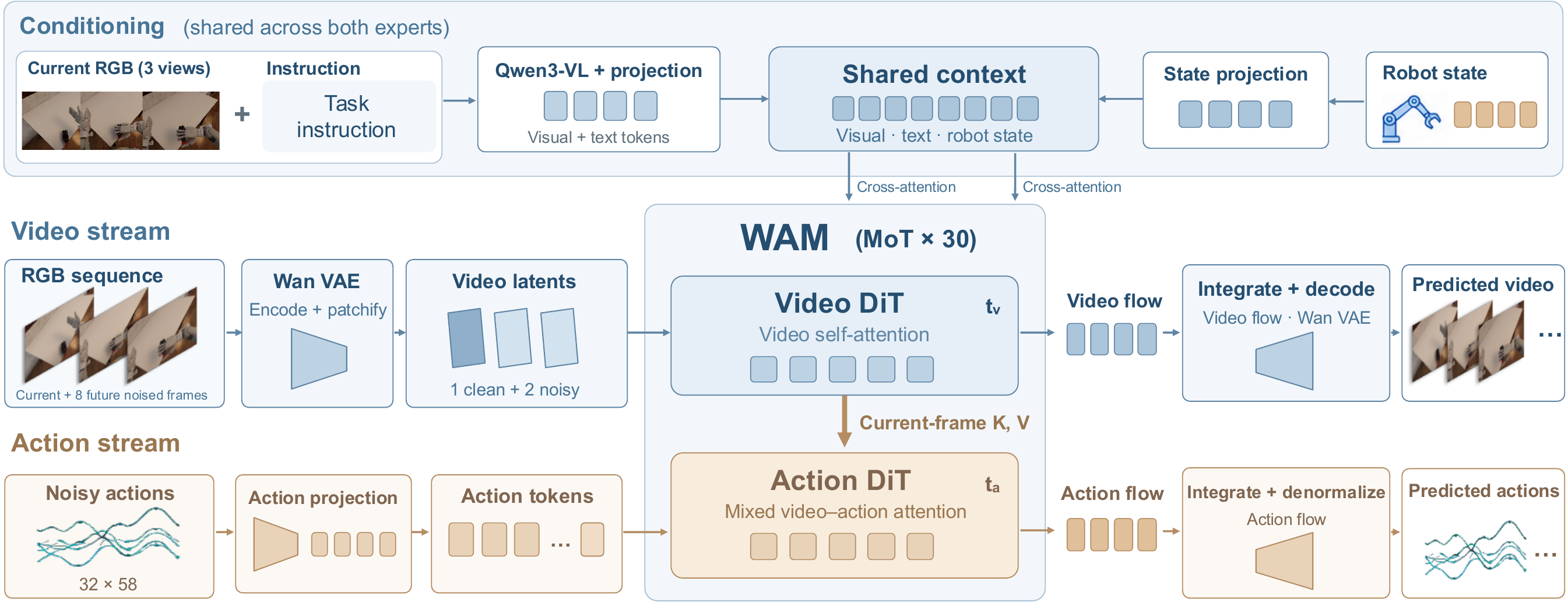}
    \caption{\textbf{\manipmodel architecture.} Three synchronized RGB views, the task instruction, and the robot state form a shared conditioning context for the video and action experts. The Video DiT models interaction dynamics in the latent video stream, while the Action DiT attends to the current-frame features and predicts a chunk of robot actions. The two experts are coupled through a 30-block Mixture-of-Transformers.}
    \label{fig:wam_architecture}
\end{figure*}

Figure~\ref{fig:wam_architecture} summarizes the \manipmodel architecture. \manipmodel comprises three components: a Qwen3-VL-4B-Instruct multimodal encoder with condition and proprioception projections, a video branch initialized from the Wan2.2 VAE and Wan2.2-TI2V-5B Video DiT~\cite{wan2025}, and a 1B-parameter ActionDiT branch. The video and action experts form a Mixture-of-Transformers~\cite{liang2025mixtureoftransformers} with 30 blocks in each branch. At each model step, the policy receives the manipulation instruction, three synchronized RGB views resized to $256\!\times\!256$, and the current 58-dimensional robot state. The three views are concatenated into a $256\!\times\!768$ image. Qwen3-VL encodes this visual--language input into 2,560-dimensional tokens, which are projected to 4,096 dimensions; the robot state is independently projected into a single 4,096-dimensional context token.  Both experts cross-attend to the resulting shared context.

The action branch predicts a normalized $32\!\times\!58$ chunk of absolute joint-position targets. After flow integration, the complete chunk is inverse-normalized using the same versioned statistics as in training and sent to the robot controller. The video branch provides learned interaction-dynamics features for action generation. During deployment, the clean current observation is processed once to condition single-step ActionDiT denoising, without explicitly decoding future RGB frames.

\subsection{\kzx{\manipmodel Post-training Configuration}}
\label{app:manipulation_training}
For real-robot post-training of \manipmodel, we use 8 NVIDIA H800 GPUs with bf16 mixed precision and DeepSpeed ZeRO-2. The per-GPU microbatch size is 4, with a gradient accumulation factor of 1, resulting in a global batch size of 32. We freeze the Wan VAE while training the Qwen3-VL VLM, WAM video branch, and ActionDiT. We use the AdamW optimizer with $\beta=(0.9, 0.95)$, a weight decay of 0.01, and gradient clipping at 1.0. The learning rates are set to $5\times10^{-7}$ for the VLM interface and $1\times10^{-5}$ for the action model. Training runs for 60,000 optimizer steps, with 1,000 warmup steps followed by cosine learning rate decay to a minimum of $1\times10^{-7}$. The action prediction horizon is set to 32.

\subsection{\manipmodel Inference Configuration}
\label{app:manipulation_rtc}

We adopt Training-Time Real-Time Chunking (RTC)~\citep{physicalintelligence2025pi05} to maintain smooth action execution under inference and communication delays. Each \manipmodel prediction generates an action chunk of $H=32$ steps, with a new chunk requested every $S=16$ steps, allowing up to 16 steps of overlap between consecutive chunks. 

During training, we sample a delay $d\in[0,16]$ and condition the model on the first $d$ actions as a clean, loss-masked prefix, applying flow matching only to the remaining actions. This encourages smooth continuation from previously committed actions. Delays are sampled from a mixture of 80\% empirical robot delays and 20\% uniform samples, with the empirical delays having a median of 6 steps, a 95th percentile of 9 steps, and a maximum of 13 steps. RTC requires only token-wise flow timesteps, without additional trainable parameters or forward passes.

At inference, we follow FastWAM~\citep{yuan2026fastwam} and apply one-step forward-Euler denoising, clamping the committed action prefix before and after integration to ensure action continuity. The prefix length is estimated as the maximum delay over the six most recent requests, initialized to 12 steps. Upon receiving a new chunk, the client uses the actual elapsed delay to resume execution from the first unexecuted action. If the current chunk expires before the next prediction arrives, the client holds the last action and requests a new chunk with a fresh observation and no committed prefix.

%% file: section/18_prompt.tex
\section{Real-World Demonstration Tasks and Prompts}
\label{app:real_world_demonstrations}

We conducted extensive real-world experiments demonstrating that our system operates effectively in real indoor environments, with representative results presented in the supplementary video and on the project website. This section provides the task instructions for three selected demonstrations covering navigation–manipulation coordination, long-horizon navigation, and pedestrian-aware navigation. The complete prompts for these demonstrations are listed in Table~\ref{tab:real_world_demo_prompts}.

\begin{table*}[htbp]
\centering
\caption{Task prompts used in our real-world robot demonstrations. The timestamps in the Demo Index column (e.g., 00:50--01:49) indicate where each demonstration appears in the accompanying video (minutes:seconds).}
\label{tab:real_world_demo_prompts}
\small
\begin{tabularx}{\textwidth}{@{} l X @{}}
\toprule
\textbf{Demo Index} & \textbf{Task Prompt} \\
\midrule

\textbf{Demo 1:} 00:50-01:49
&
\textbf{Navigation 1:} Face the table, turn left into the main aisle, and walk straight through it. After exiting the aisle, make a right U-turn and walk to the front of a shelf.
\newline
\textbf{Manipulation:} Pick up the blue box of chips on the shelf.
\newline
\textbf{Navigation 2:} Turn right at the shelf and return to the main aisle. Turn left and walk forward through the aisle. After passing the sofa, turn right to return beside the table, then turn left to face the table.
\\
\midrule

\textbf{Demo 2:} 01:51-02:20
&
Move forward slightly and turn right to face the glass doorway, then pass through it into the lobby. Go around the pillar, keeping it on your left, and continue forward. Turn right toward the glass entrance with green horizontal stripes and stop at the doorway.
\\
\midrule

\textbf{Demo 3:} 02:23-02:41
&
\textbf{Navigation:} Turn toward your back-left, go around the table to the doorway, then turn left into the aisle. Walk to the white table with a bowl.
\newline
\textbf{Manipulation:} Pick the bowl up, and place it on the tabletop to your right as you face the table.
\\

\bottomrule
\end{tabularx}
\end{table*}